\pdfoutput=1
\documentclass{article}
\usepackage[english]{babel}
\usepackage{microtype}
\usepackage{graphicx}
\usepackage{booktabs}
\usepackage{hyperref}
\usepackage[preprint]{icml2026}
\usepackage{amsmath,amssymb,amsthm,mathtools}
\usepackage{array,enumitem,xurl,xcolor,tabularx,multirow}
\makeatletter
\newcommand{\setres}[2]{\expandafter\gdef\csname res@#1\endcsname{#2}}
\newcommand{\res}[1]{\ifcsname res@#1\endcsname\csname res@#1\endcsname\else\textcolor{gray}{[\texttt{\detokenize{#1}}]}\fi}
\newcommand{\rest}[1]{\ifcsname res@#1\endcsname\csname res@#1\endcsname\else\textcolor{gray}{\scriptsize pend.}\fi}
\makeatother
\theoremstyle{plain}
\newtheorem{theorem}{Theorem}
\newtheorem{proposition}[theorem]{Proposition}
\newtheorem{lemma}{Lemma}

\theoremstyle{definition}

\newtheorem{postulate}{Postulate}

\newenvironment{restatement}[3]{\par\smallskip\noindent\textbf{#1~\ref{#2} (#3).}\ \itshape\ignorespaces}{\par\smallskip}
\newcommand{\R}{\mathbb{R}}
\newcommand{\E}{\mathbb{E}}
\newcommand{\bx}{\mathbf{x}}
\newcommand{\ba}{\mathbf{a}}
\newcommand{\bh}{\mathbf{h}}
\newcommand{\bs}{\mathbf{s}}
\newcommand{\btheta}{\boldsymbol{\theta}}
\newcommand{\blambda}{\boldsymbol{\lambda}}
\newcommand{\Lag}{\mathcal{L}}

\newcommand{\sparsemax}{\mathrm{sparsemax}}
\newcommand{\TCz}{\mathsf{TC}^0}
\newcommand{\NCo}{\mathsf{NC}^1}
\newcommand{\wake}{\mathrm{wake}}
\newcommand{\sleep}{\mathrm{sleep}}
\definecolor{mydarkblue}{rgb}{0,0.08,0.45}

\setres{T:rhythm:relerr}{0.01\%}
\setres{T:rhythm:slope}{-0.50}
\setres{T:rhythm:lag}{89}
\setres{T:herd:slope}{-1.00}
\setres{T:samp:slope}{-0.50}
\setres{T:sparse:err}{0.003}
\setres{T:sparse:bound}{0.14}
\setres{T:herd:err}{1.0\times10^{-5}}
\setres{T:samp:err}{1.6\times10^{-3}}
\setres{T:ramp:rate}{3.0}
\setres{T:ramp:other}{0.66}
\setres{T:ramp:pools}{row 1 and box 1}
\setres{T:rest:sweep}{952}
\setres{T:rest:maxprice}{53.2}
\setres{T:bridge:ratio}{1.333}
\setres{T:bridge:err}{0.09}
\setres{T:hinge:err}{0.2\%}
\setres{T:hinge:forb}{0.004}
\setres{T:hinge:n}{43}
\setres{T:hinge:nslack}{125}
\setres{T:hinge:slack}{100\%}

\setres{E1:k:recall}{100.0}
\setres{E1:k:false}{14}
\setres{E1:k:mass}{1.61}
\setres{E1:k:pred}{4.22}
\setres{E1:sudoku:first}{0.0}
\setres{E1:sudoku:valid}{100.0}
\setres{E1:sudoku:recall}{98.6}
\setres{E1:sudoku:pools}{243}
\setres{E1:sudoku:pairs}{7290}
\setres{E1:sudoku:solve}{100.0}
\setres{E2:full}{99.99}
\setres{E2:full:solved}{422{,}759}
\setres{E2:full:early}{7}
\setres{E2:full:capped}{20}
\setres{E2:follow:early}{7}
\setres{E2:follow:capped}{20}
\setres{E2:follow:maxupd}{$3.1\times10^{6}$}
\setres{E2:rest}{100.00}
\setres{E2:rest:solved}{422{,}786}
\setres{E2:rest:maxupd}{$3.1\times10^{6}$}
\setres{E2:rest:median}{$5.0\times10^{4}$}
\setres{E2:rest:check}{1{,}000/1{,}000}
\setres{E2:main}{100.0}
\setres{E2:wrong}{0.0}
\setres{E2:updates}{$4.7\times10^{4}$}
\setres{E2:ms}{104.4}
\setres{E2:spearman}{0.62}
\setres{E2:spurious}{3}
\setres{E2:blocked}{4.0}
\setres{E2:minN}{100}
\setres{E2:minN:excl}{72{,}306}
\setres{E2:minconf}{37.0}
\setres{E2:breakout}{37.3}
\setres{E2:compose}{100.0}
\setres{E2:compose:std}{0.0}
\setres{E2:abl:pools}{91.8}
\setres{E2:abl:pairs}{62.5}
\setres{E2:abl:noise}{24.2}
\setres{E2:abl:rent}{0.0}
\setres{E2:abl:frozen}{0.0}
\setres{E2:abl:noleak}{97.8}
\setres{E3:optimal}{100.0}
\setres{E3:exact}{100.0}
\setres{E3:large}{100.0}
\setres{E3:noprice:valid}{0.0}
\setres{E4:law}{100.0}
\setres{E4:law:40}{100.0}
\setres{E4:tf}{0.0}
\setres{E4:tf:40}{0.0}
\setres{E4:tf6}{0.0}
\setres{E4:tf6:40}{14.3}
\setres{E4:loop}{0.0}
\setres{E4:loop:40}{14.4}
\setres{E4:loopn}{0.0}
\setres{E4:loopn:40}{68.2}
\setres{E4:lstm}{85.8}
\setres{E4:lstm:40}{99.9}
\setres{E5:paradox}{200/200}
\setres{E5:rivalry:cv}{0.55}
\setres{E5:rivalry:low}{0.02}
\setres{E5:rivalry:lowpct}{2.4}
\setres{E5:rivalry:mean}{2.6}

\hypersetup{colorlinks=true,linkcolor=mydarkblue,citecolor=mydarkblue,urlcolor=mydarkblue,
  pdftitle={The Lagrangian Brain: A Constraint Mechanics of Mind and Machine},pdfauthor={Yuhang He, Si Wu}}
\icmltitlerunning{The Lagrangian Brain}
\makeatletter
\renewcommand{\printAffiliationsAndNotice}[1]{\global\icml@noticeprintedtrue%
  {\let\thefootnote\relax\footnotetext{\hspace*{-\footnotesep}\textsuperscript{*}Corresponding author
      \textsuperscript{1}School of Psychological and Cognitive Sciences, Peking University
      \textsuperscript{2}IDG/McGovern Institute for Brain Research, Peking University
      \textsuperscript{3}Peking-Tsinghua Center for Life Sciences, Academy for Advanced Interdisciplinary Studies, Peking University
      \textsuperscript{4}Center of Quantitative Biology, Peking University
      \textsuperscript{5}Key Laboratory of Machine Perception (Ministry of Education), Peking University.
      Emails: Yuhang He \textless{}yuhanghe@stu.pku.edu.cn\textgreater{}, Si Wu \textless{}siwu@pku.edu.cn\textgreater{}.
      \ \\
      \Notice@String}}}
\makeatother
\begin{document}
\twocolumn[
\icmltitle{The Lagrangian Brain:\\ A Constraint Mechanics of Mind and Machine}
\begin{icmlauthorlist}
\mbox{\bf Yuhang He\,\textsuperscript{1,2,3,4,5}}\qquad\mbox{\bf Si Wu\,\textsuperscript{1,2,3,4,5,*}}
\end{icmlauthorlist}
\icmlkeywords{theoretical neuroscience, Lagrange multipliers, constraint mechanics, saddle flow, inhibition, neural oscillations, sleep, herding, learning from positive examples, planning, length generalization, reasoning, Sudoku, mazes, circuit complexity}
\vskip 0.2in
]
\printAffiliationsAndNotice{}

\begin{abstract}
A brain is in one state at a time, yet most of what decides its next state is absent from it: the patterns its world never produces, the statistics of a lifetime, a goal that lies ahead. In mechanics a wire acts on a bead only through a force that appears when the bead presses on it, and Lagrange computed such forces with one multiplier for each constraint. We propose that a brain follows the same mechanics. Everything a circuit knows, is told or wants is a constraint on its single state and acts through its own multiplier, which accumulates the violation and pushes the state back. Neurons and synapses carry the multipliers, each with its own time constant: inhibitory interneurons carry what never happens, adaptation the cost of staying in one state, synapses the statistics of experience, and feedback to apical dendrites the future. From this hypothesis follow Dale's law, the reciprocal wiring and balance of excitation and inhibition, the paradoxical responses of cortex and the frequencies of brain rhythms. Belief becomes the time one trajectory spends in each state, learning becomes sleep proposing laws and waking refuting the false ones, and planning becomes a wave from the goal meeting a wave from the present. Connectionism and symbolism are the limits of frozen and of instantaneous multipliers, and between them the time a circuit runs becomes the depth of its computation. A machine built on these principles, trained without gradients on 1000 solved puzzles, solves all 422{,}786 Sudoku-Extreme test puzzles with one trajectory each. It rediscovers the rules of Sudoku by sleeping and waking, plans shortest paths in mazes of 64 times the training area, and tracks permutations exactly on strings 25 times longer than in training, where Transformers fail. A mind, on this view, is life that learns its own constraints.

\end{abstract}

\section{Introduction}
\label{sec:intro}

\begin{figure*}[t]
\centering
\includegraphics[width=0.88\textwidth]{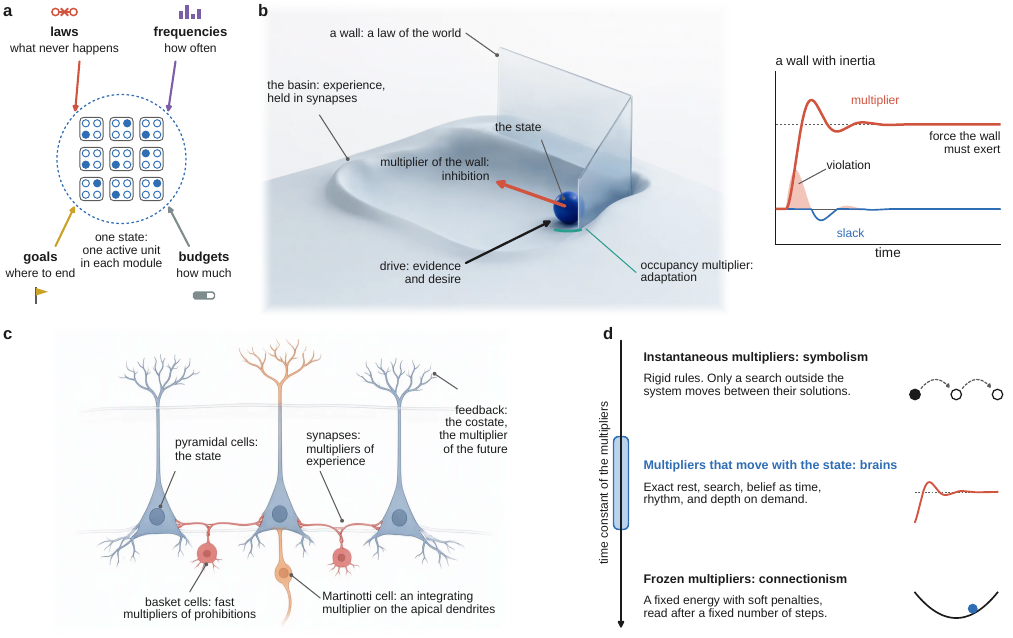}
\vspace{-5mm}
\caption{\textbf{The Lagrangian brain.} \textbf{(a)}~A circuit is in one state at a time, and what decides its next state is mostly absent from it. \textbf{(b)}~Each absent thing enters as a constraint, a wall of the region that knowledge allows, and acts through a multiplier that integrates the violation and pushes the state back. In a brain the multiplier has a time constant, a wall with inertia (right): pressed by the drive, its force grows, overshoots and settles at the value that holds the state on the wall. \textbf{(c)}~The carriers in cortex: pyramidal cells hold the state; basket and Martinotti cells carry prohibitions; synapses carry experience; apical feedback carries the costate of a goal. \textbf{(d)}~The time constant of the multipliers orders three accounts of intelligence.}
\label{fig:principle}
\vspace{-3mm}
\end{figure*}

A brain is one physical object, and at each instant it is in one state. Yet almost nothing that decides its next state is present in that state. A reader of English expects some words and not others, although no group of neurons holds the grammar as a list of rules. A rat at a junction turns toward food it cannot see. A viewer of an ambiguous figure sees one interpretation at a time, and the two alternate. The rules of the world, the patterns it never produces, the statistics of a lifetime and a goal in the future are all absent from the state, and all of them act on it. This paper is about how the absent acts on the present.

Neuroscience has a mechanism for each of these influences: attractors hold percepts \citep{hopfield1982}, interneurons balance excitation \citep{isaacson2011}, adaptation moves activity on \citep{benda2003}, plasticity and sleep store and reorganize experience \citep{rasch2013}, and apical feedback carries context and goals \citep{larkum2013}. Each has its own theory and its own time scale. What is missing is the principle that makes them one, a theory of principle in Einstein's sense, which starts from a general law and derives the phenomena from it \citep{einstein1919}.

The principle comes from mechanics. A bead threaded on a wire moves as though the wire were written into its laws of motion, although the wire is nowhere in the bead. The wire acts through a force that appears only while the bead presses on it and is exactly as strong as it needs to be. Lagrange computed such forces with one multiplier per constraint \citep{lagrange1788}, and Hertz set out to build all of mechanics from free motion and constraints \citep{hertz1899}. Economics found the same object in prices, which rise while demand exceeds supply and so carry knowledge that no participant holds \citep{hayek1945,arrow1958}.

We propose that a brain carries everything absent in this way. Whatever a circuit knows, is told or wants is a constraint on its state and acts only through its own multiplier, a variable that integrates the violation and pushes the state back. In a brain a multiplier is carried by a neuron or a synapse and has a time constant: it is a wall with inertia. Inhibitory interneurons carry the multipliers of prohibitions, adaptation the cost of staying in one state, synapses the multipliers of experience, and feedback to apical dendrites the multiplier of the future, which optimal control calls the costate. One state moves forward and many multipliers accumulate behind it; the state climbs a single Lagrangian and the multipliers descend it (Figure~\ref{fig:principle}). We call this hypothesis the Lagrangian brain.

A penalty is a fixed price; a multiplier is a price that keeps rising until the rule is kept. At rest the rule holds exactly, and inhibition, the force that enforces it, accounts for Dale's law, reciprocal wiring, balance and the paradoxical responses of cortex (Section~\ref{sec:consequences}). Until the rule holds the multiplier grows, so no state that violates a law can last: the circuit searches by its own dynamics, and the time it spends in each state is a belief. Because it has inertia, the multiplier overshoots and turns with the state in the gamma and theta bands. Learning sets the multipliers of experience: sleep proposes laws and waking refutes the false ones. Such a law says only what may not occur next to what, so it applies at every size, and the circuit runs until nothing changes: its running time is the depth of its computation. A goal's multiplier travels backward through the synapses that carry activity forward. Section~\ref{sec:brain} follows these results from milliseconds to days.

The same principle builds machines (Section~\ref{sec:experiments}). A multiplier network learns the laws of a task from solved examples and thinks by letting its multipliers move until its state rests, with no gradient and no search procedure besides its flow. Trained on 1000 solved puzzles, it solves all 422{,}786 Sudoku-Extreme test puzzles, \res{E2:full}\% within its budget, where the best reasoning model trained by gradient descent reports 98.75\% \citep{sghaier2026ptrm}. Sleeping and waking, it rediscovers the rules of Sudoku; it plans shortest paths in mazes of 64 times the training area; and a law learned from strings of length at most 40 tracks permutations exactly at length 1000, where Transformers are exact on no string.

\section{The Lagrangian Brain}
\label{sec:principle}

\subsection{One state, many constraints}
\label{sec:state}

The excitatory neurons of a circuit form \emph{modules}, populations whose recurrent excitation and shared inhibition let one unit, or one bump of activity, dominate at a time: a column of visual cortex settling on one orientation, the head direction system holding one bump \citep{zhang1996,kim2017}, a cell of a Sudoku grid holding one digit. Such attractor networks hold the percepts and memories of a circuit \citep{amari1977,hopfield1982,khona2022}. We write the firing rates of the excitatory units as a vector $\bx$ with entries between 0 and 1. At high gain, when rates rise steeply with input, each module holds one active unit and the state is a choice of one value per module.

\label{sec:knowledge}Much of what a circuit knows has a simple logical form. Whatever can happen can also happen with parts left out: a scene with an object unattended, a chord with a note missing. Knowledge with this property, which we call \emph{closed under omission}, is fixed by its \emph{minimal forbidden patterns}, or \emph{prohibitions}, the smallest sets of units never active together (Lemma~\ref{lem:closure}). Two equal digits never share a row of a solved Sudoku; no English word begins with /bn/; an object is never in two places at once. Popper saw that a good theory ``forbids certain things to happen'' \citep{popper1963}, and Ashby that a law of nature is a constraint \citep{ashby1956}. For graded rates a prohibition is an inequality $\ba\!\cdot\!\bx\le b$ with nonnegative weights $\ba$. Experience also teaches \emph{invariants}, sums that never vary (an animal is in one place), and \emph{frequencies}; a task supplies evidence and goals; the body supplies budgets. We write each constraint $k$ through its violation $c_k(\bx)$: it holds when $c_k\le0$, \emph{binds} when $c_k=0$, and is \emph{slack} when $c_k<0$.

\subsection{Multipliers}
\label{sec:prices}

Each constraint acts on the state through a \emph{multiplier} $\theta_k$ that integrates its violation, and the state feels the multiplier as a force against the direction $\nabla c_k$ in which the violation grows:
\begin{equation}
\begin{aligned}
\tau_x\,\dot\bx&=-\bx+f\big(\beta\,[\,\bh+J\bx-\textstyle\sum_k\theta_k\nabla c_k(\bx)\,]\big),\\
\tau_k\,\dot\theta_k&=c_k(\bx)-\rho_k\,\theta_k\qquad(\theta_k\ge0\ \text{for an inequality}).
\end{aligned}
\label{eq:flow}
\end{equation}
Here $\bh$ is the drive from evidence and desire, $J\ge0$ the excitatory associations, $f$ the logistic function, $\beta$ the gain, $\tau_x$ and $\tau_k$ the time constants of state and multiplier, and $\rho_k$ the leak. The first line is the rate equation of an excitatory population with one input added per constraint; the second makes a multiplier grow while its constraint is violated and decay while it is slack. Together they are the saddle flow of the Lagrangian
\begin{equation}
\begin{aligned}
\Lag(\bx,\btheta)={}&\bh\!\cdot\!\bx+\tfrac12\,\bx^{\!\top}\!J\bx+\tfrac1\beta\textstyle\sum_i\mathcal H(x_i)\\
&-\textstyle\sum_k\theta_k\,c_k(\bx)+\sum_k\tfrac{\rho_k}{2}\,\theta_k^2,
\end{aligned}
\label{eq:lagrangian}
\end{equation}
with $\mathcal H$ the binary entropy: the state climbs $\Lag$, the multipliers descend it, and a rest point is a saddle point. Arrow, Hurwicz and Uzawa wrote this flow to compute economic equilibria, in which the multipliers are prices \citep{arrow1958}.

\begin{postulate}[The Lagrangian brain]
(i)~A circuit is one trajectory $\bx(t)$. (ii)~What it knows, is told or wants acts on it only as constraints, and what experience teaches is local. (iii)~Each constraint acts only through its own multiplier, which integrates its violation at its own rate and forgets only through its leak, as in Eq.~\eqref{eq:flow}. (iv)~On a map, a set of sites with a relation between neighbours, the same law holds at every site.
\end{postulate}

Without leak a multiplier grows until its constraint holds, so at rest the constraint holds exactly against any constant drive: this is the internal model principle of control \citep{francis1976}, which makes bacterial chemotaxis adapt exactly \citep{barkai1997,yi2000}. A multiplier is also a memory, the sum of its past violations. Together with the state it pushes, it forms a conjugate pair, like momentum and position, which turns about the rest point at a frequency set by the two time constants. In mechanics the multiplier is instantaneous and the turning cannot be seen; in a brain it has a time constant, and the turning is a rhythm.

\subsection{The dictionary}
\label{sec:dictionary}

\begin{figure*}[t]
\centering
\includegraphics[width=0.9\textwidth]{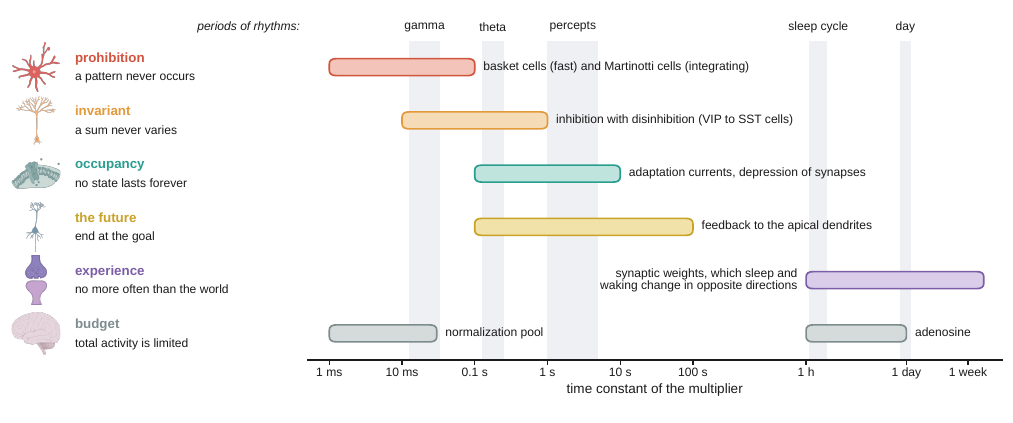}
\vspace{-5mm}
\caption{\textbf{The dictionary across time scales.} Each kind of multiplier (left: its constraint and a drawing of its carrier) is placed at the time constant of its carrier. Basket cells carry prohibitions within a gamma cycle and Martinotti cells integrate them; inhibition with disinhibition (VIP onto SST cells) holds invariants; adaptation carries occupancy over a percept; apical feedback carries a goal over one plan; synapses carry experience over days; a normalization pool and adenosine hold budgets. Grey columns mark the periods of brain rhythms, which appear where a multiplier turns with the state (Theorem~\ref{thm:rhythm}).}
\label{fig:dictionary}
\vspace{-3mm}
\end{figure*}

The postulate makes every variable of a circuit other than its excitatory state the multiplier of a constraint, and the kinds of multiplier differ only in their constraint and their speed (Figure~\ref{fig:dictionary}; Appendix~\ref{app:dictionary}). A prohibition is carried by an interneuron excited by the units of the forbidden pattern and inhibiting them in return: basket cells act within milliseconds, and somatostatin cells, among them the Martinotti cells, integrate over longer times \citep{silberberg2007,tremblay2016}. An invariant needs a multiplier of either sign, which inhibition paired with disinhibition provides, as when VIP interneurons inhibit somatostatin interneurons \citep{pfeffer2013,pi2013}. \emph{Occupancy}, the constraint that a unit's average activity stays below a target so that no state lasts forever, is carried by adaptation and synaptic depression \citep{benda2003,abbott1997}. Experience is carried by synapses onto interneurons, the future by apical feedback, and budgets by a normalization pool and, for energy, adenosine \citep{porkka1997}. The time constants of a circuit tell which constraints it carries.

\section{What Follows}
\label{sec:consequences}

Each result is stated in words, for multipliers without leak unless a leak is named; Appendix~\ref{app:proofs} restates it in full and proves it.

\subsection{Inhibition is the force of constraint}
\label{sec:structure}

The multiplier $\lambda\ge0$ of a prohibition $\ba\!\cdot\!\bx\le b$ integrates $\ba\!\cdot\!\bx-b$, so the neuron that carries it is excited by the units of the forbidden pattern with weights $\ba\ge0$, and its force $-\lambda\ba$ inhibits the same units with the same weights.

\begin{theorem}[Structure]\label{thm:structure}
(i)~The multiplier of a prohibition is carried by a neuron that receives excitation from the units of the forbidden pattern and returns inhibition to them, with input and output weights proportional cell by cell; it obeys Dale's law. (ii)~At rest it is active only if its prohibition binds, and the prohibition then holds exactly; the inhibition each unsaturated unit receives cancels its excitation, up to a term that vanishes at high gain. (iii)~At a stable rest point held by binding prohibitions, with symmetric connections, the patterns of drive to these neurons that lower their own rates are as many as the unstable excitatory modes.
\end{theorem}

Part~(i) is Dale's principle \citep{strata1999} read as the sign of a multiplier; its reciprocal half has been measured \citep{znamenskiy2024}. Part~(ii) is complementary slackness: inhibition is the force of constraint of what a circuit knows, zero until a wall is pressed, never pulling, and returned to exactly the neurons that pressed. The balance of excitation and inhibition \citep{wehr2003,okun2008,xue2014} and the plasticity that restores it \citep{vogels2011} follow. Part~(iii) counts the paradoxical responses of networks stabilized by inhibition \citep{tsodyks1997,ozeki2009}. Budgets obey the same logic. The multiplier of a shared budget $\sum_ix_i\le E$ grows with the summed drive and divides it out,
\begin{equation}
x_i=\frac{E\,h_i}{\sigma+\sum_jh_j},
\label{eq:normalization}
\end{equation}
which is divisive normalization with semisaturation constant $\sigma$ \citep{heeger1992,carandini2012} (Appendix~\ref{app:normalization}).

\subsection{Search and belief in time}
\label{sec:time}

\begin{figure*}[t]
\centering
\includegraphics[width=0.98\textwidth]{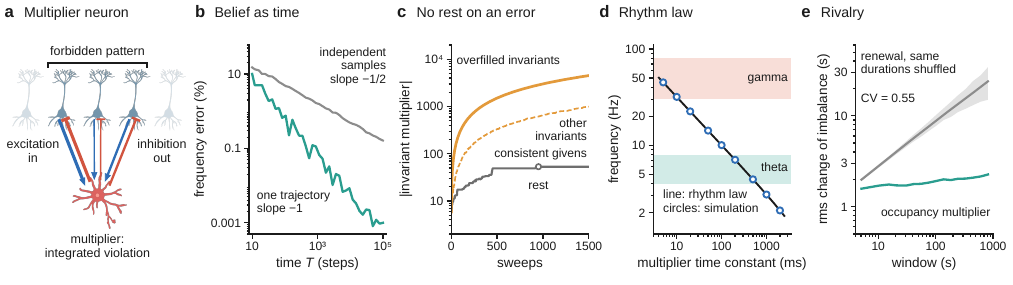}
\vspace{-5mm}
\caption{\textbf{Multipliers at work.} \textbf{(a)}~A multiplier neuron: excitation in, the same pattern of inhibition out. \textbf{(b)}~Belief as time: one trajectory reproduces target frequencies with error falling as $1/T$ (slope \res{T:herd:slope}), independent samples as $1/\sqrt T$ (slope \res{T:samp:slope}). \textbf{(c)}~A Sudoku-Extreme puzzle rests at its solution after \res{T:rest:sweep} sweeps through its 81 cells; one contradictory given makes the multipliers of the two invariants it overfills (\res{T:ramp:pools}) ramp linearly, naming the conflict. \textbf{(d)}~The rhythm law (line) against simulation (circles) over 2.6 decades of the multiplier's time constant; at $\tau_k=5$\,ms the multiplier lags by \res{T:rhythm:lag}$^\circ$. \textbf{(e)}~Rivalry under occupancy multipliers: dominance durations as variable as in human rivalry (coefficient of variation \res{E5:rivalry:cv}) and a cumulative imbalance that grows less than in 100 renewal processes with the same durations.}
\label{fig:structure}
\vspace{-3mm}
\end{figure*}

A state that violates a prohibition cannot stay: the violated multiplier grows until an offending unit gives way. A constraint on a statistic, such as a percept dominating half of the time, is never met by one state, so its multiplier never settles and the time spent in each state realizes the statistic. Both follow from one identity: a multiplier is the sum of its past violations.

\begin{theorem}[Exactness in time]\label{thm:time}
(i)~Over a horizon $T$, the average violation of a constraint is at most $\tau_k$ times the change of its multiplier, divided by $T$. (ii)~A state that violates a prohibition is never a rest point. (iii)~If each state is the one the multipliers favour most, one trajectory reproduces achievable target frequencies with error of order $1/T$, where independent samples err by $1/\sqrt T$. (iv)~Under a leaky occupancy multiplier, time shares approach a sparsemax of the preferences. (v)~If the constraints cannot hold together even on average, a combination of multipliers grows linearly and names the conflict.
\end{theorem}

Thought, on this account, is the itinerary of a state that no multiplier lets rest; an answer is where it rests, and a belief is where it spends its time. Part~(ii) is the breakout principle of local search \citep{morris1993breakout}, with inhibitory neurons as the growing weights. Part~(iii) is herding \citep{welling2009herding} inside a circuit that knows laws, a deterministic correction to theories of neural sampling \citep{fiser2010,aitchison2016}. Part~(iv) is the conditional gradient method of Frank and Wolfe \citep{bach2012equivalence}. Part~(v) turns contradiction into a ramp (Figure~\ref{fig:structure}c).

\subsection{Rhythms are multipliers turning}
\label{sec:rhythm}

Balance is the force of constraint at rest, and a rhythm is the same force in motion: the state presses into a wall, the multiplier pushes it back, the state retreats, the multiplier relaxes, and the state presses again.

\begin{theorem}[Rhythm law]\label{thm:rhythm}
Near a rest point where constraint $k$ binds, state and multiplier turn at the angular frequency
\begin{equation}
\omega_k^2=\frac{\sigma_k^2}{\tau_x\tau_k}-\frac14\Big(\frac{d_k}{\tau_x}-\frac{\rho_k}{\tau_k}\Big)^2,
\label{eq:rhythm}
\end{equation}
with $\sigma_k$ the coupling of the constraint to the excitatory units and $d_k$ the eigenvalue of the excitatory curvature along it. The multiplier lags by a quarter cycle or more whenever it forgets more slowly than the state relaxes, and the mode rings only while its constraint binds.
\end{theorem}

The frequency is set by the geometric mean of two time constants: fast perisomatic inhibition turns in the gamma band, with interneurons firing after pyramidal cells \citep{csicsvari2003,cardin2009}, and adaptation in the theta band \citep{richardson2003}; slowing a carrier shifts its band as $\tau_k^{-1/2}$ (Figure~\ref{fig:structure}d). A fast multiplier rings only while its wall is pressed, so gamma nests within the phase of slower rhythms \citep{canolty2006}.

\subsection{Sleep proposes, waking refutes}
\label{sec:learning}

\begin{figure*}[t]
\centering
\includegraphics[width=0.98\textwidth]{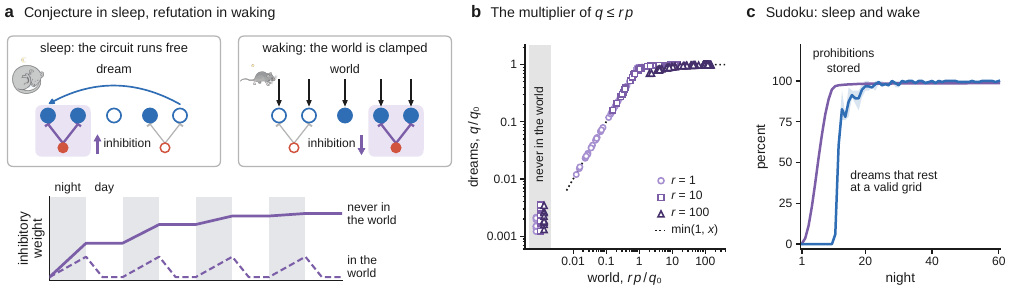}
\vspace{-5mm}
\caption{\textbf{Learning sets multipliers.} \textbf{(a)}~In sleep the circuit runs free and every pattern its dreams contain strengthens the inhibition that forbids it; in waking the world is clamped and every pattern it contains weakens its inhibition. Inhibition survives for patterns the circuit produces and the world never shows. \textbf{(b)}~Each inhibitory weight is the multiplier of $q\le r\,p$: pairs with positive weight are dreamed exactly $r$ times as often as the world shows them (median error \res{T:hinge:err}). \textbf{(c)}~Sleep and wake on 1000 Sudoku solutions: valid dreams rise from \res{E1:sudoku:first}\% to \res{E1:sudoku:valid}\% over 60 nights, and the stored prohibitions level off at \res{E1:sudoku:recall}\% of the 7290 rules, the ones the dreams violated.}
\label{fig:learning}
\vspace{-3mm}
\end{figure*}

Synapses carry multipliers of constraints on the circuit's own activity. In the \emph{free} phase the circuit produces activity by itself, as in sleep; in the \emph{clamped} phase the world drives it, as in waking. Let each prohibition $S$, a set of units, be carried by an interneuron that returns inhibition of strength $I_S\ge0$, its \emph{inhibitory weight}, when all of $S$ is active, and let $I_S$ rise at rate $\eta_s$ while $S$ is active in sleep and fall at rate $\eta_w$ while $S$ is active in waking. With $T_s$ and $T_w$ the time asleep and awake, and $q_S$ and $p_S$ the fractions of sleep and of waking in which $S$ is active, write
\begin{equation}
r=\eta_wT_w/(\eta_sT_s).
\label{eq:ratio}
\end{equation}

\begin{theorem}[Learning]\label{thm:learning}
(i)~Associations are learned in waking and unlearned in sleep; prohibitions are learned in sleep and unlearned in waking; together the two rules descend one contrastive objective. (ii)~Each inhibitory weight is the multiplier of $q_S\le r\,p_S$, and stochastic free activity tends to the distribution closest to the one it has without inhibitory weights, among those that produce no pattern more than $r$ times as often as the world. (iii)~As $r$ grows, the prohibitions become the patterns the circuit has produced and $N$ waking examples never contain; a fresh waking state is wrongly forbidden with probability at most $\sum_Sp_S(1-p_S)^N\le M/(eN)$, summed over the $M$ allowed patterns that carry inhibitory weights. (iv)~A pattern the circuit never dreams is never stored.
\end{theorem}

Part~(i) is the Boltzmann machine \citep{ackley1985} with its weights split by sign, so that the phase without the world writes inhibition. At $r=1$ the circuit learns the distribution of maximal entropy within the world's frequencies \citep{jaynes1957}; as $r$ grows, it learns to forbid what it dreams and the world never shows. Sleep conjectures laws and waking refutes the false ones, Popper's logic of discovery \citep{popper1959}. Languages cannot in general be learned from positive examples \citep{gold1967}; small prohibitions can \citep{valiant1984}, at the rate of Part~(iii). And a circuit stores only what corrects its own imagination.

\subsection{Laws have no size, and time is depth}
\label{sec:depth}

A law learned this way is a list of what may not occur next to what. Applied at every site of a map it makes no reference to size, and a circuit that obeys it computes until nothing changes, so the time it runs is the depth of its computation. Two complexity classes make this precise: $\TCz$ holds the problems solved by threshold circuits of constant depth and polynomial size, $\NCo$ those solved by circuits of gates with two inputs and depth logarithmic in the input size. The product of a string of permutations of five objects, the word problem of $S_5$, is complete for $\NCo$ \citep{barrington1989}, maze connectivity is hard for it \citep{allender2009grid}, and Transformers of fixed depth and logarithmic precision lie in $\TCz$ \citep{merrill2023parallelism,strobl2024}.

\begin{theorem}[Closure]\label{thm:closure}
(i)~On an instance of any size, no configuration that contains a forbidden pattern is a rest point, and every lawful configuration that agrees with the evidence is one when the drive favours none of them. (ii)~Recording what $N$ small solved instances never show recovers the law with high probability once $N$ exceeds the size of a neighbourhood times the logarithm of the number of values, divided by the least frequency of an allowed neighbourhood.
\end{theorem}

\begin{theorem}[Time is depth]\label{thm:depth}
Leaky occupancy multipliers with refractoriness turn a goal into a front that reaches each place at a time equal to its distance from the goal, and descending the multipliers it leaves gives a shortest path. With constant parameters the front decides maze connectivity, and a chain law with 120 states computes the word problem of $S_5$, at every size, in time proportional to the diameter. No family of Transformers of fixed depth and logarithmic precision does either at every size unless $\TCz=\NCo$.
\end{theorem}

\subsection{The future is a multiplier}
\label{sec:future}

\begin{figure*}[t]
\centering
\includegraphics[width=0.98\textwidth]{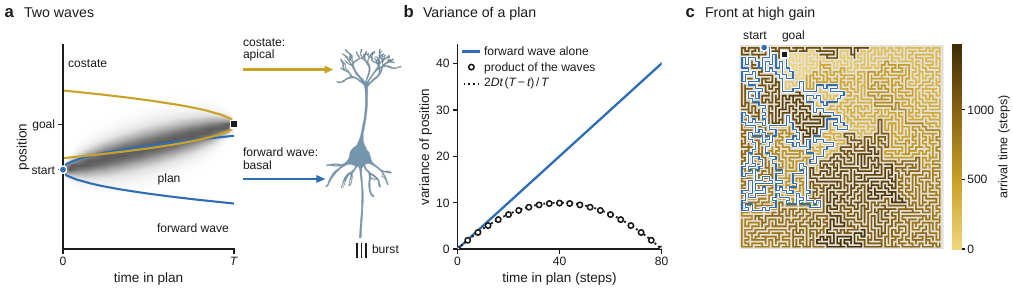}
\vspace{-5mm}
\caption{\textbf{The future as a multiplier.} \textbf{(a)}~Activity spreads forward from the present to the basal dendrites; the costate spreads backward from the goal through the same synapses to the apical tuft; a neuron bursts where the two arrive together. \textbf{(b)}~Their product is the distribution of a plan, with variance $2Dt(T-t)/T$, largest midway and zero at both ends (ratio at $T/2$ and $T/4$: \res{T:bridge:ratio}, predicted $4/3$); the forward wave alone spreads without bound. \textbf{(c)}~At high gain the costate is a front: launched at the goal of a $120\times120$ maze it reaches every open cell once, and descending it gives a shortest path.}
\label{fig:future}
\vspace{-3mm}
\end{figure*}

A goal is a constraint on where a history ends, and it acts through a multiplier like every other constraint. In optimal control this multiplier is carried backward in time by the \emph{costate}, which measures how much each state is worth given where the history has to arrive; in a circuit with learned transitions $K$ it travels backward through the transposed synapses, the direction of apical feedback.

\begin{theorem}[Two waves]\label{thm:future}
For a history that follows $K$ and ends at a goal $\hat\phi$, the multiplier of the dynamics is $-\log\hat\psi$, with the costate $\hat\psi=\hat\phi+K^{\!\top}\hat\psi$, and the history has the marginal $q\propto\psi\odot\hat\psi$, where $\psi=\phi+K\psi$ is the forward wave from the present $\phi$. For a diffusion with coefficient $D$ the goal acts as the drift $2D\nabla\log\hat\psi$, and a plan of duration $T$ has variance $2Dt(T-t)/T$. At high gain both waves are fronts that meet on shortest paths. The expected use of a transition, against which learning compares experience, is the product of forward wave, transition and costate.
\end{theorem}

The forward wave is what can happen from here, the backward wave is what leads to the goal, and their product is what will happen on the way. Schr\"odinger found this structure in 1931 for a diffusion observed at both ends of an interval \citep{schrodinger1931}, and planning as inference rediscovered it \citep{todorov2009,botvinick2012}. Planning and credit assignment become one computation: the backward wave that plans is the signal that learning needs, and it arrives at the apical tuft while the forward wave arrives at the basal dendrites.

\subsection{Connectionism and symbolism are limits}
\label{sec:limits}

\begin{proposition}[Limits]\label{prop:limits}
As $\tau_k\to\infty$ the multipliers freeze into soft penalties and the flow follows a fixed energy, a network of fixed depth when read at a fixed time. As $\tau_k\to0$ they are instantaneous, the state is held to rigid rules, and only an outside search moves between solutions. In between the flow gains exact rest, search, belief in time, rhythm and depth on demand.
\end{proposition}

Connectionist networks descend a fixed energy or compute a fixed function \citep{hopfield1982,rumelhart1986}, symbolic systems hold rules and search \citep{newell1976}, and brains lie between them (Figure~\ref{fig:principle}d), in the regime the dynamical hypothesis anticipated \citep{vangelder1998}. There knowledge composes: a prohibition that does not bind contributes no force, so laws learned apart act together at once, and the systematicity of thought \citep{fodor1988} is the sameness of a law at every site.

\section{The Brain Across Time Scales}
\label{sec:brain}

\begin{table*}[t]
\centering\footnotesize
\caption{\textbf{Predictions}, each from one result (protocols in Appendix~\ref{app:predictions}).}
\label{tab:predictions}
\setlength{\tabcolsep}{4pt}
\renewcommand{\arraystretch}{1.0}
\begin{tabular}{@{}l>{\raggedright\arraybackslash}p{14.4cm}l@{}}
\toprule
P1 & An interneuron that carries knowledge receives weights proportional, cell by cell, to those it returns; sleep keeps the proportion. & Th.~\ref{thm:structure}i\\
P2 & An integrating interneuron is silent until its pooled input reaches a fixed level, then holds it there at any drive. & Th.~\ref{thm:structure}ii\\
P3 & Patterns of interneuron drive that lower the interneurons' own rates are as many as the unstable excitatory modes. & Th.~\ref{thm:structure}iii\\
P4 & Slowing the carrier of a band shifts its frequency as $\tau^{-1/2}$; inhibition lags excitation by about a quarter cycle. & Th.~\ref{thm:rhythm}\\
P5 & In rivalry and theta alternation, cumulative dominance imbalance grows less than in a renewal process with equal durations. & Th.~\ref{thm:time}i,iv\\
P6 & Synapses onto interneurons potentiate on patterns active in sleep and depress on those active in waking. & Th.~\ref{thm:learning}i,ii\\
P7 & Over nights of learning a task, reactivations that violate its rules decline to zero, and rule learning then stops. & Th.~\ref{thm:learning}iv\\
P8 & Apical input to a place cell encodes distance to the current goal and changes before somatic tuning when the goal moves. & Th.~\ref{thm:future}\\
P9 & The positional variance of goal directed theta sequences is largest midway and vanishes at both ends. & Th.~\ref{thm:future}\\
P10 & Contradictory evidence produces linearly ramping inhibition on the violated rules, ending in a switch of interpretation. & Th.~\ref{thm:time}v\\
\bottomrule
\end{tabular}
\vspace{-3mm}
\end{table*}

Read from the fastest carrier to the slowest, the results above give one account of a cortical circuit (in full, with twenty-two findings, in Appendix~\ref{app:brain}).

\textbf{Milliseconds.} Parvalbumin basket cells connect with most pyramidal cells around them, mostly in both directions \citep{holmgren2003,packer2011,hofer2011}, and inhibit most strongly the cells that excite them most \citep{znamenskiy2024}: the reciprocal wiring of Theorem~\ref{thm:structure}, which connectomes of whole cortical columns can now test in full \citep{schneidermizell2025,microns2025} (P1). Cortex operates in the regime stabilized by inhibition \citep{sanzeni2020}, whose paradoxical responses the theorem counts (P3); the count holds in \res{E5:paradox} simulated circuits (Appendix~\ref{app:exp:circuits}). The same cells generate gamma, firing just after the pyramidal cells \citep{csicsvari2003,cardin2009,sohal2009}: a fast multiplier turning with the state (P4).

\textbf{Tens of milliseconds to a second.} Martinotti cells respond little to a single spike and increasingly to sustained firing, and recurrent inhibition shifts to the dendrites in proportion to the rate of a spike train \citep{silberberg2007,pouille2004,kapfer2007}: an integrating multiplier (P2). VIP cells inhibit somatostatin cells and release pyramidal dendrites \citep{pfeffer2013,lee2013,fu2014}, the positive half of a signed multiplier, and divisive normalization \citep{carandini2012} is the multiplier of a budget.

\textbf{Seconds.} Rivalry alternates more regularly than a Poisson process \citep{levelt1965,blake2002}, theta cycles represent alternative futures in turn \citep{kay2020,vollan2025}, replay favours paths not recently taken \citep{gupta2010}, and attention is slow to return where it has just been \citep{klein2000}: one occupancy multiplier, not yet decayed (P5). Signals of response conflict grow with the conflict \citep{botvinick2001}, as the multipliers of incompatible constraints ramp (P10).

\textbf{One plan.} A pyramidal neuron bursts when apical and basal input coincide \citep{larkum1999,larkum2013}, and apical input gates perception \citep{takahashi2016,takahashi2020}: the costate meeting the forward wave (P8), the signal that theories of dendritic credit assignment place on the tuft \citep{guerguiev2017,richards2019,payeur2021}. Plasticity windows of seconds \citep{bittner2017} span the delay of the costate, reverse replay starts at the reward \citep{foster2006,ambrose2016}, and sequences before a run lead to the goal \citep{pfeiffer2013} (P9).

\textbf{Hours to days.} Excitatory synapses grow in waking and are downscaled in sleep \citep{vyazovskiy2008,tononi2014}, inhibition changes across sleep stages \citep{niethard2016}, and sleep favours the discovery of hidden rules \citep{wagner2004}: associations are learned awake and prohibitions asleep (P6, P7). Adenosine accumulates with the energy used in waking and promotes sleep \citep{porkka1997,basheer2004}, so sleep, in which a circuit proposes new laws, is scheduled by the cost of keeping its state lawful.

Four predictions set the theory apart: P2 from the error units of predictive coding \citep{rao1999,bastos2012}, P6 and P7 from replay as rehearsal, and P9 from a forward model, whose uncertainty only grows.

\section{Machines Built from Multipliers}
\label{sec:experiments}

\begin{figure*}[t]
\centering
\includegraphics[width=\textwidth]{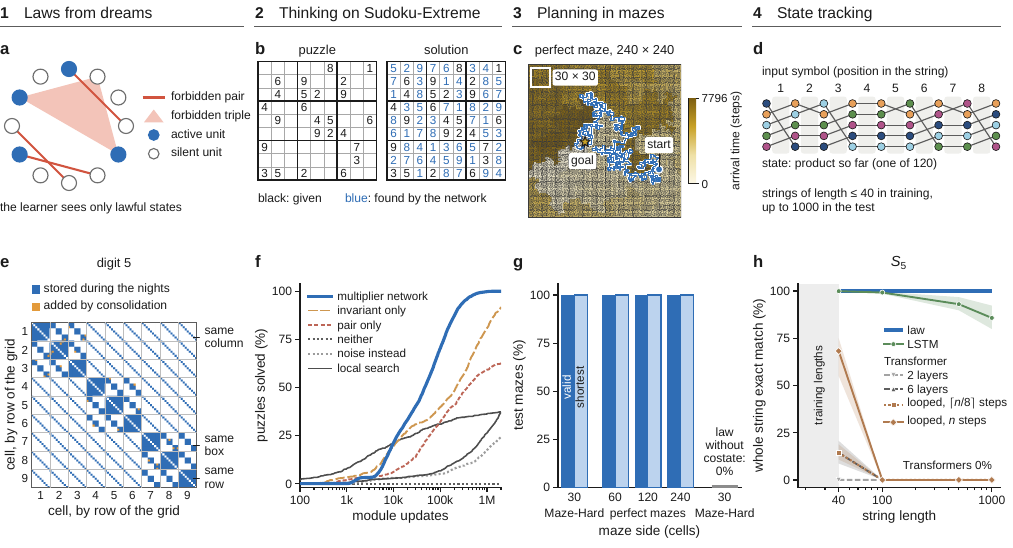}
\vspace{-5mm}
\caption{\textbf{Four experiments.} Top, the tasks; bottom, the main results. \textbf{(a)}~A planted world: binary units, some pairs and triples of which never occur together; the learner sees only lawful states. \textbf{(b)}~A Sudoku-Extreme puzzle and its solution; the network has one unit per digit and cell and is told nothing about rows, columns or boxes. \textbf{(c)}~A perfect maze of $240\times240$ cells, 64 times the training area (square), planned by the front from the goal (shading, arrival time). \textbf{(d)}~A word problem of $S_5$: a string of permutations of five objects and the product so far. \textbf{(e)}~The rules of Sudoku learned by sleep and wake, as prohibitions among the 81 cells for one digit (blue, stored in the nights; orange, added by consolidation): rows, columns and boxes appear. \textbf{(f)}~Puzzles solved in a sample of 1000 against thinking time for the multiplier network, its ablations and local search with the same rules. \textbf{(g)}~Shortest paths at every maze size, and none without a costate. \textbf{(h)}~Exact match of whole strings on $S_5$ against length; training ends at 40 (grey).}
\label{fig:experiments}
\vspace{-3mm}
\end{figure*}

\subsection{A guide to the experiments}
\label{sec:exp:overview}

If knowledge is a set of prohibitions carried by multipliers, a machine built from multipliers alone should do four things that machine learning finds hard: learn the laws of a world from examples of what happens, with no counterexamples; solve hard problems with no search procedure besides its flow; plan by giving the future a multiplier; and apply a law learned on small instances at every size. Experiments~1 to~4 test these claims in turn, with one machine (Figure~\ref{fig:experiments}). Sudoku-Extreme and Maze-Hard are the benchmarks of recurrent reasoning models \citep{wang2025hrm,jolicoeur2025trm}, and $S_5$ is the standard test of state tracking \citep{liu2023shortcuts,merrill2024illusion} (protocols in Appendix~\ref{app:experiments}).

\subsection{The multiplier network}
\label{sec:exp:machine}

An instance of a task is laid out on sites, and each site carries a module whose units are its possible values (a digit per Sudoku cell, a state per position of a string); the units given by the instance are clamped. Knowledge is learned from solved instances by \emph{counting}, the limit of Theorem~\ref{thm:learning} in which every pattern is conjectured and every pattern seen is refuted at once, or by sleep and wake. It takes three forms: \emph{invariants}, groups whose summed activity is the same in every solution; \emph{prohibited pairs} inside invariants; and \emph{window laws}, patterns over a small window that never occur. Thought is the flow of Eq.~\eqref{eq:flow} at high gain, with input
\begin{equation}
h_a=\sum_{P\ni a}w_P\;-\;\sum_b\,\big(\bar\lambda_{ab}+\lambda_{ab}\big)\,x_b
\label{eq:machine}
\end{equation}
to unit $a$, where $w_P$ is the signed multiplier of invariant $P$, $\lambda_{ab}$ the integrating multiplier of a prohibited pair, and $\bar\lambda_{ab}$ the slow multiplier of experience, fixed during thought. Modules update one at a time to their unit of largest input, sweeping through all modules in turn, and the answer is read when ten sweeps change nothing and no multiplier rises; there is no noise, no gradient and no backtracking. A solution is a strict rest point. Away from it, overfull groups push their members down, empty groups pull them up, and each repeated conflict accumulates inhibition that makes it harder to reenter (Appendix~\ref{app:machine}).

\subsection{Experiment 1: laws from dreams}
\label{sec:exp:dreams}

\textbf{What it tests.} Theorem~\ref{thm:learning}: alternating sleep and waking learns the laws of a world from lawful examples alone, with errors of one kind and bounded size. \textbf{The data.} In a planted world (Figure~\ref{fig:experiments}a), $n=12$, 24 or 48 binary units switch on with density 0.3, except that $n/2$ pairs and $n/4$ triples never occur together; each pair inside a forbidden triple is allowed, so triples cannot be inferred from pairs. The learner gets 1000 lawful states. The second world is Sudoku: 1000 solved grids, and nothing about rows, columns or boxes. \textbf{How it learns.} Each night the examples are clamped and lower the weights of the patterns they contain; then the network dreams and raises the weights of the patterns its dreams contain; after 60 nights on Sudoku, groups of mutually prohibited units whose activity sums to one in every solution are consolidated into invariants. \textbf{What happens.} Every hidden law of nine planted worlds is recovered; the \res{E1:k:false} false prohibitions all forbid patterns absent from training and wrongly forbid \res{E1:k:mass}\% of fresh lawful states in the largest worlds, below the bound of \res{E1:k:pred}\%. On Sudoku, valid dreams rise from \res{E1:sudoku:first}\% to \res{E1:sudoku:valid}\%, the stored prohibitions reach \res{E1:sudoku:recall}\% of the 7290 rules, and consolidation completes all 243 invariants (Figure~\ref{fig:experiments}e). \textbf{What it shows.} With no counterexample the learner still learns what never happens: its dreams supply the candidates, and waking removes the false ones.

\subsection{Experiment 2: thinking on Sudoku-Extreme}
\label{sec:exp:sudoku}

\textbf{What it tests.} Theorems~\ref{thm:structure} and~\ref{thm:time}: a circuit whose knowledge is prohibitions searches by itself and rests exactly at lawful states. \textbf{The data.} Sudoku-Extreme \citep{wang2025hrm} has 422{,}786 test puzzles from eight public collections, the hardest known among them (Figure~\ref{fig:experiments}b); reasoning models trained by gradient descent on 1000 puzzles reach 55.0\% to 98.75\%. \textbf{How it thinks.} Counting 1000 training solutions gives exactly the 243 invariants and 7290 prohibitions of Sudoku (the invariants already from \res{E2:minN} solutions). One trajectory runs per puzzle, with a budget of $2\times10^6$ module updates; ablations remove or replace the integrating multipliers, and min-conflicts \citep{minton1992} and breakout \citep{morris1993breakout} use the same rules. \textbf{What happens.} The flow rests at the solution of \res{E2:full:solved} puzzles within the budget (\res{E2:full}\%, Table~\ref{tab:main}), and the other 27 reach it when they run longer. On a sample of 1000 puzzles every seed solves every puzzle, in a median of \res{E2:updates} updates and \res{E2:ms}\,ms on one CPU core. Without integrating multipliers the network solves none; invariant multipliers alone solve \res{E2:abl:pools}\%, pair multipliers alone \res{E2:abl:pairs}\%, noise in their place \res{E2:abl:noise}\%, and min-conflicts and breakout about \res{E2:minconf}\% (Figure~\ref{fig:experiments}f). Laws of standard Sudoku and of the diagonals, counted apart, solve all 200 X-Sudoku puzzles on first encounter. \textbf{What it shows.} The flow alone solves the hardest public Sudoku puzzles: reasoning models learn a function of bounded depth from 1000 puzzles, while the multiplier network learns a law from as many solutions and obeys it for as long as a puzzle needs.

\subsection{Experiment 3: planning in mazes}
\label{sec:exp:maze}

\textbf{What it tests.} Theorems~\ref{thm:depth} and~\ref{thm:future}: a goal acts through a multiplier carried back from it, a front whose parameters do not depend on size. \textbf{The data.} Maze-Hard \citep{wang2025hrm}: 1000 training and 1000 test mazes of $30\times30$ cells with a shortest path as target; and 100 perfect mazes (one path between any two cells) at each side 60, 120 and 240 (Figure~\ref{fig:experiments}c). \textbf{How it plans.} The goal is activated; a silent cell becomes active when a neighbour is active and its occupancy multiplier is low, and then stays refractory; the path descends these multipliers from the start. Only the order of directions for ties and the choice to start the front at the goal are fitted on training mazes. As a control, the network relaxes without a costate the law that every training path obeys, counted over windows of $3\times3$ cells. \textbf{What happens.} Paths are shortest in \res{E3:optimal}\% of test mazes and match the stored path in \res{E3:exact}\% (reasoning models: 74.5 to 86.73\%), the same rule plans in \res{E3:large}\% of perfect mazes at every size, and the law without a costate finds no path (Figure~\ref{fig:experiments}g). \textbf{What it shows.} A local law cannot see where a path should go; the costate supplies the direction, at every size.

\subsection{Experiment 4: state tracking at every length}
\label{sec:exp:s5}

\textbf{What it tests.} Theorems~\ref{thm:closure} and~\ref{thm:depth}: a law learned on small instances is exact at every size. \textbf{The data.} Strings of generators of $\mathbb Z_2$, $\mathbb Z_5$, $S_3$ or $S_5$, with the product so far as the target at each position (Figure~\ref{fig:experiments}d); 10{,}000 training strings of length 1 to 40; 1000 test strings at each length 40, 100, 500 and 1000, correct only if every position is. \textbf{How it runs.} The law forbids every triple of state, symbol and next state absent from training, and the chain relaxes until nothing changes. Transformers of two and six layers, a looped Transformer applied $\lceil n/8\rceil$ or $n$ times, and an LSTM are trained by gradient descent on the same targets, five seeds each. \textbf{What happens.} The law is exact at every length on every group. On $S_5$ no Transformer is exact on any string of length 100 or more; the looped Transformer applied once per symbol falls from \res{E4:loopn:40}\% at length 40 to \res{E4:loopn}\% at 1000, and the LSTM from \res{E4:lstm:40}\% to \res{E4:lstm}\% (Figure~\ref{fig:experiments}h). \textbf{What it shows.} Time gives depth, and the form of a law gives exactness: a counted prohibition does not drift with length.

\begin{table}[t]
\centering\footnotesize
\caption{\textbf{Exact accuracy (\%)} on Sudoku-Extreme (all 422{,}786 test puzzles; $^\dagger$a sample of 1000), Maze-Hard, perfect mazes up to 64 times the area, and $S_5$ at length 40 and 1000. Published numbers as reported; TRM entries are its best variant per task.}
\label{tab:main}
\setlength{\tabcolsep}{2.2pt}
\resizebox{\columnwidth}{!}{%
\begin{tabular}{@{}lcrrrrr@{}}
\toprule
method & grad. & Sudoku & Maze & large & $S_5$@40 & $S_5$@1000\\
\midrule
HRM \citep{wang2025hrm} & yes & 55.0 & 74.5 & -- & -- & --\\
TRM \citep{jolicoeur2025trm} & yes & 87.4 & 85.3 & -- & -- & --\\
PTRM \citep{sghaier2026ptrm} & yes & 98.75 & 86.73 & -- & -- & --\\
Transformer, 2 layers & yes & -- & -- & -- & \rest{E4:tf:40} & \rest{E4:tf}\\
Transformer, 6 layers & yes & -- & -- & -- & \rest{E4:tf6:40} & \rest{E4:tf6}\\
looped, $\lceil n/8\rceil$ steps & yes & -- & -- & -- & \rest{E4:loop:40} & \rest{E4:loop}\\
looped, $n$ steps & yes & -- & -- & -- & \rest{E4:loopn:40} & \rest{E4:loopn}\\
LSTM & yes & -- & -- & -- & \rest{E4:lstm:40} & \rest{E4:lstm}\\
min-conflicts, same rules & no & \rest{E2:minconf}$^\dagger$ & -- & -- & -- & --\\
breakout, same rules & no & \rest{E2:breakout}$^\dagger$ & -- & -- & -- & --\\
\midrule
multiplier network & no & \rest{E2:full} & \rest{E3:exact} & \rest{E3:large} & \rest{E4:law:40} & \rest{E4:law}\\
\bottomrule
\end{tabular}}
\vspace{-3mm}
\end{table}

\subsection{What the experiments show together}
\label{sec:exp:together}

One machine, with one state and a multiplier for every constraint, learns laws from lawful examples, solves every Sudoku-Extreme puzzle, plans at every size and tracks states exactly where trained networks fail.

\section{Related Work}
\label{sec:related}

Multipliers are the forces of constraints in mechanics \citep{lagrange1788,hertz1899}; their saddle flow computes economic equilibria \citep{samuelson1941,arrow1958} and underlies Lagrangian relaxation and learning with low regret \citep{bertsekas1982,freund1999}, and in constrained learning it is known to oscillate \citep{stooke2020}. Neural circuits have computed with multipliers \citep{platt1988,seung1998minimax}, and spiking networks have been designed as solvers of convex programs \citep{boerlin2013,deneve2016}, with constraints chosen by the designer. Balanced networks and networks stabilized by inhibition \citep{vanvreeswijk1996,rubin2015} and cognitive models of constraint satisfaction \citep{marr1976,mcclelland1981,smolensky1986,thagard1998} treat constraints with fixed weights. Learning with a clamped and a free phase \citep{movellan1991,hinton1995,scellier2017} and theories of dreaming \citep{crick1983,hoel2021,deperrois2022} change associative weights in the phase without the world. Planning as inference \citep{attias2003,kappen2005}, the successor representation \citep{stachenfeld2017}, prioritized replay \citep{mattar2018} and dendritic credit assignment \citep{sacramento2018,richards2019} describe pieces of the two waves. Recurrent reasoning networks \citep{palm2018,wang2019satnet,du2022irem,tonshoff2023}, looped and adaptive models \citep{graves2016act,schwarzschild2021,saunshi2025}, studies of length generalization \citep{anil2022,deletang2023} and local and analog solvers \citep{hutter2002,ercseyravasz2011,leleu2019,jonke2016} are the closest relatives of the multiplier network. Here the constraints are learned knowledge, every variable besides the excitatory state is a multiplier, and sleep writes inhibition (Appendix~\ref{app:related}).

\section{Discussion}
\label{sec:discussion}

Features of the brain usually taken as given become consequences. The multiplier of a prohibition cannot be negative, so the neuron that carries it has one sign of output; multipliers differ in constraint and speed, and so do the interneurons that carry them. A multiplier grows until its force cancels the drive against its wall, which balances excitation and inhibition, and with inertia the same force turns with the state, which makes them oscillate. The brain sleeps because laws are proposed by activity that the world does not drive and refuted by activity that it does.

The theory also says where representation lies. The state is one configuration among those the walls allow; the multipliers stand for the world, each for its constraint through the violations it has absorbed. Knowledge, belief and desire are the multipliers of prohibitions, statistics and goals, and the model of the world that Craik placed in the mind \citep{craik1943} and Conant and Ashby required of every regulator \citep{conant1970} is a list of multipliers.

The same mechanics governs life. A cell keeps its essential variables within the region where it can live by integral feedback \citep{cannon1932,barkai1997,khammash2021}, and since concentrations are never negative it meets the problem of Dale's law with two species that annihilate each other \citep{briat2016,aoki2019} (Appendix~\ref{app:life}). A brain is the same architecture turned outward: it learns the prohibitions of its world and keeps itself within them as a cell keeps itself within its viability. A mind, on this view, is life that learns its own constraints, and the many mechanisms by which a brain carries the absent into the present are one mechanism, a multiplier for every constraint, repeated at every time scale from the gamma cycle to the day.

\section*{Impact Statement}
This paper presents work whose goal is to advance theoretical neuroscience and machine learning: a hypothesis about how neural circuits carry constraints, its mathematical consequences, and machines built from it without gradients, tested in simulation and on public puzzle benchmarks. No human participants or animal experiments were involved. There are many potential societal consequences of our work, none of which we feel must be specifically highlighted here.

\section*{AI Use Statement}

\bibliography{references}
\bibliographystyle{icml2026}

\clearpage
\appendix
\onecolumn
\newcommand{\appentry}[2]{\noindent\hspace*{0em}\hyperref[#1]{\textbf{\ref*{#1}}\hspace{0.8em}\textbf{#2}}\dotfill\pageref*{#1}\par\vspace{2pt}}
\newcommand{\appsub}[2]{\noindent\hspace*{1.6em}\hyperref[#1]{\ref*{#1}\hspace{0.8em}#2}\dotfill\pageref*{#1}\par\vspace{-3pt}}
\section*{Contents of the Appendix}
\label{app:contents}
\vspace{2pt}
\noindent\hyperref[app:part1]{\textsc{Part I. Theory}}\par\vspace{4pt}
\appentry{app:proofs}{Proofs}
\appsub{app:proofs:closure}{Closure under omission}
\appsub{app:proofs:structure}{Theorem~\ref*{thm:structure}: structure}
\appsub{app:proofs:time}{Theorem~\ref*{thm:time}: exactness in time}
\appsub{app:proofs:rhythm}{Theorem~\ref*{thm:rhythm}: rhythm law}
\appsub{app:proofs:learning}{Theorem~\ref*{thm:learning}: learning}
\appsub{app:proofs:closure5}{Theorem~\ref*{thm:closure}: closure}
\appsub{app:proofs:depth}{Theorem~\ref*{thm:depth}: time is depth}
\appsub{app:proofs:future}{Theorem~\ref*{thm:future}: two waves}
\appsub{app:proofs:limits}{Proposition~\ref*{prop:limits}: limits}
\vspace{5pt}
\appentry{app:derivations}{Further derivations}
\appsub{app:normalization}{Divisive normalization as the multiplier of a budget}
\appsub{app:signed}{Signed multipliers and disinhibition}
\appsub{app:life}{Life: integral feedback as a multiplier}
\vspace{5pt}
\appentry{app:dictionary}{The dictionary in full}
\vspace{3pt}
\appentry{app:brain}{The brain across time scales in full}
\appsub{app:brain:fast}{Milliseconds: fast inhibition and the gamma rhythm}
\appsub{app:brain:dendrite}{Tens of milliseconds to a second: integration, sign and budget}
\appsub{app:brain:occupancy}{Seconds: adaptation and the cost of staying}
\appsub{app:brain:plan}{One plan: the costate on the apical dendrites}
\appsub{app:brain:sleep}{Days: synapses, sleep and the learning of laws}
\appsub{app:brain:energy}{A day: energy and the timing of sleep}
\appsub{app:brain:summary}{One mechanism at every time scale}
\appsub{app:findings}{Twenty-two findings}
\appsub{app:predictions}{Ten predictions and how to test them}
\vspace{5pt}
\appentry{app:related}{Related work in full}
\appsub{app:related:prices}{Multipliers, saddle flows and prices}
\appsub{app:related:circuits}{Multipliers in neural circuits}
\appsub{app:related:balance}{Inhibition and balance}
\appsub{app:related:constraints}{Constraint satisfaction in cognition}
\appsub{app:related:belief}{Belief, sampling and herding}
\appsub{app:related:sleep}{Sleep and learning}
\appsub{app:related:planning}{Planning, dendrites and credit assignment}
\appsub{app:related:reasoning}{Reasoning networks, depth and length}
\appsub{app:related:solvers}{Constraint solvers in neural and physical systems}
\appsub{app:related:cognition}{Accounts of cognition}
\clearpage
\noindent\hyperref[app:part2]{\textsc{Part II. Machines and Experiments}}\par\vspace{4pt}
\appentry{app:machine}{The multiplier network}
\appsub{app:machine:overview}{How the network works}
\appsub{app:machine:knowledge}{Knowledge}
\appsub{app:machine:thought}{Thought}
\appsub{app:machine:sleep}{Learning by sleep and wake (rule L1)}
\appsub{app:machine:plan}{Planning with the future as a multiplier}
\appsub{app:machine:chain}{Chains}
\appsub{app:machine:impl}{Implementation}
\vspace{5pt}
\appentry{app:experiments}{The experiments in full}
\appsub{app:exp:overview}{A guide to the experiments}
\appsub{app:exp:dreams}{Experiment 1: laws from dreams}
\appsub{app:exp:sudoku}{Experiment 2: thinking on Sudoku-Extreme}
\appsub{app:exp:maze}{Experiment 3: planning in mazes}
\appsub{app:exp:s5}{Experiment 4: state tracking at every length}
\appsub{app:exp:circuits}{Experiment 5: circuit signatures}
\appsub{app:exp:theory}{Numerical checks of the theorems}
\appsub{app:exp:compute}{Compute and data}
\appsub{app:exp:tables}{Result tables}
\clearpage

\phantomsection\label{app:part1}
\begin{center}{\large\textsc{Part I. Theory}}\end{center}
\vspace{2pt}

\section{Proofs}
\label{app:proofs}

This section proves the results of Section~\ref{sec:consequences}. Each subsection restates one result in full, explains in a few plain sentences why it holds, and then gives the proof. Two properties of a multiplier that integrates the violation of its constraint carry most of the arguments. Without leak it rests only where its constraint holds and is positive at rest only where the constraint binds, and at every moment its value records the accumulated history of the violation.

\textbf{Notation.} Throughout, $\bx\in[0,1]^n$ is the excitatory state, $u_i=h_i+(J\bx)_i-\sum_k\theta_k\partial_ic_k(\bx)$ the input of unit $i$, $G=\mathrm{diag}(\beta f'(\beta u_i))$ the excitatory gains at a rest point, and $H=G^{-1}-J$ the curvature of $-U$ there (the Hessian of $-\beta^{-1}\sum_i\mathcal H(x_i)$ is $G^{-1}$). $J$ is symmetric where stated. At high gain each module holds one unit and prohibitions take their binary form: a minimal forbidden pattern $S$ is the constraint $\prod_{i\in S}x_i=0$, whose force on unit $j\in S$ is $-\lambda_S\prod_{i\in S\setminus j}x_i$. Multipliers of inequalities are projected onto $\theta_k\ge0$.

\subsection{Closure under omission}
\label{app:proofs:closure}

\begin{lemma}[Closure]\label{lem:closure}
A family $\mathcal P$ of binary patterns is closed under omission if and only if it is the set of patterns that contain none of its minimal nonmembers (its \emph{minimal forbidden patterns}); equivalently $\bx\in\mathcal P$ iff $\sum_{i\in S}x_i\le|S|-1$ for every minimal forbidden pattern $S$. A polyhedron $K\subseteq\R^n_{\ge0}$ that contains, with every point, all smaller nonnegative points is $\{\bx\ge0:A\bx\le\mathbf b\}$ with $A\ge0$ and $\mathbf b\ge0$.
\end{lemma}

\textbf{Idea.} If leaving units out of an allowed pattern always gives an allowed pattern, then a pattern is forbidden exactly when it contains one of the smallest forbidden patterns, so these patterns describe the whole family. For graded activity the same property says that lowering a rate never leaves the allowed region, so every inequality that limits activity has nonnegative weights.

\begin{proof}
If $\mathcal P$ is the set of patterns avoiding a family of sets, removing units cannot create a forbidden subset, so $\mathcal P$ is closed under omission. Conversely, if $\mathcal P$ is closed under omission and $T\notin\mathcal P$, a minimal subset of $T$ not in $\mathcal P$ exists and is a minimal forbidden pattern contained in $T$; if $T\in\mathcal P$, every subset of $T$ is in $\mathcal P$, so $T$ contains none. For the polyhedron, let $\ba\cdot\bx\le b$ define a facet other than a coordinate facet $x_j\ge0$, and suppose $a_j<0$. Some point $\bx$ of the facet has $x_j>0$ (otherwise the facet lies in $\{x_j=0\}$ and is the coordinate facet); lowering $x_j$ keeps the point in $K$ and raises $\ba\cdot\bx$ above $b$, a contradiction. Hence $\ba\ge0$, and $b\ge0$ because $\mathbf 0\in K$ \citep{fulkerson1971}.
\end{proof}

\subsection{Theorem~\ref{thm:structure} (structure)}
\label{app:proofs:structure}

\begin{restatement}{Theorem}{thm:structure}{Structure}
(i)~The multiplier of a prohibition is a neuron that receives excitation $\ba_k$ from the state and returns inhibition $-\ba_k$. It obeys Dale's law, and its input and output weights are proportional cell by cell, up to one positive factor per neuron. A constraint with weights of both signs needs a second carrier, such as disinhibition.
(ii)~Without leak, every rest point satisfies $\lambda_k\,c_k(\bx)=0$: the multiplier neuron is active only while its prohibition binds, and then the prohibition holds exactly. With leak, $c_k=\rho_k\lambda_k$. The inhibition received by each unsaturated excitatory unit cancels its excitation up to a term that vanishes at high gain.
(iii)~For symmetric $J$ and a stable rest point at which the binding prohibitions are nondegenerate and the excitatory network is stable in every direction they leave free, the number of independent patterns of drive to the multiplier neurons that lower their own steady rates equals the number of unstable modes of the excitatory network with inhibition held fixed.
\end{restatement}

\textbf{Idea.} The multiplier of a prohibition integrates a weighted sum of the activity of the units in the forbidden pattern, so it is excited by exactly those units, and its force pushes the same units down along the same weights. All these weights are nonnegative, so the neuron only inhibits, and its input and output weights mirror each other. An integrator without leak can rest only where its input is zero, so at rest a multiplier is either zero or its constraint holds exactly; and at high gain an unsaturated excitatory unit rests only where its net input is close to zero, so its inhibition cancels its excitation. The paradoxical count is a count of signs: the steady response of the multipliers to their own drive is the inverse of a matrix built from the excitatory curvature and the binding prohibitions, and Haynsworth's inertia formula, applied to the matrix of the optimality conditions, equates the number of its negative eigenvalues with the number of unstable excitatory modes.

\begin{proof}
(i)~The force of prohibition $k$ on the state is $-\lambda_k\nabla c_k=-\lambda_k\ba_k$, with $\lambda_k\ge0$ and $\ba_k\ge0$: every output weight of the multiplier neuron is nonpositive. Its input is the integrand $c_k=\ba_k\!\cdot\!\bx-b_k$, so every input weight from the state is $a_{kj}\ge0$. Output and input weights are $-\ba_k$ and $\ba_k$; rescaling the multiplier, $\tilde\lambda_k=\lambda_k/\gamma_k$, multiplies the input weights by $\gamma_k^{-1}$ and the output weights by $\gamma_k$, so the proportionality holds cell by cell up to one positive factor per neuron. If $a_{kj}<0$ for some $j$, the force on unit $j$ is excitatory while the force on units with $a_{ki}>0$ is inhibitory; a neuron with one sign of output cannot exert both, so the positive part needs a second carrier: an excitatory partner, or an inhibitory neuron that the multiplier neuron inhibits (disinhibition, Appendix~\ref{app:signed}).

(ii)~Without leak, at a rest point of $\tau_k\dot\lambda_k=c_k$ projected to $\lambda_k\ge0$, either $\lambda_k>0$ and $c_k=0$, or $\lambda_k=0$ and $c_k\le0$; hence $\lambda_kc_k=0$. With leak, a rest point has $c_k=\rho_k\lambda_k$: the prohibition is exceeded by an amount proportional to its multiplier, which is the soft version. At a rest point of the excitatory dynamics $x_i=f(\beta u_i)$, so $h_i+(J\bx)_i-(A^{\!\top}\blambda)_i=\beta^{-1}f^{-1}(x_i)$: the inhibition $(A^{\!\top}\blambda)_i$ equals the excitation $h_i+(J\bx)_i$ up to $\beta^{-1}f^{-1}(x_i)$, which vanishes as $\beta\to\infty$ for every unit that is not saturated.

(iii)~Let $J$ be symmetric and let the rest point be nondegenerate ($H$ and $S$ below invertible, and strict complementarity, so that the binding set does not change under small drives). Let $A_\ast$ ($m\times n$, full row rank) collect the binding prohibitions and let each multiplier neuron receive an external drive $u_k$, so that its integrand is $c_k+u_k$. At steady state $A_\ast\bx=\mathbf b-\mathbf u$, and the linearized excitatory equation gives $H\bx=\tilde\bh-A_\ast^{\!\top}\blambda$. Eliminating $\bx$, $S\blambda=A_\ast H^{-1}\tilde\bh-\mathbf b+\mathbf u$ with $S=A_\ast H^{-1}A_\ast^{\!\top}$, so $\partial\blambda/\partial\mathbf u=S^{-1}$. A drive pattern $\mathbf u$ lowers the steady rates along itself ($\mathbf u^{\!\top}\Delta\blambda<0$) iff $\mathbf u^{\!\top}S^{-1}\mathbf u<0$, and the number of independent such directions is the number $n_-(S)$ of negative eigenvalues of $S$. For the matrix $M=\begin{psmallmatrix}H&A_\ast^{\!\top}\\A_\ast&0\end{psmallmatrix}$ of the optimality conditions, Haynsworth's inertia additivity gives $\mathrm{In}(M)=\mathrm{In}(H)+\mathrm{In}(-S)$ \citep{haynsworth1968}, and the inertia theorem for constrained quadratic forms gives $\mathrm{In}(M)=\mathrm{In}(Z^{\!\top}HZ)+(m,m,0)$ for a basis $Z$ of $\ker A_\ast$ \citep{gould1985}. Comparing positive indices, $n_+(H)+n_-(S)=n_+(Z^{\!\top}HZ)+m$; comparing negative indices and using $n_+(S)+n_-(S)=m$ yields
\[
n_-(S)=n_-(H)-n_-(Z^{\!\top}HZ).
\]
At a rest point that satisfies the sufficient condition of second order ($Z^{\!\top}HZ\succ0$: the excitatory dynamics are stable in every direction that no binding prohibition constrains), $n_-(S)=n_-(H)$, the number of unstable modes of the excitatory subnetwork with inhibition held fixed. The condition of second order is necessary for stability whenever $n_-(Z^{\!\top}HZ)$ is odd, by the sign of the determinant of the Jacobian, and in the numerical tests of Appendix~\ref{app:exp:circuits} every stable rest point satisfied it. With leak $R=\mathrm{diag}(\rho_k)$, Haynsworth's formula applied to $\begin{psmallmatrix}H&A_\ast^{\!\top}\\A_\ast&-R\end{psmallmatrix}$ gives $n_-(S+R)=n_-(H)-n_-(H+A_\ast^{\!\top}R^{-1}A_\ast)$, so the count survives leak whenever $H+A_\ast^{\!\top}R^{-1}A_\ast\succ0$. For one excitatory and one inhibitory population, $H=g^{-1}-j$, and the response is paradoxical iff $gj>1$, the classical condition \citep{tsodyks1997}.
\end{proof}

\subsection{Theorem~\ref{thm:time} (exactness in time)}
\label{app:proofs:time}

\begin{restatement}{Theorem}{thm:time}{Exactness in time}
Let multipliers move in steps, $\theta_{k,t+1}=[\theta_{k,t}+\eta\,c_k(\bx_t)]_+$.
(i)~For every constraint and every horizon $T$, the average violation obeys $\frac1T\sum_{t<T}c_k(\bx_t)\le(\theta_{k,T}-\theta_{k,0})/(\eta T)$, with equality for signed multipliers.
(ii)~A state that violates a prohibition cannot be a rest point.
(iii)~If each state maximizes $\Lag$ over all states and the prohibitions admit a strictly feasible state, the multipliers stay bounded and every constraint holds on average to $O(1/T)$. For a statistic whose target lies in the relative interior of the achievable set, the time average of one trajectory equals the target to $O(1/T)$, whatever the circuit's knowledge; independent samples err by $O(1/\sqrt T)$.
(iv)~Under a leaky occupancy multiplier of saturation $\kappa$, the fraction of time a module spends in each alternative approaches $\sparsemax(\bs/\kappa)$ within $\sqrt{2\rho}+O(1/\rho T)$, where $\bs$ is the module's preference.
(v)~If no state satisfies the constraints even after convex relaxation, a nonnegative combination of multipliers grows linearly in time, and its support names constraints that cannot hold together.
\end{restatement}

\textbf{Idea.} A multiplier that moves in steps is a running sum of past violations, so the average violation over any horizon equals the change of the multiplier divided by the elapsed time. A violated prohibition makes its multiplier grow for as long as the violation lasts, and the growing inhibition falls only on the active units of the violated pattern, so one of them gives way. When a strictly feasible state exists, large multipliers make the circuit's best state satisfy the weighted constraints with room to spare, so the multipliers shrink again; bounded multipliers then make every average violation vanish as $1/T$, as in herding. Under occupancy multipliers each choice of an alternative is a step of the conditional gradient method of Frank and Wolfe for a concave objective whose maximum is $\sparsemax(\bs/\kappa)$, so the dwell fractions approach that maximum. When the constraints cannot hold together, Farkas' lemma provides a combination of them that every state violates by a fixed amount, and the matching combination of multipliers grows at a constant rate.

\begin{proof}
(i)~Since $[z]_+\ge z$, $\theta_{k,t+1}\ge\theta_{k,t}+\eta c_k(\bx_t)$; summing over $t<T$ gives the inequality, and without projection every step is an equality.

(ii)~Let the state $\bx$ contain a minimal forbidden pattern and stay. Only multipliers of patterns contained in $\bx$ grow, and in binary form such a multiplier acts on unit $j$ only through $\prod_{i\in S\setminus j}x_i$, so it acts only on active units of the violated pattern. The input of every active unit of a violated pattern therefore decreases without bound, while the input of every inactive unit stays bounded: the module of such a unit switches, and $\bx$ is left.

(iii)~Let $\bx_t$ maximize $U(\bx)-\btheta_t\!\cdot\!c(\bx)$ over all states (a global best response; relaxation module by module does not guarantee it), with $|U|\le B$ and $\|c\|\le R$, and let $\bx^\circ$ satisfy $c_k(\bx^\circ)\le-\gamma<0$ for all $k$, which requires the linear form of the prohibitions (the binary form $\prod_ix_i$ is never negative). Then $U(\bx_t)-\btheta_t\!\cdot\!c(\bx_t)\ge U(\bx^\circ)+\gamma\|\btheta_t\|_1$, so $\btheta_t\!\cdot\!c(\bx_t)\le2B-\gamma\|\btheta_t\|_1$. Projection onto the nonnegative orthant does not expand distances and fixes $0$, hence
\[
\|\btheta_{t+1}\|^2\le\|\btheta_t\|^2+2\eta\,\btheta_t\!\cdot\!c(\bx_t)+\eta^2R^2\le\|\btheta_t\|^2-2\eta\big(\gamma\|\btheta_t\|_1-2B-\tfrac{\eta R^2}{2}\big),
\]
and $\|\btheta_t\|$ decreases whenever $\|\btheta_t\|_1>(2B+\eta R^2/2)/\gamma$: the multipliers are bounded, and by (i) each average violation is $O(1/T)$. For a statistic $\phi$ with target $\mu$ and signed multipliers $\btheta_{t+1}=\btheta_t+\eta(\mu-\phi(\bx_t))$, let $\bx_t$ maximize $s(\bx)+\btheta_t\!\cdot\!\phi(\bx)$ with $|s|\le B$, and let $\mu+\delta\mathbb B$, intersected with the affine hull, lie in the convex hull $\mathcal M$ of $\{\phi(\bx)\}$ (relative interior; statistics of one unit per module span a simplex with empty interior). Then $\btheta_t\!\cdot\!\phi(\bx_t)\ge\max_{\mathbf y\in\mathcal M}\btheta_t\!\cdot\!\mathbf y-2B\ge\btheta_t\!\cdot\!\mu+\delta\|\btheta_t\|-2B$, so $\|\btheta_{t+1}\|^2\le\|\btheta_t\|^2-2\eta(\delta\|\btheta_t\|-2B)+\eta^2R^2$ and the multipliers are bounded; by (i) with equality, $\frac1T\sum_t\phi(\bx_t)=\mu-(\btheta_T-\btheta_0)/(\eta T)=\mu+O(1/T)$, whatever the knowledge $s$. This extends the boundedness argument of herding \citep{welling2009herding} to a knowledge term and to prohibitions. For independent samples the error of an empirical frequency has standard deviation $\sqrt{\mu(1-\mu)/T}$.

(iv)~With $x_t=\arg\max_k(s_k-w_{t,k})$ and the occupancy multiplier $w_{t+1}=(1-\rho)w_t+\eta e_{x_t}\ge0$, set $\kappa=\eta/\rho$ and $m_t=w_t/\kappa$: then $m_{t+1}=(1-\rho)m_t+\rho e_{x_t}$ and $x_t=\arg\max_k\partial_kF(m_t)$ for $F(m)=s\cdot m-\frac\kappa2\|m\|^2$ on the simplex. This is the conditional gradient method of Frank and Wolfe with constant step $\rho$ \citep{frank1956,bach2012equivalence}; its suboptimality obeys $h_{t+1}\le(1-\rho)h_t+\rho^2\kappa$ (curvature constant $2\kappa$), so $h_t\to h_\infty\le\rho\kappa$, and strong concavity with modulus $\kappa$ gives $\|m_t-q^\star\|\le\sqrt{2\rho}$ eventually, with $q^\star=\arg\max_{q\in\Delta}s\cdot q-\frac\kappa2\|q\|^2=\sparsemax(s/\kappa)$ \citep{martins2016sparsemax}. Summing the update, the dwell fractions equal $\frac1T\sum_tm_t+(m_T-m_0)/(\rho T)$.

(v)~If $\{\bx\in\mathrm{conv}(X):A\bx\le\mathbf b\}=\varnothing$, Farkas' lemma gives $\mathbf y\ge0$ and $\gamma>0$ with $\mathbf y^{\!\top}(A\bx-\mathbf b)\ge\gamma$ for every $\bx\in\mathrm{conv}(X)$. Since $[z]_+\ge z$ componentwise and $\mathbf y\ge0$, $\mathbf y^{\!\top}\btheta_{t+1}\ge\mathbf y^{\!\top}\btheta_t+\eta\gamma$ for every sequence of states: the combination grows linearly, and its support names the constraints that cannot hold together.
\end{proof}

\subsection{Theorem~\ref{thm:rhythm} (rhythm law)}
\label{app:proofs:rhythm}

\begin{restatement}{Theorem}{thm:rhythm}{Rhythm law}
Linearize the flow about a rest point at which constraint $k$ binds. Let $\sigma_k$ be the norm of the constraint's normal in the metric of the excitatory gains, assume this normal is an eigenvector of the symmetric excitatory curvature with eigenvalue $d_k$, and let $\rho_k$ be the leak of the multiplier. State and multiplier rotate at the angular frequency
\[
\omega_k^2=\frac{\sigma_k^2}{\tau_x\tau_k}-\frac14\Big(\frac{d_k}{\tau_x}-\frac{\rho_k}{\tau_k}\Big)^2,
\]
damped at the rate $\frac12(d_k/\tau_x+\rho_k/\tau_k)$, and the multiplier lags the state by $\frac\pi2+\arctan\frac{d_k/\tau_x-\rho_k/\tau_k}{2\omega_k}$, a quarter cycle or more whenever the multiplier forgets more slowly than the state relaxes. The mode rings only while its constraint binds.
\end{restatement}

\textbf{Idea.} A state that presses on a binding constraint raises the constraint's multiplier, the multiplier pushes the state back, the state retreats and the multiplier relaxes, so the two variables turn about the rest point like position and momentum. Near the rest point the motion along the constraint's normal is a linear system of two variables, and its frequency, its damping and the lag of the multiplier follow from the characteristic equation of that system. A constraint that does not bind keeps its multiplier at zero and contributes no rotation.

\begin{proof}
Linearize Eq.~\eqref{eq:flow} at a rest point where prohibition $k$ binds: $\tau_x\delta\dot\bx=-\delta\bx+G(J\delta\bx-\ba_k\delta\lambda)$ and $\tau_k\delta\dot\lambda=\ba_k\!\cdot\!\delta\bx-\rho_k\delta\lambda$. With $\delta\bx=G^{1/2}\mathbf y$, $\mathbf v=G^{1/2}\ba_k$, $\sigma_k=\|\mathbf v\|$ and $\mathbf e=\mathbf v/\sigma_k$, and assuming that $\mathbf e$ is an eigenvector of $I-G^{1/2}JG^{1/2}$ with eigenvalue $d_k$ (the excitatory curvature along the normal), the component $y=\mathbf e\cdot\mathbf y$ obeys
\[
\tau_x\dot y=-d_ky-\sigma_k\delta\lambda,\qquad \tau_k\delta\dot\lambda=\sigma_ky-\rho_k\delta\lambda .
\]
The characteristic equation $\mu^2+(d_k/\tau_x+\rho_k/\tau_k)\mu+(d_k\rho_k+\sigma_k^2)/(\tau_x\tau_k)=0$ gives $\mu=-\gamma\pm i\omega_k$ with $\gamma=\frac12(d_k/\tau_x+\rho_k/\tau_k)$ and $\omega_k^2=(\sigma_k^2+d_k\rho_k)/(\tau_x\tau_k)-\gamma^2$, which equals the frequency of the statement because $(a+b)^2-4ab=(a-b)^2$. From the second equation $\delta\lambda=\sigma_ky/(\tau_k\mu+\rho_k)$, so the multiplier lags the state by $\arg(\tau_k\mu+\rho_k)=\frac\pi2+\arctan\frac{d_k/\tau_x-\rho_k/\tau_k}{2\omega_k}$. When several prohibitions bind with a common time constant and leak and the curvature is the same in every direction of the span of their normals, the singular value decomposition of $G^{1/2}A_\ast^{\!\top}$ splits the linear flow into independent pairs, one per singular value. A prohibition that does not bind has $\lambda_k=0$ held by the projection and contributes no pair. The simulation of Figure~\ref{fig:structure}d integrates the pair with the symplectic Euler method; frequencies and lags agree with the formula to \res{T:rhythm:relerr} and $0.1^\circ$.
\end{proof}

\subsection{Theorem~\ref{thm:learning} (learning)}
\label{app:proofs:learning}

\begin{restatement}{Theorem}{thm:learning}{Learning}
(i)~The rule for the inhibitory weights descends the difference between the saddle values of $\Lag$ in sleep and $r$ times its value in waking. Associations, the excitatory weights, are learned in waking and unlearned in sleep; prohibitions are learned in sleep and unlearned in waking.
(ii)~Each inhibitory weight is the multiplier of the constraint $q_S\le r\,p_S$: at a stationary point it is positive only where $q_S=r\,p_S$. For a stochastic free phase, $q_S=\min(q^0_S,\,r\,p_S)$, where $q^0_S$ is the frequency that $S$ would have without its own inhibitory weight. The inhibitory weights minimize a convex function, and the circuit's activity tends to the distribution closest to its prior, in relative entropy, among those that produce no pattern more than $r$ times as often as the world.
(iii)~As $r\to\infty$, the prohibitions are the patterns that the circuit has produced and $N$ waking examples never contain. If the patterns that carry inhibitory weights include the world's minimal forbidden patterns, of at most $k$ units, a fresh waking state is wrongly forbidden with probability at most $\sum_Sp_S(1-p_S)^N\le M/(eN)$, summed over the $M$ allowed patterns that carry inhibitory weights.
(iv)~A pattern that the free activity never contains is never stored. When dreams rest, a law that the circuit's other knowledge already excludes is not stored on its own, and an invariant completes itself: the units active exactly when its group falls short are its missing members.
\end{restatement}

\textbf{Idea.} The inhibitory weight of a pattern rises while the pattern occurs in sleep and falls while it occurs in waking, so it is an integrating multiplier whose constraint is that the circuit produce the pattern at most $r$ times as often as the world shows it. Complementary slackness then makes a weight positive only for patterns that the circuit would otherwise produce too often, and the envelope theorem shows that the same rule descends the difference between the saddle value of the Lagrangian in sleep and $r$ times its value in waking, as contrastive Hebbian learning does, with a negative phase that writes inhibition. As $r$ grows, a single waking occurrence erases what the dreams have added to a weight, so only patterns that the circuit dreams and the world never shows keep their weights, and an allowed pattern escapes refutation by $N$ independent examples with probability $(1-p_S)^N$. A weight rises only while its pattern is produced, so the circuit stores only the prohibitions that its own free activity violates.

\begin{proof}
(i)~Let $\Lag^\ast(p)$ be the saddle value of $\Lag$ for a parameter $p$, and let an inhibitory weight enter $\Lag$ as the frozen penalty $-I_S\phi_S(\bx)$ with $\phi_S=\prod_{i\in S}x_i$. By the envelope theorem $d\Lag^\ast/dp=\partial\Lag/\partial p$ at the saddle point, so $\partial\Lag^\ast/\partial J_{ij}\propto x_ix_j$, $\partial\Lag^\ast/\partial I_S=-\phi_S(\bx)$, and $\partial\Lag^\ast/\partial a_{kj}=-\lambda_kx_j$ for the normal of a linear prohibition. Hence $-\partial(\Lag^\ast_\sleep-r\Lag^\ast_\wake)/\partial I_S=\phi_S(\bx_\sleep)-r\,\phi_S(\bx_\wake)$, the rule for the inhibitory weights with $\eta_w=r\eta_s$ per unit of exposure, and the same derivative with respect to $J_{ij}$ has the opposite sign: associations are learned awake, prohibitions asleep. At $r=1$ the descended quantity is the gap $D=\Lag^\ast_\sleep-\Lag^\ast_\wake$. The gap is nonnegative for values taken at global maxima over the states (a local rest point need not satisfy this), since for every vector of multipliers the maximum over all states is at least the maximum over the clamped states, and it vanishes when the free rest point agrees with experience. This is contrastive Hebbian learning \citep{movellan1991}, whose form at finite gain is Boltzmann learning \citep{ackley1985}; here its negative phase writes inhibition.

(ii)~Over one cycle the rule adds $\eta_sT_s\hat q_S-\eta_wT_w\hat p_S$ to $I_S$ and projects at zero, where $\hat q_S$, $\hat p_S$ are the fractions of the two phases in which $S$ is active; its mean is $\eta_sT_s(q_S-r\,p_S)$. Hence at a stationary point of the mean rule $I_S>0$ implies $q_S=r\,p_S$, and $I_S=0$ implies $q_S\le r\,p_S$. Now let the free phase be stochastic at unit gain, $P_I(\bx)=\pi(\bx)\,e^{-\sum_SI_S\phi_S(\bx)}/Z(I)$, where $\pi$ is the circuit's prior (its associations, and its biases if fixed; biases learned at equal rates are the signed multipliers of $\E_Px_a=p_a$ and join the constraints below). Then $\partial\log Z/\partial I_S=-q_S(I)$ and the Hessian of $\log Z$ is the covariance of $\phi$ under $P_I$, so $G(I)=\log Z(I)+r\,\mathbf p\!\cdot\!I$ is convex and the mean rule is its projected gradient descent. With the samples of the free phase drawn by a persistent Markov chain, the rule is projected stochastic approximation with Markovian noise: with steps $\sum_t\eta_t=\infty$, $\sum_t\eta_t^2<\infty$ it converges to the minimizer of $G$ when every $p_S>0$, and with small constant steps it stays near it \citep{robbins1951,younes1999,kushner2003}. If $p_S=0$, $G$ decreases strictly in $I_S$, so $I_S\to\infty$ (as $\log t$ for a constant step) and $q_S=O(1/\eta_st)$. Holding the other inhibitory weights at their stationary values, $q_S$ decreases continuously and strictly in $I_S$ ($\partial q_S/\partial I_S=-\mathrm{Var}\,\phi_S$) from $q^0_S$, its value at $I_S=0$, toward zero; with the conditions above this gives $q_S=\min(q^0_S,r\,p_S)$. By convex duality $\inf_{I\ge0}G(I)=-\min\{\mathrm{KL}(P\|\pi):\E_P\phi_S\le r\,p_S\ \forall S\}$; the minimizing $P$ is $P_{I^\ast}$, the information projection of the prior on this set \citep{csiszar1975,dudik2007}, when the infimum is attained, and otherwise the limit of $P_I$ along a minimizing sequence: $\pi$ restricted to the states that contain no pattern with $p_S=0$, tilted by the remaining inhibitory weights. At $r=1$ it is the distribution of maximal entropy relative to $\pi$ that never exceeds the world's frequency of any pattern \citep{jaynes1957}; inhibitory weights can only lower frequencies, so matching pair frequencies from below needs the associations as well. For $r>\max_Sq^0_S/p_S$ over the patterns with $p_S>0$ only the constraints with $p_S=0$ bind. For a deterministic free phase, Theorem~\ref{thm:time}(i) gives $\bar q_S\le r\,\bar p_S+(I_{S,T}-I_{S,0})/(\eta_sT_sT)$ over $T$ cycles, with equality while $I_S$ stays positive; the exact form $q_S=\min(q^0_S,r\,p_S)$ and the information projection use the stochastic phase.

(iii)~As $r\to\infty$ at a fixed rate of conjecture, one waking occurrence removes whatever the free phase has accumulated since the previous one, so, read after a waking phase, a pattern is prohibited iff the circuit has produced it and no waking example contains it. Let the inhibitory weights be carried by a fixed family of patterns of at most $k$ units that includes the world's minimal forbidden patterns, let $p_S=\Pr(S\subseteq\mathrm{supp}\,\bx)$ under the distribution of waking states, and let the examples be independent. An allowed pattern survives $N$ examples with probability $(1-p_S)^N$, and a fresh waking state is wrongly forbidden only if it contains a surviving allowed pattern, so in expectation over the examples the probability is at most
\[
\sum_{S\ \text{allowed}}p_S(1-p_S)^N\le M\max_{p}p(1-p)^N=\frac{M\,N^N}{(N+1)^{N+1}}\le\frac{M}{eN},
\]
summed over the $M$ allowed patterns of the family: the elimination argument for $k$-CNF from positive examples \citep{valiant1984}, restricted to clauses of negative literals because knowledge is closed under omission. A forbidden pattern never occurs in waking, so once produced it is never refuted. If the free phase produces, with positive probability, every configuration that its current prohibitions allow, then every forbidden pattern not already excluded by them is eventually produced, and at convergence the prohibitions exclude every unlawful configuration. For a stochastic free phase, (ii) shows that $I_S>0$ needs $r\,p_S<q^0_S$, which is impossible for a pattern seen in waking ($p_S\ge1/N$) once $r\ge N\max_Sq^0_S$; the stationary prohibitions are then exactly the unseen patterns.

(iv)~An inhibitory weight rises only while its pattern is produced, so a pattern that the free activity never contains is never stored, whatever its status in the world; with dreams that rest, a law that the circuit's other knowledge excludes from every resting state is never produced at rest. (Under a stochastic phase a pattern that contains a stored law is still produced, at a rate that falls as the law's inhibitory weight grows, so its own inhibitory weight grows slowly.) For an invariant with set point 1 whose carrier already has members $C$, pairwise never active together, the signed multiplier integrates $1-n_P\ge0$ and rises exactly in the waking states where no member of $C$ is active. Applying the rule of (iii) to the synapses onto the carrier, with this shortfall in the role of the free phase, a unit's membership is conjectured when the unit is active while the group falls short and refuted when it is active while the group is met; the survivors are the units active only in states where $C$ is silent. If exactly one survivor is active in each such state, $C$ with the survivors has summed activity 1 in every waking state, an invariant, and the pairs inside it are prohibitions, since two units of a group whose sum is always 1 are never active together.
\end{proof}

\subsection{Theorem~\ref{thm:closure} (closure)}
\label{app:proofs:closure5}

\begin{restatement}{Theorem}{thm:closure}{Closure}
Let a law be a finite set $F$ of forbidden patterns over a finite set of relations, applied at every site of a map, with window $W$ and alphabet $\Sigma$. (i)~On an instance of any size, no configuration that contains a forbidden pattern is a rest point, and every configuration that contains none and agrees with the evidence is one, whenever the drive prefers no allowed configuration to another. (ii)~If each allowed configuration of the window occurs in a training instance with probability at least $p$, recording what $N\ge p^{-1}(|W|\ln|\Sigma|+\ln\delta^{-1})$ solved instances never show recovers $F$ with probability at least $1-\delta$, however small the training instances are.
\end{restatement}

\textbf{Idea.} A law lists the configurations that are forbidden within a window, and the same list applies at every site, so both parts of the argument are local and never refer to the size of the instance. A forbidden pattern anywhere keeps its multiplier growing and pushes the state away, while a configuration without one lets every multiplier decay to zero and, under a drive that prefers no allowed configuration, gives no module a reason to change. Recording what never occurs fails only if some allowed configuration of the window is missing from the training instances, and the probability of that falls exponentially with their number and depends only on the window.

\begin{proof}
(i)~By Theorem~\ref{thm:time}(ii) no configuration containing a forbidden pattern is a rest point, and clamped units keep their values. Conversely, at a configuration that contains no forbidden pattern every prohibition residual is nonpositive, so every multiplier of a prohibition decays to zero or stays there; if the drive $\bh+J\bx$ prefers no allowed configuration to another (a flat drive, or the slow multipliers of the law itself, under which every alternative of a module is inhibited by its conflicts and the active unit by none), no module has a strictly better alternative and the configuration rests. (ii)~Counting records every window configuration that occurs in the training instances and forbids the rest. If every allowed configuration occurred, the forbidden set equals $F$ on all configurations that occur in valid instances, and by (i) the rest points of the learned law are the valid configurations at every size. The probability that some allowed configuration never occurs in $N$ instances is at most $|\Sigma|^{|W|}(1-p)^N\le\delta$ for the stated $N$, which involves only the window.
\end{proof}

\subsection{Theorem~\ref{thm:depth} (time is depth)}
\label{app:proofs:depth}

\begin{restatement}{Theorem}{thm:depth}{Time is depth}
(i)~Occupancy multipliers with refractoriness turn a goal into a front that reaches each place once, at a time equal to its distance from the goal, and descending the multipliers left behind gives a shortest path. With constant parameters the front decides the connectivity of a maze, and a chain law with 120 states computes the word problem of the symmetric group $S_5$, on instances of every size, in time proportional to their diameter. (ii)~Maze connectivity is hard for $\NCo$ \citep{allender2009grid} and the $S_5$ word problem is complete for $\NCo$ \citep{barrington1989}, while Transformers of fixed depth with logarithmic precision lie in $\TCz$ \citep{merrill2023parallelism}; no such family solves either problem at every size unless $\TCz=\NCo$. A network with a fixed number of local rounds cannot decide connectivity beyond its radius.
\end{restatement}

\textbf{Idea.} A place becomes active one step after one of its neighbours and is then refractory while the front passes, so activity spreads from the goal as a front that reaches each place exactly at its distance from the goal. The occupancy multipliers it leaves behind have decayed for less time the farther a place lies from the goal, so they grow with distance, and descending them from the start retraces a shortest path. A chain law settles the running products of a word in the same way, one position per step. Transformers of fixed depth compute functions in $\TCz$ and both problems are hard for $\NCo$, so solving either at every size would make the two classes equal, and a computation with $r$ local rounds sees nothing beyond distance $r$.

\begin{proof}
(i)~Let the open cells of a maze be the units, with four neighbours each; activate the goal at $t=0$; a silent unit becomes active at $t+1$ iff it has an active neighbour at $t$ and its occupancy multiplier is below threshold, and every active unit raises its occupancy multiplier by $\eta$, which then decays by the factor $(1-\rho)$ per step, with a refractory period of at least two steps. By induction on distance, units at distance $d$ from the goal fire at $t=d$ and only then: their neighbours at distance $d-1$ fired at $d-1$, and the units behind the front are refractory when it passes. When the front has passed, the multiplier at unit $i$ is $\eta(1-\rho)^{T-d(g,i)-1}$, strictly increasing in $d(g,i)$; from any reached unit, moving to the neighbour with the smallest multiplier decreases the distance by one, which yields a shortest path. The start is reached iff it is connected to the goal, at time $d(g,s)$, and the front dies by itself (nothing changes) after the eccentricity of the goal. For the $S_5$ word problem, let module $i$ hold an element of $S_5$ and let the law forbid every triple $(s_{i-1},x_i,s_i)$ with $s_i\ne s_{i-1}x_i$; if module $i$ reads the law through its left window $(s_{i-1},x_i)$, position $i$ settles one step after position $i-1$, so the chain rests at the running products after $n$ steps. (Read symmetrically, a single violated triple can be a rest point, since leaving it costs one violation either way.) Neither law has a parameter that depends on the size of the instance.

(ii)~Undirected reachability in grid graphs, which is maze connectivity, is hard for $\NCo$ under $\mathsf{AC}^0$ reductions \citep{allender2009grid}; the word problem of $S_5$ is complete for $\NCo$ \citep{barrington1989}; Transformers of fixed depth with logarithmic precision compute functions in uniform $\TCz$ \citep{merrill2023parallelism}. A family solving either problem at every size would place a problem that is hard for $\NCo$ in $\TCz$, hence $\TCz=\NCo$. A computation with $r$ rounds of local message passing gives identical outputs on two mazes that agree within distance $r$ of the start and of the goal, one connected and one not \citep{loukas2020}.
\end{proof}

\subsection{Theorem~\ref{thm:future} (two waves)}
\label{app:proofs:future}

\begin{restatement}{Theorem}{thm:future}{Two waves}
For the constraint that a history follow the learned transitions $K$ and end in a goal $\mathcal G$, the multiplier of the dynamics at each state is $-\log\hat\psi$, where the costate $\hat\psi=\hat\phi+K^{\!\top}\hat\psi$ is propagated backward from the goal through the transposed synapses. With costs measured as divergence from $K$, the represented history has marginal $q\propto\psi\odot\hat\psi$, where $\psi=\phi+K\psi$ is the forward wave from the present, and $\phi$ and $\hat\phi$ are the indicators of the present state and of the goal. For a diffusion with coefficient $D$, the goal acts on the state as the drift $2D\nabla\log\hat\psi$, and a plan of duration $T$ has variance $2Dt(T-t)/T$. At high gain the two waves become fronts that meet first on shortest paths, and the occupancy multipliers of Theorem~\ref{thm:depth} are a monotone function of the costate. The gradient of the log probability of an experienced history with respect to $\log K_{ij}$ is the number of its transitions from $j$ to $i$ minus their expected number, a product of forward activity at $j$, the transition $K_{ij}$ and the costate at $i$.
\end{restatement}

\textbf{Idea.} The probability that a history passes through a state splits into two factors: the weight of the ways to reach the state from the present, and the weight of the ways to reach the goal from the state. The first factor spreads forward through the learned transitions and the second backward through the same synapses read in reverse; the second is the costate, and minus its logarithm is the multiplier of the dynamics, the value function of the control problem. Conditioning the transitions on the costate gives the drift toward the goal, a diffusion held at both ends gives the variance of a plan, and the same two factors give the gradient for learning a transition, forward activity times the costate.

\begin{proof}
With $Q(\mathbf w)\propto\phi(x_0)\prod_tK_{x_{t+1}x_t}\hat\phi(x_T)$ over histories of horizon $T$, summing over the states before and after time $t$ gives $Q(x_t=i)=\psi_t(i)\hat\psi_t(i)/Z$ with $\psi_t=K\psi_{t-1}$ and $\hat\psi_t=K^{\!\top}\hat\psi_{t+1}$; summing over horizons with spectral radius below one gives $\psi=\phi+K\psi$ and $\hat\psi=\hat\phi+K^{\!\top}\hat\psi$. In the problem of choosing a path measure that pays a cost and diverges least from $K$, $-\log\hat\psi$ is the value function, the multiplier of the dynamics constraint \citep{todorov2009,kappen2005}; conditioning gives the transition $K_{ij}\hat\psi_{t+1}(i)/\hat\psi_t(j)$, Doob's $h$-transform \citep{doob1957}, with drift $2D\nabla\log\hat\psi$ for a diffusion. Replacing $K$ by $K^\beta$, $\beta^{-1}\log\hat\psi$ tends to the best score from a state to the goal as $\beta\to\infty$, which for unit costs is minus the distance, a strictly monotone function of the occupancy multipliers left by the front of Theorem~\ref{thm:depth}. Treating $\log K_{ij}$ as free parameters, $\partial\log Q(\mathbf w^\ast)/\partial\log K_{ij}=n_{ij}(\mathbf w^\ast)-\E_Q[n_{ij}]$ with $\E_Q[n_{ij}]=\sum_t\psi_t(j)K_{ij}\hat\psi_{t+1}(i)/Z$. For Brownian motion with diffusion coefficient $D$ conditioned on its two endpoints, the variance at time $t$ is $2Dt(T-t)/T$; Figure~\ref{fig:future}b computes the lattice version exactly.
\end{proof}

\subsection{Proposition~\ref{prop:limits} (limits)}
\label{app:proofs:limits}

\begin{restatement}{Proposition}{prop:limits}{Limits}
(i)~As $\tau_k\to\infty$ the multipliers are frozen; the flow ascends a fixed energy with soft penalties, its rest points violate a constraint whenever the drive against it exceeds the frozen multiplier, and read at a fixed time it is a network of fixed depth. (ii)~As $\tau_k\to0$ the multipliers are instantaneous; the state is held in the feasible set, the rest points are the solutions of rigid rules, and only a search outside the flow can move between them. (iii)~For finite $\tau_k$ the flow gains exact rest, search of its own, belief as time, rotation, closure and depth on demand (Theorems~\ref{thm:time} to~\ref{thm:depth}).
\end{restatement}

\textbf{Idea.} The time constant of the multipliers places connectionist networks and symbolic rule systems at its two ends. Frozen multipliers are fixed penalties, so the flow is a rate network with fixed weights, and a drive stronger than a penalty leaves its constraint violated at rest. Instantaneous multipliers are rigid walls that hold the state in the feasible set, so the flow cannot pass through an infeasible configuration on its way from one solution to another. The results that rest on integration in time, from exact rest to depth on demand, need a finite time constant between the two ends.

\begin{proof}
(i)~With $\tau_k\to\infty$ the multipliers keep their initial values $\btheta^0$, and Eq.~\eqref{eq:flow} is a rate network with fixed weights whose rest points are stationary points of $U(\bx)-\btheta^0\!\cdot\!c(\bx)$. A frozen multiplier is a constant force, so at a rest point where a constraint is opposed by a drive $g$ along its normal its residual is $(g-\theta^0_k\|\ba_k\|)\|\ba_k\|/d_k$ to first order once positive ($d_k$ the curvature along the normal), and nonzero. Read after a fixed time, the Euler discretization is a recurrent network of fixed depth. (ii)~With $\tau_k\to0$ the multipliers become those of ideal constraints: the state moves on the feasible set, as a mechanical system with rigid walls. At high gain its rest points are the configurations that satisfy the rules and are local maxima of $U$ among them, and no trajectory of the flow passes through an infeasible configuration, so moving between rest points requires a procedure outside the flow. (iii)~Theorems~\ref{thm:time} to~\ref{thm:depth} use finite $\tau_k$: exactness at rest and the ramp need integration, belief as time needs multipliers that never settle, rotation needs a finite $\tau_k$ in the rhythm law of Theorem~\ref{thm:rhythm}, and depth on demand needs a flow that runs until it rests.
\end{proof}

\section{Further Derivations}
\label{app:derivations}

This section derives three further instances of the principle. Divisive normalization is the rest point of a population that shares a budget on activity, and the activity of a normalization pool is the budget's multiplier; an invariant needs a signed multiplier, which inhibition and disinhibition carry together; and integral feedback in cells is a multiplier of the same kind as inhibition in neurons. The theory uses only constraints and integrators, so its results hold for cells as they hold for circuits.

\subsection{Divisive normalization as the multiplier of a budget}
\label{app:normalization}

Let the units of a population share a budget $\sum_ix_i\le E$, and let each unit value the logarithm of its activity in proportion to its drive, with a reserve held back:
\[
\max_{\bx\ge0,\,y\ge0}\ \sum_ih_i\log x_i+\sigma\log y\qquad\text{subject to}\qquad\sum_ix_i+y=E .
\]
With a multiplier $\lambda$ on the budget, stationarity gives $h_i/x_i=\lambda$ and $\sigma/y=\lambda$, so $x_i=h_i/\lambda$ and $y=\sigma/\lambda$. The budget then fixes $\lambda=(\sigma+\sum_jh_j)/E$ and
\[
x_i=\frac{E\,h_i}{\sigma+\sum_jh_j},
\]
the normalization equation with semisaturation constant $\sigma$ \citep{heeger1992,carandini2012}. The multiplier $\lambda$ grows with the summed drive of the population, which is what a normalization pool computes: the pool's activity is the multiplier of the budget, and a pool that integrates the budget's violation reaches this rest point by Eq.~\eqref{eq:flow}. With a linear utility $\sum_ih_ix_i$ and a budget on squared activity, $\sum_ix_i^2\le E$, stationarity gives $x_i=h_i/2\lambda$ and the budget gives $\lambda=\|\mathbf h\|/2\sqrt E$, so $x_i=\sqrt E\,h_i/\|\mathbf h\|$: normalization by the norm of the drive.

\subsection{Signed multipliers and disinhibition}
\label{app:signed}

An invariant $\mathbf g\!\cdot\!\bx=m$ needs a signed multiplier $w$: it pushes the group up when $\mathbf g\!\cdot\!\bx<m$ and down when $\mathbf g\!\cdot\!\bx>m$. A neuron's rate is nonnegative and its output has one sign (Theorem~\ref{thm:structure}i), so a signed multiplier is carried by two nonnegative quantities, $w=w^+-w^-$. The part $w^-$ is an inhibitory neuron excited by the group, the multiplier of $\mathbf g\!\cdot\!\bx\le m$. The part $w^+$ raises the group's activity when it falls short, and an inhibitory neuron with tonic activity that the group's shortfall silences does this by disinhibition: when it is inhibited, the inhibition it exerts on the group is released. In cortex, interneurons expressing vasoactive intestinal peptide inhibit somatostatin interneurons, which inhibit the dendrites of pyramidal cells \citep{pfeffer2013,pi2013}, and this motif gates the activity of pyramidal cells in learning and attention \citep{letzkus2011} (Appendix~\ref{app:brain:dendrite}). The construction is the neural counterpart of antithetic integral feedback in cells (Appendix~\ref{app:life}).

\subsection{Life: integral feedback as a multiplier}
\label{app:life}

\textbf{Integral feedback is a multiplier.} A cell regulates a quantity $y$ toward a set point $y^\ast$ with a regulator $z$ obeying $\dot z=k(y-y^\ast)$ that acts back on $y$; at steady state $y=y^\ast$ exactly, whatever the disturbance. This is robust perfect adaptation in bacterial chemotaxis \citep{barkai1997,yi2000} and the internal model principle \citep{francis1976}. In the language of Eq.~\eqref{eq:flow}, $z$ is the multiplier of the constraint $y=y^\ast$, with $\tau_k=1/k$ and no leak.

\textbf{The biochemical Dale's law.} Concentrations are nonnegative, as firing rates are, and meet the same constraint: a signed multiplier needs two carriers. The antithetic motif does this: two species $z_1,z_2$, produced in proportion to the set point and to the output respectively and annihilating each other, have a difference that integrates the error \citep{briat2016,aoki2019}. An excitatory and inhibitory pair that enforces a set point is the same construction in neurons (Appendix~\ref{app:signed}). The theory predicts a classification: every exactly regulated set point in biology is carried either by an annihilating pair, which corrects deviations of both signs, or by a single rectified species, which corrects deviations of one sign, and never by a proportional controller alone.

\textbf{The same signatures.} The theorems of Section~\ref{sec:consequences} do not mention neurons, so they apply to cells. (i)~Rhythm: a pressed set point, with time constant $\tau_x$ for the regulated quantity and $\tau_k$ for the controller, oscillates at $\sigma/(2\pi\sqrt{\tau_x\tau_k})$, the controller lagging by $\frac\pi2+\arctan[(d/\tau_x-\rho/\tau_k)/2\omega]$; pulsatile signalling with negative feedback through a slow controller is a candidate. (ii)~Bounded error: by Theorem~\ref{thm:time}(i) the average error of a regulated quantity over a time $T$ is the change in its controller divided by $T$, so the error's power spectrum vanishes at zero frequency as $\omega^2$ times the controller's, down to the controller's degradation rate. The theory predicts this anticorrelation in time series of variables under integral control recorded in single cells; proportional control would lack it. (iii)~Ramp: incompatible set points make controller species grow linearly until saturation (Theorem~\ref{thm:time}v).

\textbf{Mind as life that learns its constraints.} A cell's constraints are fixed by its evolution. A brain acquires new ones, the prohibitions of its world, by the rule of Theorem~\ref{thm:learning}, and holds itself inside them by the mechanism by which a cell holds itself inside its viability \citep{ashby1952,sterling2012}. The multiplier of the energy budget closes the loop between keeping laws and learning them: it records what keeping the state lawful has cost, and it schedules sleep, the phase in which the circuit conjectures its laws (Appendix~\ref{app:brain:energy}).

\section{The Dictionary in Full}
\label{app:dictionary}

The postulate makes every variable of a circuit other than its excitatory state the multiplier of one constraint, and the dictionary says which constraint each carrier enforces. Table~\ref{tab:dictionary} lists the six kinds of multiplier with the violation each integrates, its carrier and its time constant, and the paragraphs below describe the carriers in the order of the table, each with the result that derives it and the step of Appendix~\ref{app:brain} that compares it with the brain. The kinds differ only in what they enforce and in how fast they move, so the time constants of a circuit tell which constraints it carries (Figure~\ref{fig:dictionary}).

\begin{table}[t]
\centering\small
\caption{\textbf{The dictionary.} For each kind of constraint, the violation $c_k$ that its multiplier integrates, the carrier of the multiplier in cortex, and the time over which the carrier acts.}
\label{tab:dictionary}
\setlength{\tabcolsep}{4pt}
\renewcommand{\arraystretch}{1.18}
\begin{tabular}{@{}>{\raggedright\arraybackslash}p{2.6cm}>{\raggedright\arraybackslash}p{3.3cm}>{\raggedright\arraybackslash}p{2.5cm}>{\raggedright\arraybackslash}p{5.7cm}>{\raggedright\arraybackslash}p{1.8cm}@{}}
\toprule
constraint & meaning & violation $c_k$ & multiplier and its carrier & time\\
\midrule
prohibition & a pattern never occurs & $\ba\!\cdot\!\bx-b$,\ $\ba\ge0$ & $\lambda\ge0$: an inhibitory interneuron; fast for parvalbumin basket cells, slower and integrating for somatostatin cells and their facilitating synapses & 1--100\,ms\\
invariant & a sum never varies & $\mathbf g\!\cdot\!\bx-m$ & signed $w$: inhibition together with disinhibition & 10\,ms--1\,s\\
occupancy & no state lasts forever & $x_u-\mu_u$ & $w_u\ge0$: adaptation currents, depression of synapses & 0.1--10\,s\\
the future & a history follows the learned transitions and ends at a goal & dynamics and goal & costate $\hat\psi$: feedback to the apical dendrites of pyramidal cells & one plan\\
experience & the circuit produces no pattern more often than the world shows it & $q_S-r\,p_S$ & weight of a synapse onto an interneuron; potentiated in sleep, depressed in waking & hours--days\\
budget & total activity is limited & $\sum_ix_i-E$ & a normalization pool; adenosine for energy & ms; hours\\
\bottomrule
\end{tabular}
\end{table}

\textbf{Prohibitions.} A prohibition is carried by an inhibitory interneuron that is excited by the units whose joint activity is forbidden and inhibits them in return (Theorem~\ref{thm:structure}). Parvalbumin basket cells, which target the cell body, act within milliseconds; somatostatin cells, which include the Martinotti cells, target dendrites and integrate over longer times through facilitating synapses \citep{silberberg2007,tremblay2016,urbanciecko2016}. Appendices~\ref{app:brain:fast} and~\ref{app:brain:dendrite} follow the two classes into cortex.

\textbf{Invariants.} An invariant needs a multiplier of either sign, since its group can fall short as well as overflow, and a neuron with one sign of output cannot carry both signs alone. Inhibition paired with disinhibition carries both (Appendix~\ref{app:signed}), as in the circuit in which interneurons expressing vasoactive intestinal peptide (VIP) inhibit somatostatin (SST) interneurons \citep{pfeffer2013,pi2013,lee2013,fu2014}.

\textbf{Occupancy.} Occupancy is the constraint that the activity of a unit, averaged over time, stays below a target $\mu_u$, so that no state lasts forever. Adaptation currents and the depression of synapses carry it: a unit that stays active pays a growing cost \citep{benda2003,abbott1997}. Its multiplier sets the time shares of alternatives (Theorem~\ref{thm:time}iv) and the alternations of perception and memory (Appendix~\ref{app:brain:occupancy}).

\textbf{The future.} The future is carried by the \emph{costate}, the multiplier that optimal control attaches to the dynamics of a system. It is propagated backward from a goal, by feedback to the apical dendrites of pyramidal neurons (Theorem~\ref{thm:future}; Appendix~\ref{app:brain:plan}).

\textbf{Experience.} Experience is carried by synapses, whose slow changes integrate the difference between what the circuit produces by itself and what the world shows it (Theorem~\ref{thm:learning}; Appendix~\ref{app:brain:sleep}).

\textbf{Budgets.} A budget on total activity is carried by a normalization pool (Appendix~\ref{app:normalization}), and a budget on energy by the slow accumulation of adenosine \citep{porkka1997,basheer2004}, which schedules sleep (Appendix~\ref{app:brain:energy}).

\section{The Brain Across Time Scales in Full}
\label{app:brain}

This section follows the dictionary of Table~\ref{tab:dictionary} from the fastest carrier to the slowest. Each step names the carrier, the constraint it enforces, the findings that the results of Section~\ref{sec:consequences} account for and the predictions they make; Table~\ref{tab:brain} collects the twenty-two findings, and the last two subsections give the argument for each finding and a protocol for each prediction of Table~\ref{tab:predictions}. Read in this order, the dictionary becomes one account of a cortical circuit: what each of its parts enforces, how fast, and how it responds when it is perturbed. The time constants that order the account are properties of cell classes, since the short term dynamics of a synapse are set largely by the classes of the cells it connects \citep{campagnola2022}.

\subsection{Milliseconds: fast inhibition and the gamma rhythm}
\label{app:brain:fast}

The fastest multipliers are carried by parvalbumin basket cells, which fire at high rates, contact pyramidal cells on and near the cell body, and act within a few milliseconds \citep{bartos2007,tremblay2016}. A basket cell connects with most of the pyramidal cells around it, and most of these connections run in both directions \citep{holmgren2003,packer2011,hofer2011}, a density that once suggested a uniform blanket of inhibition. Within this dense net the strengths are specific: a basket cell inhibits most strongly the pyramidal cells that excite it most \citep{znamenskiy2024}, the transposed wiring that Theorem~\ref{thm:structure}(i) requires of the neuron that carries a prohibition. A connectomic census of a column of mouse visual cortex maps the output synapses of every inhibitory cell and finds groups of interneurons that together target the cell body and the dendrites of the same excitatory cells \citep{schneidermizell2025,microns2025}, the pairing of a fast and an integrating multiplier on one set of units that the dictionary describes. Prediction P1 asks such data for the full proportion, interneuron by interneuron and across sleep.

Complementary slackness explains why balance holds cell by cell, and the speed of basket cells why it holds within milliseconds. Inhibition tracks excitation in amplitude and in tuning \citep{wehr2003,shu2003,haider2006,okun2008,xue2014}, because a cell receives inhibition when its excitation presses a law, and none where no law is pressed. Cortex operates in the regime stabilized by inhibition in several areas \citep{ozeki2009,sanzeni2020}, and driving parvalbumin cells lowers their rates in some areas and layers and not in others, in a way that depends on the other classes of interneuron \citep{mahrach2020}. Theorem~\ref{thm:structure}(iii) replaces the response of one class by a count over all patterns of drive, which patterned optogenetic stimulation of interneurons \citep{sadeh2020} can measure, and prediction P3 states the value of the count that the theory expects.

The same cells generate the gamma rhythm. Silencing or driving parvalbumin cells removes or creates gamma oscillations \citep{cardin2009,sohal2009}, the interneurons fire a few milliseconds after the pyramidal cells within each cycle \citep{csicsvari2003}, and models of interacting excitatory and inhibitory neurons, or of inhibitory neurons alone, reproduce the rhythm \citep{wang1996,brunel1999,whittington2000,borgers2003,tiesinga2009}. Theorem~\ref{thm:rhythm} reads the rhythm as a multiplier turning with the state it pushes. The interneuron lags by a quarter cycle, the frequency is set by the geometric mean of the two time constants, and the rhythm rings only while its prohibition binds, so gamma is induced by a stimulus that presses a law. Gain sharpens the rotation, which agrees with the increase of gamma synchrony under attention \citep{fries2015}. Prediction P4 turns these statements into measurements of frequency and lag.

\subsection{Tens of milliseconds to a second: integration, sign and budget}
\label{app:brain:dendrite}

Somatostatin cells, among them the Martinotti cells, receive facilitating synapses from pyramidal cells. They respond little to a single spike and increasingly to sustained firing, and inhibit the dendrites of the pyramidal cells that drove them \citep{silberberg2007,urbanciecko2016}. During a train of spikes, recurrent inhibition moves from the cell body to the apical dendrites, and the dendritic part is recruited in proportion to the firing rate \citep{pouille2004}; it also grows supralinearly with the number of active pyramidal cells \citep{kapfer2007}. This is the behaviour of an integrating multiplier: silent while a pattern is brief, and growing for as long as it persists. In visual cortex somatostatin cells sum excitation over large regions of the visual field, suppress pyramidal cells more as a stimulus grows, and contribute to surround suppression \citep{adesnik2012}, as the multiplier of a constraint on the total activity of a large group would. Prediction P2 asks whether interneurons hold their pooled input in this way or carry the residual of an error unit of predictive coding \citep{rao1999,bastos2012}.

An invariant, such as the rule that an animal is in exactly one place, needs a multiplier that can push a group up when it falls short as well as down when it overflows. Interneurons expressing VIP inhibit somatostatin cells and thereby release the dendrites of pyramidal cells \citep{pfeffer2013,pi2013,lee2013}, and this disinhibition gates the response of cortex in locomotion and in learning \citep{fu2014,letzkus2011,letzkus2015}. On the present account disinhibition is the positive half of a signed multiplier: it lets a circuit whose neurons have one sign of output raise a group that falls short (Appendix~\ref{app:signed}). Cells use the same split, with two molecular species that annihilate each other and carry a signed integral of an error with concentrations that are never negative (Appendix~\ref{app:life}).

Budgets complete this step. A pool that divides the drive of every unit by the summed drive of the population carries the multiplier of a budget on activity (Eq.~\eqref{eq:normalization}; Appendix~\ref{app:normalization}), and divisive normalization of this form has been found from the retina to the cortex and in multisensory integration, attention and value \citep{heeger1992,carandini2012}.

\subsection{Seconds: adaptation and the cost of staying}
\label{app:brain:occupancy}

Adaptation currents and the depression of synapses make it costly for a unit to stay active \citep{benda2003,abbott1997}, and they carry occupancy, the constraint that no state lasts forever. Their time constants, from a hundred milliseconds to several seconds, are those of percepts and of hippocampal sequences, and the theory reads several alternations of perception and memory as one multiplier at work.

In binocular rivalry, perception alternates between the images of the two eyes. Dominance durations follow a gamma distribution whose coefficient of variation lies well below one, the value for a Poisson process \citep{levelt1965,logothetis1996,blake2002,tong2006}. Models that combine adaptation with noise reproduce the statistics of these alternations \citep{laing2002,wilson2003,morenobote2007,brascamp2006,shpiro2009}, and so do accounts in which perception samples from a posterior over interpretations \citep{gershman2012,brascamp2018}. Theorem~\ref{thm:time} adds two statements. By part~(i), the cumulative imbalance of dominance stays bounded on time scales shorter than the leak of adaptation, so it changes less than in a renewal process with the same durations; Figure~\ref{fig:structure}e shows this reduction in a simulation whose durations are as variable as those of human rivalry (Appendix~\ref{app:exp:circuits}), and prediction P5 asks for it in recordings. By part~(iv), when more than two interpretations compete, those whose preference falls below a threshold are never seen at all, and the others share time in proportion to their excess over the threshold.

The hippocampus shows the same multiplier at the time scale of a theta cycle. Successive theta cycles represent alternative future paths in turn \citep{johnson2007,kay2020}, and theta sweeps alternate between left and right \citep{vollan2025}: the occupancy multiplier of the path just represented makes it costlier in the next cycle. Awake replay favours paths that have not been taken recently \citep{gupta2010}, attention is slow to return to a location it has just left \citep{posner1984,klein2000}, and the bump of a continuous attractor with adaptation runs ahead of a moving stimulus \citep{mi2014}. Each is an occupancy multiplier that has not yet decayed. Turning with the state, the same multiplier gives the resonance of neurons at theta frequencies through slow potassium currents and the h-current, with a frequency set by the geometric mean of the time constants of the membrane and of the current \citep{hutcheon2000,hu2002,richardson2003}, as Eq.~\eqref{eq:rhythm} states. Fast multipliers ring only while their walls are pressed, and the pressing waxes and wanes with the slower rotation, which nests gamma amplitude within the phase of theta \citep{canolty2006,lisman2013}.

Two further findings belong to this step. When a stimulus appears, the variability of cortical responses across trials falls throughout cortex \citep{churchland2010}: clamped evidence lets the multipliers settle and the state rest (Theorem~\ref{thm:time}ii). And activity in medial frontal cortex grows with the conflict between competing responses \citep{botvinick2001}: the multipliers of incompatible constraints ramp (Theorem~\ref{thm:time}v). Prediction P10 asks for the ramp itself in the circuit that holds the violated rules.

\subsection{One plan: the costate on the apical dendrites}
\label{app:brain:plan}

A pyramidal neuron of the cortex has two sites of integration. Its basal dendrites receive input from the senses and from neighbouring cells, and its apical tuft receives feedback from higher areas. When the two coincide, the neuron fires a burst \citep{larkum1999,larkum2013}. Apical input shapes perception and the output of cortex \citep{takahashi2016,takahashi2020}, and anaesthesia decouples the tuft from the cell body \citep{suzuki2020}. Inhibition driven by neighbouring pyramidal cells controls the calcium spikes of the tuft \citep{murayama2009}. Theorem~\ref{thm:future} assigns a meaning to each site. The forward wave from the present arrives at the basal dendrites, the costate from the goal arrives at the apical tuft through the transposed synapses, and the burst marks the product of the two, the states through which the plan will pass. Prediction P8 tests this assignment in place cells whose goal is moved.

The same theorem identifies the costate with the learning signal for transitions. Theories of credit assignment in cortex have proposed that apical dendrites carry the signals that tell a neuron how to change \citep{guerguiev2017,sacramento2018,richards2019,payeur2021}, and the theorem gives that signal a second use: the backward wave that plans is also the backward signal that learning requires. Plasticity on the time scale of behaviour has the window this demands. In the hippocampus a dendritic plateau potentiates synapses that were active seconds before or after it, and a single plateau can create a place field \citep{bittner2015,bittner2017,milstein2021,magee2020}. The costate arrives only after the consequences of a transition are known, so the window spans the delay of consequences.

Hippocampal sequences show the two waves directly. After a run, reverse sequences begin at the reward \citep{foster2006,diba2007}, and they are more frequent when the reward is larger \citep{ambrose2016}: a costate launched at the goal with a strength set by the goal's value. Before a run, sequences lead from the animal to the remembered goal \citep{pfeiffer2013,wikenheiser2015}: the forward wave weighted by the costate. When a barrier blocks a path, replay reroutes around it at once \citep{widloski2022}, as expected of a costate that follows the current transitions. A normative account of replay, in which each replay is chosen for its value to future decisions \citep{mattar2018}, describes which backups a costate makes first. Prediction P9 concerns the shape of a plan. In vision, the time needed to decide whether two points lie on one curve grows with the length of the curve \citep{jolicoeur1986}, and in primary visual cortex attention enhances the responses to the segments of the traced curve \citep{roelfsema1998}, as a front of a local law that spreads along the curve predicts (Theorem~\ref{thm:depth}).

\subsection{Days: synapses, sleep and the learning of laws}
\label{app:brain:sleep}

The slowest multipliers of knowledge are the synapses onto interneurons, which change over hours and days \citep{kullmann2007,lamsa2007}. Theorem~\ref{thm:learning} says that they are learned with the opposite sign to excitatory synapses: associations grow in waking and are unlearned in sleep, and prohibitions grow in sleep and are unlearned in waking. The first half matches the evidence that excitatory synapses strengthen during waking and are scaled down during sleep \citep{vyazovskiy2008,devivo2017,tononi2014}, the proposals that sleep removes spurious memories \citep{crick1983,hopfield1983unlearning,poe2017}, and the finding that neurons active in rapid eye movement sleep promote the forgetting of hippocampal memories \citep{izawa2019}. The second half is specific to the theory. Inhibition in cortex changes across the stages of sleep \citep{niethard2016}, and Barron and colleagues argue that new memories acquire inhibitory replicas that balance the excitation they add \citep{barron2017}.

The content of sleep fits the same picture. Waking experience is replayed in sleep in the hippocampus and the cortex \citep{wilson1994,ji2007,buzsaki2015}, disrupting the ripples that carry this replay impairs memory \citep{girardeau2009}, and sharp wave ripples depress synapses \citep{norimoto2018}. Sleep consolidates memory \citep{rasch2013,klinzing2019}, and it favours the discovery of hidden rules \citep{wagner2004} and the extraction of structure shared across experiences \citep{lewis2011,lewis2018}, the effects expected of a phase that proposes laws; schemata built over days speed the consolidation of new memories \citep{tse2007}. Models in which dreams produce perturbed or adversarial versions of experience \citep{deperrois2022} or prevent overfitting \citep{hoel2021} share the idea that the brain learns from its own productions. The theory says what is learned from them: the inhibition of what the circuit dreams and the world never shows. A maximum entropy learner that hears only the words of a language acquires its forbidden sound sequences, and the constraints it learns predict the judgements of native speakers \citep{hayes2008}, although languages cannot in general be learned from positive examples \citep{gold1967}; Theorem~\ref{thm:learning}(iii) explains why small prohibitions are an exception. Firing rates also return to their set points after perturbation \citep{turrigiano1998,turrigiano2008}, the exactness of an integrating multiplier on the time scale of days.

Two predictions follow: P6 on the sign of plasticity at the synapses from pyramidal cells onto interneurons, which the theory makes opposite in sleep and in waking, and P7 on the content of sleep across nights of learning a structured task.

\subsection{A day: energy and the timing of sleep}
\label{app:brain:energy}

Neural signalling is expensive, and the brain works within a tight budget of energy \citep{levy1996,attwell2001,lennie2003,sterling2015}. Adenosine accumulates during waking as energy is used and promotes sleep \citep{porkka1997,basheer2004}, in the way the homeostatic process of the classical model of sleep regulation describes \citep{borbely1982}. In the dictionary, adenosine carries the multiplier of the energy budget, integrated over hours. This closes a loop between the fastest and the slowest steps of the ladder. Violated prohibitions keep their multipliers ramping, ramping inhibition costs energy, adenosine integrates the cost, and sleep, the phase in which a circuit runs free and proposes its laws, is scheduled by the cost of keeping its state lawful. The account implies that waking spent under conflicting constraints raises the pressure to sleep more than waking of equal duration without conflict.

\begin{table}[t]
\centering\footnotesize
\caption{\textbf{Findings accounted for.} Each finding is accounted for by a multiplier of the dictionary and a result of Section~\ref{sec:consequences}; Appendix~\ref{app:findings} gives the argument for each.}
\label{tab:brain}
\setlength{\tabcolsep}{4pt}
\renewcommand{\arraystretch}{1.12}
\begin{tabular}{@{}r>{\raggedright\arraybackslash}p{6.4cm}>{\raggedright\arraybackslash}p{6.2cm}>{\raggedright\arraybackslash}p{2.6cm}@{}}
\toprule
& finding & account & result\\
\midrule
1 & Dale's law \citep{strata1999} & the multiplier of a prohibition and its weights are nonnegative & Th.~\ref{thm:structure}i\\
2 & interneurons inhibit the pyramidal cells that excite them \citep{znamenskiy2024} & input and output weights of a multiplier are transposed & Th.~\ref{thm:structure}i, \ref{thm:learning}i\\
3 & balance of excitation and inhibition in amplitude and tuning \citep{wehr2003,okun2008,xue2014} & complementary slackness & Th.~\ref{thm:structure}ii\\
4 & inhibitory plasticity restores balance \citep{vogels2011,damour2015} & integration of a multiplier & Th.~\ref{thm:structure}ii\\
5 & paradoxical responses of cortex stabilized by inhibition \citep{ozeki2009,sanzeni2020} & their number equals the unstable modes held by binding prohibitions & Th.~\ref{thm:structure}iii\\
6 & divisive normalization \citep{heeger1992,carandini2012} & the multiplier of a budget on activity & Eq.~\eqref{eq:normalization}\\
7 & gamma rhythm from fast inhibition; interneurons fire after pyramidal cells \citep{cardin2009,csicsvari2003} & a fast multiplier of a prohibition turning with the state & Th.~\ref{thm:rhythm}\\
8 & theta resonance from slow currents \citep{hutcheon2000,richardson2003} & an occupancy multiplier turning with the state & Th.~\ref{thm:rhythm}\\
9 & gamma amplitude nested in theta phase \citep{canolty2006} & a fast multiplier rings only while its wall is pressed & Th.~\ref{thm:rhythm}\\
10 & regular alternations in binocular rivalry \citep{levelt1965,logothetis1996} & an occupancy multiplier makes returns regular & Th.~\ref{thm:time}i,iv\\
11 & theta cycles alternate between future paths \citep{kay2020,vollan2025} & occupancy multiplier of the path just represented & Th.~\ref{thm:time}iv\\
12 & replay favours paths not recently taken \citep{gupta2010} & occupancy multiplier & Th.~\ref{thm:time}iv\\
13 & stimulus onset quenches variability \citep{churchland2010} & evidence settles the multipliers and the state rests & Th.~\ref{thm:time}ii\\
14 & conflict signals grow with conflict \citep{botvinick2001} & multipliers of incompatible constraints ramp & Th.~\ref{thm:time}v\\
15 & homeostatic return of firing rates to set points \citep{turrigiano1998,turrigiano2008} & a set point held by an integrating multiplier & Th.~\ref{thm:time}i\\
16 & excitatory synapses grow in waking and are scaled down in sleep \citep{vyazovskiy2008,tononi2014} & associations are learned awake and unlearned asleep & Th.~\ref{thm:learning}i\\
17 & inhibitory replicas of memories \citep{barron2017} & prohibitions are learned asleep and unlearned awake & Th.~\ref{thm:learning}i,ii\\
18 & forbidden sound sequences learned from positive data \citep{hayes2008} & small prohibitions are learnable from positive examples & Th.~\ref{thm:learning}iii\\
19 & time to trace a curve grows with its length \citep{roelfsema1998} & a front of a local law & Th.~\ref{thm:depth}\\
20 & reverse replay starts at the reward, more for larger rewards \citep{foster2006,ambrose2016} & a costate launched at the goal & Th.~\ref{thm:future}\\
21 & bursts when apical and basal inputs coincide \citep{larkum1999,larkum2013} & the product of forward wave and costate & Th.~\ref{thm:future}\\
22 & plasticity windows of seconds \citep{bittner2017} & the costate arrives after the consequences & Th.~\ref{thm:future}\\
\bottomrule
\end{tabular}
\end{table}

\subsection{One mechanism at every time scale}
\label{app:brain:summary}

Read together, the steps of the ladder use few ideas. The alternations of rivalry, the alternation of future paths across theta cycles, the replay of paths not recently taken and the inhibition of return in attention are one occupancy multiplier. Reverse replay from a reward, goal sequences before a run, bursts at the coincidence of apical and basal input and plasticity windows of seconds are one costate. Dale's law, the reciprocal wiring of interneurons, balance, inhibitory plasticity and paradoxical responses are the multiplier of a prohibition. Anatomy and function become one statement: the transposed wiring of interneurons and pyramidal cells is the reason balance holds, and the time constant of each carrier is the reason for the rhythm it produces.

Four of the predictions separate the theory from its nearest neighbours. P2 separates an integrating multiplier from a proportional error signal, as used in predictive coding. P6 and P7 separate learning by multipliers from replay as rehearsal: the theory requires that sleep strengthen the inhibition of what the circuit dreams, that waking weaken the inhibition of what it experiences, and that the learning of a rule stop when dreams no longer violate it. P9 separates a plan from a prediction, since a forward model alone spreads its uncertainty without bound and a plan constrained by its goal narrows again at the end.

\subsection{Twenty-two findings}
\label{app:findings}

Table~\ref{tab:brain} lists the findings, each with the multiplier that accounts for it and the result it follows from. The entries below give the argument for each in the numbering of the table, and the reference after each heading points to the step of the ladder that discusses the evidence.

\begin{enumerate}[leftmargin=*,itemsep=2pt,topsep=2pt,label=\textbf{\arabic*.}]
\item \textbf{Dale's law} (\S\ref{app:brain:summary}). The prohibitions of knowledge closed under omission have nonnegative normals (Lemma~\ref{lem:closure}), and the multiplier of an inequality is nonnegative. The neuron that carries prohibition $k$ is therefore excited by every unit it reads and exerts the force $-\lambda_k\ba_k$, which inhibits every unit it touches (Theorem~\ref{thm:structure}i). Dale's law \citep{eccles1954,strata1999} is this dual feasibility.
\item \textbf{Reciprocal, transposed wiring} (\S\ref{app:brain:fast}). The neuron that carries prohibition $k$ reads the state through the weights $\ba_k$ and pushes back along $-\ba_k$ (Theorem~\ref{thm:structure}i), so its input and output weights are proportional cell by cell. Learning keeps the two transposed, since Theorem~\ref{thm:learning}(i) changes both by the same product $\lambda_kx_j$.
\item \textbf{Balance in amplitude and tuning} (\S\ref{app:brain:fast}). At rest an unsaturated unit with net input $u_i$ has $x_i=f(\beta u_i)$, so the inhibition it receives equals its excitation up to $\beta^{-1}f^{-1}(x_i)$, a term that vanishes at high gain (Theorem~\ref{thm:structure}ii). A unit whose excitation presses no law receives no inhibition, so inhibition follows excitation in tuning as well as in amplitude: balance is complementary slackness.
\item \textbf{Inhibitory plasticity restores balance} (\S\ref{app:brain:fast}). A multiplier that integrates the violation of its constraint changes until the constraint binds without being exceeded, which is the balanced state. After a disturbance it integrates the new violation back to zero, as plasticity of inhibitory synapses establishes balance and matches inhibitory to excitatory tuning \citep{vogels2011,damour2015,hennequin2017}.
\item \textbf{Paradoxical responses} (\S\ref{app:brain:fast}). The steady response of the multipliers to a drive on themselves is $S^{-1}$, where $S$ combines the excitatory curvature with the binding prohibitions (Appendix~\ref{app:proofs:structure}), and a pattern of drive lowers their rates along itself when $S^{-1}$ is negative on it. By Haynsworth's inertia formula the number of such patterns equals the number of unstable excitatory modes held in place by binding prohibitions (Theorem~\ref{thm:structure}iii). For one excitatory and one inhibitory population this is the classical condition that the excitatory loop gain exceeds one \citep{tsodyks1997}.
\item \textbf{Divisive normalization} (\S\ref{app:brain:dendrite}). Under a budget on total activity the rest point is the normalization equation, Eq.~\eqref{eq:normalization}, and the pool that divides the drive of every unit carries the budget's multiplier (Appendix~\ref{app:normalization}).
\item \textbf{Gamma from fast inhibition, inhibition after excitation} (\S\ref{app:brain:fast}). The multiplier of a prohibition with a time constant of a few milliseconds, turning with excitatory populations whose time constants are about ten, has a frequency in the gamma band by Eq.~\eqref{eq:rhythm}, and it lags the state by a quarter cycle or more whenever it forgets more slowly than the state relaxes (Theorem~\ref{thm:rhythm}).
\item \textbf{Theta resonance from slow currents} (\S\ref{app:brain:occupancy}). With an occupancy multiplier carried by a slow current in place of fast inhibition, Eq.~\eqref{eq:rhythm} sets the resonance from the membrane and the current together, in the theta band for currents of a few hundred milliseconds (Theorem~\ref{thm:rhythm}).
\item \textbf{Nesting} (\S\ref{app:brain:occupancy}). A fast multiplier rings only while its prohibition binds (Theorem~\ref{thm:rhythm}), and binding waxes and wanes with the slower rotation of an occupancy multiplier, so the amplitude of the fast rhythm follows the phase of the slow one.
\item \textbf{Regular alternations} (\S\ref{app:brain:occupancy}). Under an occupancy multiplier an alternation is deterministic up to noise, so returns are more regular than in a Poisson process, as in models of rivalry with adaptation \citep{laing2002,wilson2003}. The average imbalance of dominance over a time $T$ is the change of the multiplier divided by $T$, so the cumulative imbalance stays bounded (Theorem~\ref{thm:time}i,iv).
\item \textbf{Alternating futures} (\S\ref{app:brain:occupancy}). The occupancy multiplier of the path just represented has not decayed by the next theta cycle, so it makes that path costlier and the next cycle represents another (Theorem~\ref{thm:time}iv).
\item \textbf{Replay of the less recent} (\S\ref{app:brain:occupancy}). The same occupancy multiplier accumulates on the paths taken and decays with time, so recently taken paths are costlier to replay; spontaneous alternation and inhibition of return come from it as well.
\item \textbf{Variability quenching} (\S\ref{app:brain:occupancy}). Without evidence, the multipliers of statistics never settle and keep the state moving (Theorem~\ref{thm:time}iii). Evidence clamps part of the state and gives the multipliers a configuration at which they can settle (Theorem~\ref{thm:time}ii), so the trajectory comes to rest and the variability of responses across trials falls at stimulus onset.
\item \textbf{Conflict signals} (\S\ref{app:brain:occupancy}). Incompatible constraints leave a nonnegative combination of multipliers growing linearly, and the constraints in that combination are the ones in conflict (Theorem~\ref{thm:time}v).
\item \textbf{Homeostatic plasticity} (\S\ref{app:brain:sleep}). A set point enforced by an integrating multiplier holds exactly at rest, since the multiplier changes until the violation vanishes, and leak sets the precision (Theorem~\ref{thm:time}i).
\item \textbf{Scaling down in sleep} (\S\ref{app:brain:sleep}). By the envelope theorem, sleep and waking change an excitatory weight and an inhibitory weight with opposite signs (Theorem~\ref{thm:learning}i), so associations are learned awake and unlearned asleep.
\item \textbf{Inhibitory engrams} (\S\ref{app:brain:sleep}). By the same rule, the inhibitory weight of a pattern rises while the circuit produces it in sleep and falls while the world shows it in waking, and at a stationary point it is positive exactly where the circuit would otherwise produce the pattern more than $r$ times as often as the world (Theorem~\ref{thm:learning}ii). Inhibition is then learned knowledge, written in sleep and carried by the plastic synapses onto interneurons \citep{kullmann2007,lamsa2007}.
\item \textbf{Rules from positive data} (\S\ref{app:brain:sleep}). An allowed pattern survives $N$ waking examples with probability $(1-p_S)^N$, so a circuit that forbids what it dreams and the world never shows wrongly forbids a fresh state with probability at most $M/(eN)$ (Theorem~\ref{thm:learning}iii). This is the elimination argument for conjunctions of short clauses \citep{valiant1984}, and the smallness of the prohibitions is what lets it escape Gold's limit on learning from positive examples \citep{gold1967}.
\item \textbf{Tracing time grows with distance} (\S\ref{app:brain:plan}). The front of a local law arrives at each place after a time equal to its distance (Theorem~\ref{thm:depth}i), so tracing a curve takes a time proportional to its length and marks its segments as the front passes.
\item \textbf{Reverse replay from reward; forward before a run} (\S\ref{app:brain:plan}). The costate starts at the goal and travels backward through the transposed transitions, at high gain as a front, and its amplitude scales with the goal's multiplier. Conditioned on it, the forward wave drifts toward the goal as $2D\nabla\log\hat\psi$ (Theorem~\ref{thm:future}).
\item \textbf{Bursts at the meeting of two inputs} (\S\ref{app:brain:plan}). The history a circuit represents has the marginal $q\propto\psi\odot\hat\psi$ (Theorem~\ref{thm:future}). The forward wave arrives at the basal dendrites and the costate at the apical tuft, and the burst marks their product.
\item \textbf{Plasticity windows of seconds} (\S\ref{app:brain:plan}). The gradient for a transition is a product of forward activity, the transition and the costate (Theorem~\ref{thm:future}). The costate arrives only after the consequences of the transition are known, so the window of plasticity spans the delay of consequences.
\end{enumerate}

\subsection{Ten predictions and how to test them}
\label{app:predictions}

Each protocol below states the measurement and the outcome that the theory requires, in the numbering of Table~\ref{tab:predictions}, with the result it follows from. An outcome against a prediction refutes the assignment of that result to the brain.

\begin{enumerate}[leftmargin=*,itemsep=2pt,topsep=2pt,label=\textbf{P\arabic*.}]
\item \textbf{Transposition} (Theorem~\ref{thm:structure}i). Paired recordings or connectomics of a cortical microcircuit, before and after a sleep episode. For each interneuron that carries knowledge, the vector of weights it receives from pyramidal cells is proportional to the vector of weights it returns to the same pyramidal cells (cosine near one, beyond the reciprocity expected from shared tuning), and the proportion is preserved across sleep. A failure of this prediction would refute the Lagrangian form of the circuit.
\item \textbf{Threshold and hold} (Theorem~\ref{thm:structure}ii). The rate of an interneuron recorded in vivo, against its pooled excitatory input under graded drive. The rate is zero below a threshold and linear above it, and once its prohibition binds the input is held at the threshold irrespective of the drive. An error unit of predictive coding would carry a residual proportional to the drive.
\item \textbf{Count} (Theorem~\ref{thm:structure}iii). Holographic stimulation of interneurons along many patterns \citep{sadeh2020}. The number of patterns that lower the interneurons' own steady rate equals the number of unstable directions of the excitatory network estimated with inhibition clamped. Experiment~5 carries out this protocol in simulated circuits (Appendix~\ref{app:exp:circuits}).
\item \textbf{Rhythm law} (Theorem~\ref{thm:rhythm}). Gamma or theta recorded while the carrier of the band is slowed by temperature, by channel pharmacology or by a change of synaptic decay. The band's frequency shifts as $\tau^{-1/2}$ at fixed coupling; the inhibitory conductance lags the excitatory population by $\frac\pi2+\arctan[(d/\tau_x-\rho/\tau_k)/2\omega]$; raising neuromodulatory gain raises coherence.
\item \textbf{Bounded imbalance} (Theorem~\ref{thm:time}i,iv). Long recordings of rivalry and decoded theta sequences. Measured in windows shorter than the leak of adaptation, the cumulative deviation of dominance from its fraction in the long run changes less than in renewal surrogates with the same durations, which diffuse, and its power spectrum falls toward zero frequency down to the leak rate.
\item \textbf{Sign flip} (Theorem~\ref{thm:learning}i,ii). Plasticity of excitatory synapses onto interneurons in sleep and in waking, and artificial phonotactic learning. These synapses potentiate on patterns active in sleep and depress on patterns active in waking, with more total depression than potentiation per unit of frequency ($r>1$) where rules are learned. Blocking interneuron plasticity during sleep impairs the learning of rules from positive examples and spares associations. Quantitatively (Theorem~\ref{thm:learning}ii, assuming stochastic replay), a pattern that waking experience shows with frequency $p$ is replayed with frequency $\min(q_0,r\,p)$, where $q_0$ is its replay frequency without its own inhibitory weight, so the replay of rare but possible sequences, relative to their waking frequency, measures the ratio $r$ of Eq.~\eqref{eq:ratio}.
\item \textbf{Dreams converge} (Theorem~\ref{thm:learning}iv). Replay across nights of learning a structured task, such as a grammar or a maze with forbidden transitions. The fraction of sleep reactivations that violate the task's rules declines toward zero, and the inhibitory changes that accompany learning stop when it vanishes. A rule that replay never violates, because the others already exclude it, acquires no inhibitory trace of its own until a waking task makes the circuit attempt it.
\item \textbf{Apical costate} (Theorem~\ref{thm:future}). Dendritic imaging of place cells during navigation to a goal that is then moved. Apical input encodes the distance to the current goal, and after the goal moves it changes before the somatic tuning does.
\item \textbf{Bridge} (Theorem~\ref{thm:future}). Theta sequences directed at a goal, decoded within each cycle. Their positional variance is largest midway between the animal and the goal and vanishes at both ends, as the variance $2Dt(T-t)/T$ of a plan of duration $T$ does, while a forward model alone spreads its variance without bound.
\item \textbf{Ramp} (Theorem~\ref{thm:time}v). Contradictory evidence, such as an impossible figure or an inconsistent rule set, presented while the circuit that holds the violated rules is recorded. Inhibitory activity in that circuit ramps linearly until the interpretation switches, and the rate of the ramp grows with the size of the contradiction.
\end{enumerate}

\section{Related Work in Full}
\label{app:related}

This section places the theory among ten lines of earlier work, from the multipliers of mechanics and economics to reasoning networks and accounts of cognition, and each subsection ends with what the Lagrangian brain adds to its line. Most ingredients of the theory exist separately: saddle flows that compute prices, interneurons read as multipliers, constraint satisfaction as a model of thought, herding, learning with a clamped and a free phase, and planning as inference. The theory joins them in one flow whose constraints are learned, and the same mechanics then accounts for inhibition, rhythms, sleep and planning in the brain and for reasoning in machines.

\subsection{Multipliers, saddle flows and prices}
\label{app:related:prices}

Lagrange multipliers as the forces of constraints are the foundation of analytical mechanics \citep{lagrange1788,lanczos1949}. Gauss described motion under constraints as the least deviation from free motion \citep{gauss1829}, and Hertz built a mechanics in which constraints take the place of forces \citep{hertz1899}. In economics, prices that adjust to excess demand compute an equilibrium \citep{walras1874,samuelson1941}, and Lange proposed that a planning board could do the same by raising prices where shortages appear and lowering them where surpluses accumulate \citep{lange1936}. Arrow, Hurwicz and Uzawa wrote this adjustment as the saddle flow of a Lagrangian \citep{arrow1958,kose1956}, and its convergence has since been established under general conditions \citep{feijer2010,cherukuri2017}. Multiplier methods and Lagrangian relaxation adjust the multiplier of each constraint by the size of its violation \citep{bertsekas1982,fisher1981}, and the running average of their iterates satisfies the constraints at rate $1/T$ \citep{nedic2009}. Players of a repeated game who adapt with low regret reach equilibrium on average in the same way \citep{freund1999,abernethy2017}, and the conditional gradient method has a similar reading \citep{jaggi2013}. Prices coordinate decentralized decisions in decomposition methods \citep{dantzig1960}, in the control of congestion on the internet \citep{kelly1998,low1999} and, in Hayek's account, in a whole economy \citep{hayek1945}. In machine learning, Lagrangian methods train networks under constraints \citep{nandwani2019,fioretto2021}, and in reinforcement learning with constraints the multiplier acts as an integral controller whose oscillations can be damped by proportional and derivative terms \citep{stooke2020}. In the Lagrangian brain the same oscillation is a brain rhythm, with a frequency fixed by two time constants (Theorem~\ref{thm:rhythm}).

\subsection{Multipliers in neural circuits}
\label{app:related:circuits}

Analog circuits that compute with differential multipliers \citep{platt1988,zhang1992lpnn} and networks that solve optimization problems by descending an energy \citep{hopfield1985} brought these ideas to neural computation. Seung and colleagues wrote the dynamics of excitatory and inhibitory populations coupled through transposed weights in minimax and dissipative Hamiltonian form, and gave the conditions under which they converge; outside them, limit cycles can be stable \citep{seung1998minimax}. Interneurons have been read as the multipliers of statistical constraints such as whitening \citep{pehlevan2019}. Spiking networks have been designed as solvers of convex programs whose inhibitory thresholds are the faces of a polytope \citep{boerlin2013,mancoo2020,deneve2016}, and sparse coding by competition through inhibition belongs to the same family \citep{rozell2008}. In these models the constraints are chosen by the designer, and the rotation of the saddle flow is treated as a matter of stability. The Lagrangian brain takes the constraints to be the content of knowledge, closed under omission and learned from experience, makes every variable of a circuit other than its excitatory state a multiplier, inhibition included, and identifies the rotation with the rhythms of the brain.

\subsection{Inhibition and balance}
\label{app:related:balance}

Balanced networks explain the irregular activity of cortex as the result of strong excitation cancelled by strong inhibition \citep{vanvreeswijk1996,brunel2000}, and the theory of networks stabilized by inhibition explains paradoxical and supralinear responses \citep{tsodyks1997,ozeki2009,rubin2015,sanzeni2020}. Reviews of cortical inhibition describe its many cell types, its dense connectivity, its targeting of somata and dendrites, and the disinhibitory circuits that gate plasticity and attention \citep{isaacson2011,kepecs2014,tremblay2016,letzkus2015}. Inhibitory plasticity establishes balance and matches inhibitory to excitatory tuning \citep{vogels2011,damour2015,hennequin2017}, and theories of circuits with several classes of interneuron explain how perturbing one class propagates through the others \citep{litwinkumar2016}. Efficient balanced networks interpret balance as the signature of a code that tracks its own errors \citep{deneve2016}. The Lagrangian brain reads balance as the complementary slackness of a constrained optimum, which leads to two predictions: transposed weights for every interneuron that carries knowledge, preserved by learning and sleep, and a count of paradoxical directions.

\subsection{Constraint satisfaction in cognition}
\label{app:related:constraints}

Cognitive science has long described perception and thought as the satisfaction of many constraints at once. Marr and Poggio computed stereo depth with a cooperative network in which excitation and inhibition encode the constraints of uniqueness and continuity \citep{marr1976}. Interactive activation explained the perception of words by excitation among consistent hypotheses and inhibition among inconsistent ones \citep{mcclelland1981}, and Rumelhart, Smolensky, McClelland and Hinton described schemata as the settling of a network that satisfies soft constraints \citep{rumelhart1986schemata}. Harmony theory and Optimality Theory treat cognition and grammar as the satisfaction of ranked or weighted constraints \citep{smolensky1986,prince2004,smolensky2006}, and analogy and explanation have been modelled as the maximization of coherence \citep{holyoak1989,thagard1989,thagard1998}. These models give each constraint a fixed weight. In the Lagrangian brain each constraint has a multiplier that moves, so a violated constraint gains strength until it holds, and the constraints themselves are learned by sleeping and waking.

\subsection{Belief, sampling and herding}
\label{app:related:belief}

Theories of neural uncertainty represent distributions in population activity \citep{ma2006,knill2004} or by sampling, with activity at each moment a sample from a posterior \citep{hoyer2003,fiser2010,berkes2011,buesing2011,orban2016}, and human judgements show the signatures of reasoning from few samples \citep{vul2014,sanborn2016}. Predictive coding and the minimization of free energy propagate errors weighted by their precision \citep{rao1999,friston2010,bastos2012,keller2018}. Herding replaces random samples by a deterministic sequence that accumulates the errors of its moments and matches them at rate $1/T$ \citep{welling2009herding,chen2010super}. It is equivalent to the conditional gradient method \citep{bach2012equivalence}, and the fast weights of persistent contrastive divergence speed the mixing of Boltzmann machines in a related way \citep{tieleman2009}. Aitchison and Lengyel showed that inhibitory cells can carry the momentum of a Hamiltonian sampler whose positions are carried by excitatory cells, so that excitation and inhibition become conjugate variables, with balance and oscillation as consequences \citep{aitchison2016}, and networks trained to sample quickly develop gamma oscillations, normalization and quenched variability \citep{echeveste2020}. A multiplier is a conjugate variable of the same kind whose content is a constraint, and Theorem~\ref{thm:time} places herding inside circuits with knowledge and prohibitions.

\subsection{Sleep and learning}
\label{app:related:sleep}

The Boltzmann machine learns by contrasting a clamped phase with a free phase \citep{ackley1985,hinton2002}, as do contrastive Hebbian learning \citep{movellan1991,xie2003} and equilibrium propagation \citep{scellier2017}. The wake sleep algorithm trains a generative model in one phase and a recognition model in the other \citep{hinton1995}. Crick and Mitchison proposed that dream sleep unlearns spurious memories \citep{crick1983}, and Hopfield and colleagues showed that unlearning the states a network reaches from noise suppresses its spurious attractors \citep{hopfield1983unlearning}. The synaptic homeostasis hypothesis holds that sleep scales synapses down \citep{tononi2014}. Other accounts describe dreaming as the optimization of a generative model shielded from the senses \citep{hobson2012}, as a defence against overfitting \citep{hoel2021}, or as the production of perturbed and adversarial versions of experience from which cortex learns \citep{deperrois2022}, and a phase of replay modelled on sleep reduces forgetting in artificial networks \citep{tadros2022}. Comparing a circuit's own productions with the world's resembles noise contrastive estimation \citep{gutmann2010}, with the circuit as its own source of noise. In these accounts the phase without the world changes the weights that carry associations. In the Lagrangian brain it writes inhibition, and Theorem~\ref{thm:learning} identifies what is learned: the information projection of the circuit's prior, the distribution closest to it in relative entropy among those that never exceed $r$ times the world's frequency of any pattern \citep{jaynes1957,csiszar1975,dudik2007}. As $r$ grows, the rule becomes the elimination algorithm that learns conjunctions of short clauses from positive examples \citep{valiant1984}. Constraint acquisition learns constraint networks from solutions and queries \citep{beldiceanu2012,bessiere2017,deraedt2018}, and inductive logic programming learns rules from examples \citep{muggleton1991}; the rule here is local and runs on the circuit's own dreams.

\subsection{Planning, dendrites and credit assignment}
\label{app:related:planning}

Continuous attractor networks \citep{amari1977,zhang1996,wu2008,khona2022} are the modules of the theory, and adaptation makes their bumps travel and anticipate \citep{mi2014}. Hippocampal sequences, theta sequences and replay have been read as planning and as the sampling of possible futures \citep{johnson2007,pfeiffer2013,wikenheiser2015,olafsdottir2018,kay2020}, and replay has been modelled as the prioritized backup of values \citep{mattar2018}. Planning as inference \citep{attias2003,toussaint2006,kappen2005,todorov2009,botvinick2012,levine2018}, the successor representation \citep{dayan1993,stachenfeld2017} and Schr\"odinger bridges \citep{schrodinger1931,leonard2014,chen2021liaisons,debortoli2021} supply the two waves of Theorem~\ref{thm:future}. Theories of dendritic computation propose that apical dendrites carry the signals for learning \citep{guerguiev2017,sacramento2018,richards2019,payeur2021,poirazi2020}, and that the coupling of the apical and somatic compartments underlies conscious processing \citep{aru2020}. Theorem~\ref{thm:future} identifies the apical signal with the costate, which both plans and assigns credit.

\subsection{Reasoning networks, depth and length}
\label{app:related:reasoning}

Recurrent networks that refine an answer for a fixed number of steps solve Sudoku and mazes after training on pairs of puzzles and solutions \citep{palm2018,wang2019satnet,yang2023recurrent,wang2025hrm,jolicoeur2025trm,sghaier2026ptrm}, and energy minimization has been used to learn iterative reasoning \citep{lecun2007ebm,du2022irem,du2024ired}. Differentiable solvers place an optimization problem inside a network \citep{amos2017optnet,bai2019deq}, message passing networks learn to decide satisfiability and to search for solutions of constraint problems \citep{selsam2019neurosat,tonshoff2023}, and neural algorithmic reasoning trains networks to execute classical algorithms \citep{velickovic2021}. Dense associative memories store many more patterns than they have neurons \citep{krotov2016}, and attention is one step of such a memory \citep{ramsauer2021}. Networks trained on small instances extrapolate to larger ones when they are allowed more iterations \citep{schwarzschild2021,bansal2022}, adaptive computation lets a network choose its own depth \citep{graves2016act,dehghani2019universal}, and looped Transformers can simulate programs and extrapolate on tasks with short iterative solutions \citep{giannou2023,zhou2024,fan2025looped,saunshi2025}; recurrence in depth has been scaled to language models \citep{geiping2025}. Transformers of fixed depth fail to extrapolate parity and variable tracking to longer inputs \citep{anil2022}, and their positional encodings shape how far they extrapolate \citep{kazemnejad2023}. Theory bounds what networks of fixed depth can compute: Transformers with logarithmic precision lie in $\TCz$ \citep{merrill2023parallelism,strobl2024}, self attention has limits on periodic and hierarchical structure \citep{hahn2020}, and chains of thought add depth at the cost of generated tokens \citep{wei2022cot,feng2023,merrill2024cot}. Networks trained on short inputs fail on long ones across the Chomsky hierarchy \citep{deletang2023}, Transformers learn shortcuts to automata that do not extrapolate \citep{liu2023shortcuts}, and state space models cannot track the state of groups that are not solvable \citep{merrill2024illusion} unless their transitions may have negative eigenvalues \citep{grazzi2025}. Message passing with a fixed number of rounds cannot decide properties beyond its radius \citep{loukas2020}. The multiplier network learns a law and lets it run until it rests, which places depth in time and exactness in the form of the law.

\subsection{Constraint solvers in neural and physical systems}
\label{app:related:solvers}

Local search with constraint weights \citep{selman1992gsat,selman1994noise,minton1992,morris1993breakout,shang1998} is the closest classical relative of the search of a multiplier network, and clause weighting that smooths the weights toward their mean \citep{hutter2002} corresponds to multipliers with leak. Analog dynamical systems in which auxiliary variables grow with the violation of each clause solve satisfiability problems \citep{ercseyravasz2011}, Ising machines minimize energies in physical hardware \citep{mohseni2022}, and auxiliary variables that correct the heterogeneity of amplitudes help analog spin systems escape local minima \citep{leleu2019}. Spiking networks have solved constraint satisfaction problems, Sudoku included, by stochastic dynamics \citep{habenschuss2013,jonke2016,fonsecaguerra2017}, by circuits of winner take all modules with distributed constraints \citep{rutishauser2018}, and by rhythmic inhibition that lets a network leave inconsistent states \citep{mostafa2015}. In these systems the constraints are given. In the multiplier network they are learned from solutions by a local rule, the search is deterministic and carried by integrating multipliers, and the same flow accounts for rhythms, sleep and planning.

\subsection{Accounts of cognition}
\label{app:related:cognition}

Marr separated the computation a system performs from the algorithm and the implementation that realize it \citep{marr1982}, and Tolman showed that animals hold maps of their world that guide behaviour beyond any chain of responses \citep{tolman1948}. The symbolic account treats cognition as the manipulation of structured symbols \citep{newell1976,fodor1988,marcus2001,lake2017}, the connectionist account as the settling of networks with learned weights \citep{rumelhart1986,smolensky1988,smolensky1990}, and the dynamical account as the unfolding of a dynamical system in time \citep{port1995,vangelder1998,kelso1995}. Proposition~\ref{prop:limits} obtains the symbolic and the connectionist accounts as limits of one flow and states the law of the dynamical account. Cybernetics described the organism as a regulator that keeps its essential variables within limits \citep{wiener1948,ashby1952,cannon1932}, defined a law of nature as a constraint \citep{ashby1956}, and argued that every good regulator of a system is a model of it \citep{conant1970}. The internal model principle made this last idea exact for integral control \citep{francis1976}. The present theory carries this line of thought from regulation to knowledge.

\clearpage
\phantomsection\label{app:part2}
\begin{center}{\large\textsc{Part II. Machines and Experiments}}\end{center}
\vspace{2pt}
\suppressfloats[t]

\section{The Multiplier Network}
\label{app:machine}

This section specifies the machine of Experiments~1 to~4. A multiplier network computes only by the flow of Eq.~\eqref{eq:flow} at high gain, with one multiplier for each learned constraint, and its answer is the state at which it rests. The first subsection describes how the network works, with its algorithm and a traced trajectory; the others give its knowledge, its thought, its learning in sleep and waking, its planner, its chains and its implementation in full. The insight to take from it is that two integrating multipliers, a signed one for each invariant and one for each prohibited pair, carry the state out of the traps in which local search stalls.

\subsection{How the network works}
\label{app:machine:overview}

\begin{figure}[t]
\centering
\includegraphics[width=\textwidth]{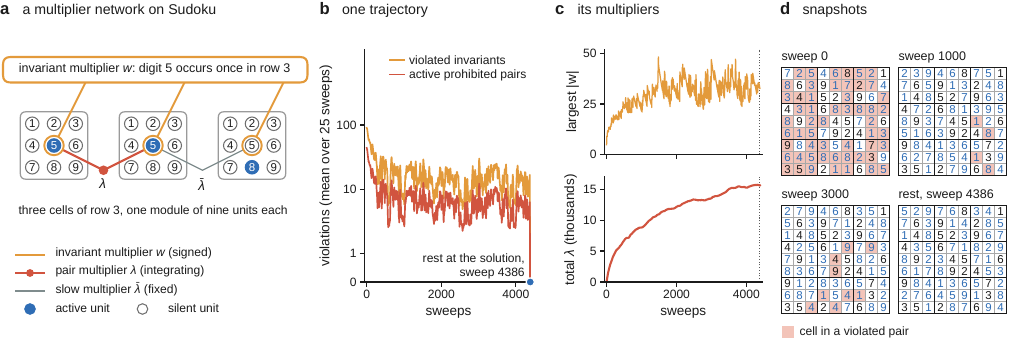}
\vspace{-5mm}
\caption{\textbf{How a multiplier network thinks.} \textbf{(a)}~Architecture on Sudoku. Each cell is a module of nine units, one per digit. An invariant, such as ``digit 5 occurs once in row 3'', has a signed multiplier that pushes its group toward exactly one active unit. A prohibited pair, such as two 5s in one row, has an inhibitory multiplier that integrates the joint activity of the pair. A slow multiplier, fixed during thought, inhibits every unit in conflict with an active one. \textbf{(b)}~One trajectory on a Sudoku-Extreme puzzle (mean over 25 sweeps). The number of violated constraints drops in the first sweeps, then fluctuates while multipliers build up and release, and reaches zero when the state comes to rest at the solution. \textbf{(c)}~The multipliers during the same trajectory: the largest invariant multiplier (top) builds up and is released again and again, and the total integrating inhibition (bottom) accumulates while violations persist. \textbf{(d)}~Snapshots of the grid; shading marks the cells in a violated pair.}
\label{fig:machine}
\vspace{-2mm}
\end{figure}

A multiplier network has units, modules, knowledge and multipliers (Figure~\ref{fig:machine}). A task lays its instances on \emph{sites}, and each site carries a module whose units are its possible values: a digit for each Sudoku cell, a state for each position of a string. At high gain the state has exactly one active unit per module. Units given by the instance, such as the digits printed in a puzzle, are clamped.

\begin{algorithm}[t]
\caption{The multiplier network (one trajectory)}
\label{alg:machine}
\begin{algorithmic}
\STATE {\bfseries Input:} clamps; invariants $(P,m_P)$, prohibited pairs and slow multipliers $\bar\lambda$; rates $\eta_P,\rho_P,\eta_Q,\rho_Q$; step $\Delta t=1/(\text{number of free modules})$
\STATE set free modules at random; $w\leftarrow0$; $\lambda\leftarrow0$; quiet $\leftarrow0$
\REPEAT
\STATE changed $\leftarrow$ false
\FOR{each free module $i$, in random order}
\STATE $h_a\leftarrow\sum_{P\ni a}w_P-\sum_b(\bar\lambda_{ab}+\lambda_{ab})x_b$ for the units $a$ of $i$
\STATE switch $i$ to $\arg\max_ah_a$ if it is strictly better than the current unit
\STATE $w_P\leftarrow w_P+\Delta t\,[\eta_P(m_P-n_P)-\rho_Pw_P]$
\STATE $\lambda_{ab}\leftarrow\lambda_{ab}+\Delta t\,[\eta_Qx_ax_b-\rho_Q\lambda_{ab}]$
\ENDFOR
\STATE quiet $\leftarrow0$ if a module changed, otherwise quiet $\leftarrow$ quiet${}+1$
\UNTIL{quiet $\ge10$ and no multiplier is rising}
\STATE {\bfseries return} the state at rest
\end{algorithmic}
\end{algorithm}

\textbf{Knowledge.} Knowledge is learned from solved instances and takes three forms, which are the invariants and prohibitions of Table~\ref{tab:dictionary} at high gain. An \emph{invariant} is a group of units whose summed activity is the same in every training solution, for example the nine units that put the digit 5 in the nine cells of a row. A \emph{prohibited pair} is a pair of units that are never active together inside an invariant. A \emph{window law} is a pattern over a small window, such as a cell and its neighbours, that never occurs in training. Learning takes one of two forms. \emph{Counting}, rule L0, is the limit of Theorem~\ref{thm:learning} in which every pattern is conjectured and every pattern seen is refuted at once. \emph{Sleep and wake}, rule L1, is the rule of Theorem~\ref{thm:learning} itself, in which the network conjectures only what it dreams. Counting records as an \emph{exclusion} every pair of units that are never active together in training, and finds the invariants as the maximal groups of mutually exclusive units whose sum never varies, with the clique search of Bron and Kerbosch \citep{bron1973}. Only the exclusions inside an invariant act, as prohibited pairs. The others, such as pairs that fail to occur together by chance, stay provisional; in Sudoku every true prohibition lies inside an invariant, and no chance exclusion does.

\textbf{Thought.} Thought is the flow of Eq.~\eqref{eq:flow} at high gain with three multipliers. The input of unit $a$ is that of Eq.~\eqref{eq:machine},
\[
h_a=\sum_{P\ni a}w_P\;-\;\sum_b\,\big(\bar\lambda_{ab}+\lambda_{ab}\big)\,x_b ,
\]
where $w_P$ is the signed multiplier of invariant $P$, which integrates $m_P-n_P$, the difference between the group's target count and its number of active units; $\lambda_{ab}$ is the integrating multiplier of the prohibited pair $(a,b)$, which grows while both units are active; and $\bar\lambda_{ab}$ is a slow multiplier, the learned inhibition of experience, fixed during thought and equal to the number of invariants that contain both units. In each sweep the free modules are updated one at a time, in a fresh random order; each takes the unit of largest input and keeps its current unit unless another is strictly better (Algorithm~\ref{alg:machine}). Both integrating multipliers leak slowly, so they remember conflicts for hundreds (invariant multipliers) to thousands (pair multipliers) of sweeps and forget them eventually. The network rests when no module has changed for ten sweeps and none of its multipliers is rising, and the answer is read then. No procedure outside the network checks the rules: its own multipliers tell when it is done. There is no noise, no gradient, no backtracking and no occupancy multiplier on single units.

\textbf{Why it works.} At a solution every invariant holds exactly one active unit and no prohibited pair is active, so the integrating multipliers decay. Each alternative unit of a module receives the slow multiplier from every active unit it conflicts with, and the active unit receives none, so once every invariant multiplier lies below one half in magnitude the solution is a strict rest point (Theorem~\ref{thm:closure}i). Away from a solution, groups with too many active units push their members down, empty groups pull their members up, and every pair that keeps conflicting accumulates a multiplier that makes the conflict harder to reenter. With leak the multipliers saturate, so the memory of a single mistake fades (Theorem~\ref{thm:time}). The invariant multipliers are the weights of violated constraints and the pair multipliers the memory of recent conflicts, the two ingredients that let local search escape its traps in breakout and discrete Lagrangian search \citep{morris1993breakout,shang1998,selman1994noise}; in the network they are the multipliers of invariants and prohibitions of Table~\ref{tab:dictionary}.

\subsection{Knowledge}
\label{app:machine:knowledge}

Clamped modules (givens, inputs, walls) keep their unit throughout a run. The three forms of knowledge are defined and found as follows.
\begin{itemize}[leftmargin=*,itemsep=1pt,topsep=2pt]
\item \textbf{Invariants.} Maximal sets of units in different modules that are never active together and whose summed activity is the same constant $m_P$ in every training solution. They are the maximal cliques of the graph of exclusions that pass the constancy test; a branch of the clique search is cut as soon as some training solution has no active unit in the current clique or among its remaining candidates. Sudoku has 243 invariants of nine units each: every digit occurs once in every row, column and box. That every cell holds one digit is enforced by its module.
\item \textbf{Prohibited pairs.} Exclusions inside an invariant. An exclusion that belongs to no invariant is \emph{provisional} and never acts: with $N$ examples some allowed pairs fail to occur together by chance, and Theorem~\ref{thm:closure}(ii) bounds the number of examples after which none remains.
\item \textbf{Window laws.} For tasks laid out on a map or a chain, the configurations of a window over a given relation (neighbours on a grid; the left neighbour on a chain) that never occur in training.
\end{itemize}

\subsection{Thought}
\label{app:machine:thought}

The three multipliers of Eq.~\eqref{eq:machine} obey
\begin{align*}
\bar\lambda_{ab}&=J_0\,m_{ab} &&\text{slow multiplier: learned inhibition, fixed during thought},\\
\dot\lambda_{ab}&=\eta_Q\,x_ax_b-\rho_Q\,\lambda_{ab} &&\text{integrating multiplier of a prohibited pair: of one sign, leaky, with long memory},\\
\dot w_P&=\eta_P\,(m_P-n_P)-\rho_P\,w_P &&\text{signed multiplier of an invariant},
\end{align*}
where $m_{ab}$ is the number of invariants that contain both $a$ and $b$, and $n_P=\sum_{a\in P}x_a$ counts the active units of invariant $P$.

\textbf{Relaxation.} An updated module breaks ties among better units by a randomly rotated order. Each module update advances time by $1/N_{\rm free}$, so one sweep is one unit of time, and the multipliers are integrated by the Euler method after every module update with this step; the uniform decay of the pair multipliers is applied as a global scale factor, which is exact.

\textbf{Rest.} Under the rest rule, ten quiet sweeps and no rising multiplier, every invariant multiplier and every pair multiplier relaxes toward zero, so the state and the multipliers are at a rest point of their joint flow; the state is then compared with the answer.

\textbf{Constants} (fixed on training puzzles of Sudoku): $J_0=1$, $\eta_P=3$, $\rho_P=3\times10^{-3}$ (saturation $\eta_P/\rho_P=1000$), $\eta_Q=1$, $\rho_Q=3\times10^{-4}$ (pair memory of about 3300 sweeps), one trajectory per puzzle with no limit on the number of updates, and random initial units for free modules. Ablations and local search, which need not come to rest, run with a budget of $2\times10^6$ module updates.

\subsection{Learning by sleep and wake (rule L1)}
\label{app:machine:sleep}

\textbf{Pairs as synapses.} Every pair of units in different modules has an inhibitory weight $F_{ab}\ge0$, held in the synapses of the interneuron that carries the pair; a pair with $F_{ab}>\vartheta$ is a prohibition. Each \emph{night}: (1)~\emph{wake}: the training solutions are presented, and every pair active together in one of them lowers its inhibitory weight by $a_w$ (projected at zero); (2)~\emph{sleep}: $D$ dreams, each a free run of the multiplier network (no clamps, random initial state) under the current prohibitions, with slow multiplier 1, integrating pair multipliers and no invariants, until it rests, by the same rule as thought, or $S_{\max}$ sweeps have passed; every pair active together in the final state of a dream raises its inhibitory weight by $a_s$. After the last night, one waking pass and consolidation: maximal cliques of prohibitions whose summed activity is 1 in every training solution are invariants; a clique whose sum is 0 in some solutions and never 2 is completed by the units active only in those solutions, provided exactly one of them is active in each (Theorem~\ref{thm:learning}iv). Only prohibitions inside invariants act afterwards, as with counting. Values: $a_s=a_w=1$ per occurrence, $D=50$ dreams and all 1000 training solutions per night (so $r=a_w\cdot1000/(a_s\cdot50)=20$: a pair is stored when it is dreamed more than 20 times as often per dream as it occurs per waking solution), $\vartheta=3.5$, $S_{\max}=3000$, 60 nights, and the constants of the solver above.

\textbf{Prohibitions of higher order as interneurons.} For planted worlds every set $S$ of two or three units carries an inhibitory weight $I_S\ge0$, held by an interneuron with threshold $|S|-\tfrac12$. The free phase is one persistent stochastic trajectory at unit gain in which a unit receives its bias minus the inhibitory weights of the patterns it would complete; biases are the signed multipliers of the frequencies of single units, learned with equal rates in waking and sleep. Per night, 100 waking examples drawn in turn from the training set and 100 states of the free trajectory change the inhibitory weights by the rule of Theorem~\ref{thm:learning} with $r=100$, and an inhibitory weight above 1 counts as a prohibition.

\subsection{Planning with the future as a multiplier}
\label{app:machine:plan}

Units are the open cells of a maze, and walls are clamped silent. The goal is activated at $t=0$; a silent unit becomes active at $t+1$ if a neighbour is active at $t$ and its occupancy multiplier is below the threshold $\vartheta=0.5$; an active unit raises its occupancy multiplier by $\eta=4$, which then decays by $(1-\rho)$ per step with $\rho=0.01$, so the unit stays refractory for $\ln(\eta/\vartheta)/\rho\approx208$ steps, longer than the passage of the front. Multipliers are stored as logarithms to avoid underflow in large mazes. From the start, the path repeatedly moves to the neighbour with the smallest multiplier. Maze-Hard scores exact match with a stored path, one of possibly several shortest paths, so ties are broken by a preference order among the four directions, and the front starts either at the goal or at the start with the path reversed; both choices are fitted on the training paths. The control without a costate relaxes the learned window law on units with two values (on the path, off the path) with a bias toward fewer path units and no occupancy multiplier.

\subsection{Chains}
\label{app:machine:chain}

For state tracking, module $i$ holds a state $s_i$ and the input $x_i$ is clamped; the law forbids the triples $(s_{i-1},x_i,s_i)$ never seen in training, and the predecessor of the first position is the identity. In each synchronous sweep module $i$ takes the state that its left window $(s_{i-1},x_i)$ allows; the run rests when nothing changes. Read in the direction of the sequence, the chain rests at the running products after $n$ sweeps (Appendix~\ref{app:proofs:depth}). The law has $|Q|^2|\Sigma|$ entries, for a set $Q$ of states and an alphabet $\Sigma$, and no parameter that depends on length.

\subsection{Implementation}
\label{app:machine:impl}

The flow, the counting rule and rule L1 are implemented in NumPy with Numba kernels, one trajectory per CPU core. A Sudoku trajectory costs a few hundred integer and floating point operations per module update. Counting the first 1000 training solutions and the clique search take well under a second.

\section{The Experiments in Full}
\label{app:experiments}

This section gives each experiment in full, with its figure and every detail of its protocol, and ends with the numerical checks of the theorems, the compute and the full result tables. Each experiment is described by what it tests, the data, how the network learns or thinks, the measures, what happens and what it shows. Experiments~1 to~4 share one design: the machine learns only from solved instances or lawful states, without gradients, and thinks by one trajectory per instance that runs until it rests, with constants fixed on training data before any test run and none that depends on the size of an instance.

\subsection{A guide to the experiments}
\label{app:exp:overview}

If knowledge is a set of prohibitions carried by multipliers, a network built from multipliers alone should be able to do four things. It should learn the laws of a world from examples of what happens, with no counterexamples. It should find solutions by letting its multipliers move, with no search procedure besides the flow. It should plan by giving the future a multiplier. And it should apply a law learned on small instances at every size. Experiments~1 to~4 test these claims one at a time with a single machine, the multiplier network of Appendix~\ref{app:machine}, and Experiment~5 checks two of the circuit predictions in simulation (Table~\ref{tab:overview}).

Each experiment is built so that a wrong theory would fail it. In Experiment~1 the learner sees only lawful examples, so any law it finds was proposed by its own dreams; the theory also predicts the kind and the rate of its errors, and both can be measured. Experiment~2 asks the flow to solve all 422{,}786 test puzzles of Sudoku-Extreme, a public benchmark that includes the hardest known puzzles, from rules it counted itself; ablations then remove one multiplier at a time to show which part of the dynamics does the work. Experiment~3 gives the same machine mazes with and without a multiplier on the goal, and then mazes with 64 times the area of those in training. Experiment~4 trains on strings of length at most 40 and tests at length 1000 on the group $S_5$, where no family of Transformers of fixed depth is exact at every length unless $\TCz=\NCo$ (Theorem~\ref{thm:depth}).

\begin{table}[t]
\centering\footnotesize
\caption{\textbf{The experiments at a glance.} Each experiment tests one claim of Section~\ref{sec:consequences}, listed with its theorems.}
\label{tab:overview}
\setlength{\tabcolsep}{3.5pt}
\renewcommand{\arraystretch}{1.15}
\begin{tabular}{@{}>{\raggedright\arraybackslash}p{2.4cm}>{\raggedright\arraybackslash}p{3.3cm}>{\raggedright\arraybackslash}p{3.4cm}>{\raggedright\arraybackslash}p{3.1cm}>{\raggedright\arraybackslash}p{3.9cm}@{}}
\toprule
experiment & claim & task & learned from & main result\\
\midrule
1 laws from dreams (\S\ref{app:exp:dreams}) & sleep proposes laws and waking refutes the false ones (Th.~\ref{thm:learning}) & worlds with hidden forbidden pairs and triples; the rules of Sudoku & 1000 lawful states, no counterexamples & every hidden law recovered; all 243 invariants and 7290 prohibitions of Sudoku recovered by sleep, waking and consolidation\\
2 thinking on Sudoku (\S\ref{app:exp:sudoku}) & no rest on an error, exact rest at a solution (Th.~\ref{thm:structure}, \ref{thm:time}) & Sudoku-Extreme, all 422{,}786 test puzzles & solutions of 1000 training puzzles, by counting & every test puzzle, one trajectory each; ablations; laws learned apart compose\\
3 planning in mazes (\S\ref{app:exp:maze}) & a goal acts through a costate (Th.~\ref{thm:depth}, \ref{thm:future}) & Maze-Hard; perfect mazes up to $240\times240$ & a preference among directions fitted on 1000 training mazes & shortest paths in \res{E3:optimal}\% of test mazes and in \res{E3:large}\% of mazes with 64 times the area; a law without a costate plans none\\
4 state tracking (\S\ref{app:exp:s5}) & a law has no size, and time is depth (Th.~\ref{thm:closure}, \ref{thm:depth}) & word problems of $\mathbb Z_2$, $\mathbb Z_5$, $S_3$, $S_5$ & 10{,}000 strings of length at most 40 & exact at length 1000 on every group; on $S_5$, Transformers exact on no string of length 100 or more\\
5 circuit signatures (\S\ref{app:exp:circuits}) & paradoxical count; balanced rivalry (Th.~\ref{thm:structure}iii, \ref{thm:time}) & simulated circuits stabilized by inhibition; rivalry & nothing & the count holds in \res{E5:paradox} circuits; rivalry as variable as in humans and balanced over minutes\\
\bottomrule
\end{tabular}
\end{table}

The tasks are public benchmarks where such benchmarks exist (Figure~\ref{fig:experiments}). Sudoku-Extreme and Maze-Hard were introduced to measure reasoning in neural networks \citep{wang2025hrm} and are the common test bed of recurrent reasoning models trained by gradient descent \citep{jolicoeur2025trm,sghaier2026ptrm}. The word problem of $S_5$ is the standard test of state tracking in sequence models \citep{liu2023shortcuts,merrill2024illusion,grazzi2025}. The rules of these tasks are laws in the sense of Section~\ref{sec:knowledge}: closed under omission, the same at every site, and never given to the machine.

\subsection{Experiment 1: laws from dreams}
\label{app:exp:dreams}

\textbf{What it tests.} Theorem~\ref{thm:learning} says that a circuit that alternates free and clamped phases learns the laws of its world from positive examples alone. Its errors are then of one kind, prohibitions of patterns its training never showed, and the probability that they forbid a fresh lawful state is at most $\sum_Sp_S(1-p_S)^N$. It also learns economically, storing only what its own dreams violate. A learner that follows the rule should therefore recover hidden laws that it never sees broken, with errors of the predicted kind and size, and it should discover the rules of Sudoku from solved grids alone.

\textbf{The data.} The experiment uses two kinds of world. In a \emph{planted world}, $n$ binary units switch on at random with density 0.3, except that $n/2$ pairs and $n/4$ triples of units never occur together. Each forbidden triple is chosen so that its three pairs are allowed, so it cannot be inferred from pairs. A lawful state is a random pattern that contains no forbidden pattern, drawn by rejection, and the learner receives $N=1000$ lawful states and no forbidden one. We use worlds with $n=12$, 24 and 48 units, three worlds of each size, and for $n=24$ also training sets of $N\in\{125,250,500,1000,2000,4000\}$ states. The other worlds are puzzles: all 576 Latin squares of size $4\times4$, and Sudoku. The units of Sudoku are the 729 pairs of a cell and a digit, its lawful states are the solutions of the first 1000 training puzzles of Sudoku-Extreme, and the learner is told only that each cell holds one digit, nothing about rows, columns or boxes.

\textbf{How it learns.} Learning follows rule L1 (Appendix~\ref{app:machine:sleep}), and each night has a waking part and a sleeping part. In waking the training examples are clamped, and every pattern they contain lowers its inhibitory weight. In sleep the circuit runs free under its current prohibitions, and every pattern its dreams contain raises its weight. In the planted worlds every pair and triple of units has a weight held by an interneuron, the free phase is one long stochastic trajectory, $r=100$, a weight above one counts as a prohibition, and learning lasts 1000 nights. On Latin squares and on Sudoku every pair of units in different cells has a weight held in a synapse. Each night the training solutions are presented once, and then 50 dreams run the multiplier network from random states, with no clamps, until they rest; pairs active together in the final state of a dream raise their weights, which on Sudoku gives $r=20$. After 60 nights a consolidation step turns each group of prohibitions whose summed activity is one in every solution into an invariant, and completes a group that falls short in some solutions with the units active exactly then (Theorem~\ref{thm:learning}iv). Latin squares run with one seed and Sudoku with three.

\textbf{Measures.} In the planted worlds: the minimal laws recovered; the false prohibitions, and the fraction of them that forbid a pattern absent from the training set (predicted: all); and the fraction of 20{,}000 fresh waking states wrongly forbidden, against $\sum_Sp_S(1-p_S)^N$ estimated on fresh states (Theorem~\ref{thm:learning}iii), for every $N$. On Latin squares: the invariants and prohibitions recovered. On Sudoku, per night: the prohibitions stored, their recall of the 7290 counted ones, the false prohibitions carried into the night, the dreams that rest at a valid grid, and the violated true pairs per dream (Figure~\ref{fig:learning}c); after consolidation: the invariants and prohibitions, and the solve rate of the solver of Experiment~2 on its fixed test sample with the knowledge learned in sleep (Table~\ref{tab:e1}).

\begin{figure}[t]
\centering
\includegraphics[width=\textwidth]{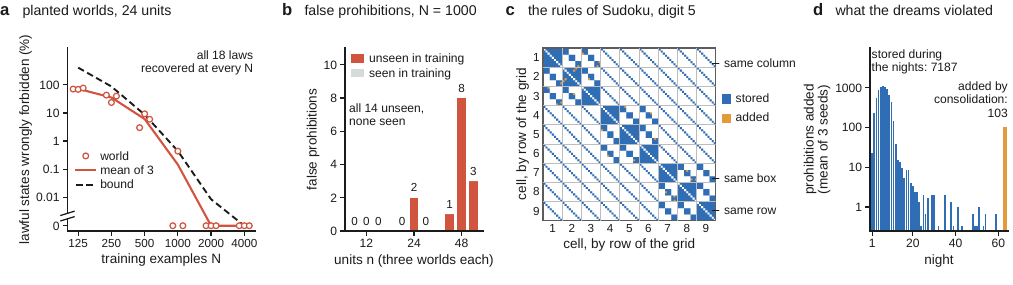}
\vspace{-5mm}
\caption{\textbf{Laws from dreams (Experiment~1).} \textbf{(a)}~Planted worlds with 24 units, three worlds per training size. The fraction of fresh lawful states that the learned laws wrongly forbid falls with the number $N$ of examples and stays below the bound $\sum_Sp_S(1-p_S)^N$ of Theorem~\ref{thm:learning}(iii) in every world (dashed, mean of the three bounds). Every hidden law is recovered at every $N$. \textbf{(b)}~Each false prohibition forbids a pattern that the training set never contains, as the theorem predicts; the panel shows the 14 false prohibitions at $N=1000$, all unseen. \textbf{(c)}~The rules of Sudoku learned by sleep and wake from 1000 solutions, shown as the prohibitions between the 81 cells for one digit after consolidation (blue: stored during the nights; orange: added by consolidation). The rows, columns and boxes of Sudoku appear, although they were never given. \textbf{(d)}~What the dreams violated: prohibitions stored in each night (blue, mean of three seeds) and those added by consolidation (orange), which completes every invariant.}
\label{fig:dreams}
\vspace{-2mm}
\end{figure}

\textbf{What happens.} In all nine planted worlds every hidden law is recovered (\res{E1:k:recall}\%). The learned laws contain \res{E1:k:false} false prohibitions in total, and every one of them forbids a pattern absent from the training set. In the largest worlds they forbid \res{E1:k:mass}\% of fresh lawful states, below the bound of \res{E1:k:pred}\%, and the error falls with $N$ as the bound predicts (Figure~\ref{fig:dreams}a). On Latin squares of size $4\times4$, sleep and wake recover all 32 invariants and 192 prohibitions of their rules. On Sudoku the fraction of dreams that rest at a valid grid rises from \res{E1:sudoku:first}\% in the first night to \res{E1:sudoku:valid}\% in the last (Figure~\ref{fig:learning}c): after about twenty nights the circuit dreams almost only lawful grids, although it has never been told a rule. The stored prohibitions level off at \res{E1:sudoku:recall}\% of the 7290 rules of Sudoku, the ones the dreams violated, and learning stops by itself once the dreams no longer break a rule (Theorem~\ref{thm:learning}iv). Consolidation completes all \res{E1:sudoku:pools} invariants and with them all \res{E1:sudoku:pairs} prohibitions, and with this knowledge the network of Experiment~2 solves \res{E1:sudoku:solve}\% of the test sample.

\textbf{What it shows.} The learner never sees a counterexample, yet it learns what never happens, because its own dreams supply the candidates and waking removes the false ones. It learns economically: the rules its dreams never violated were never stored on their own, and consolidation recovers them as members of the invariants they complete. The rows, columns and boxes of Sudoku emerge from the solutions (Figure~\ref{fig:dreams}c), as the forbidden sound sequences of a language emerge for a learner who hears only lawful words \citep{hayes2008}.

\subsection{Experiment 2: thinking on Sudoku-Extreme}
\label{app:exp:sudoku}

\textbf{What it tests.} Theorems~\ref{thm:structure} and~\ref{thm:time} say that a circuit whose knowledge is prohibitions searches by itself, because no state that violates a law can rest, and that it rests exactly at lawful states. If this is right, the flow alone, with no search procedure, should solve hard problems of constraint satisfaction from knowledge learned from examples, and each of its multipliers should contribute in the way the theory describes.

\textbf{The data.} A puzzle is a $9\times9$ grid with some digits given, and the answer is the completion in which every row, column and $3\times3$ box contains each digit once. Sudoku-Extreme, in its public release with its training puzzles, gathers 422{,}786 test puzzles from eight public collections, including the hardest known ones, and rates each puzzle by the number of backtracks a classical solver needs \citep{wang2025hrm}. Recurrent reasoning models trained by gradient descent on 1000 of its training puzzles reach 55.0\% \citep{wang2025hrm}, 87.4\% \citep{jolicoeur2025trm} and 98.75\% \citep{sghaier2026ptrm} of its test set.

\textbf{How it thinks.} Knowledge is counted by rule L0 from the solutions of the first 1000 training puzzles. The count yields exactly the 243 invariants and 7290 prohibited pairs of Sudoku, together with \res{E2:spurious} chance exclusions that belong to no invariant and stay provisional; if they were kept, they would forbid the true solution of \res{E2:blocked}\% of the fixed sample of 1000 test puzzles used below. Invariants are learned from far fewer examples than exclusions: the first \res{E2:minN} solutions already give exactly the 243 invariants, although \res{E2:minN:excl} of their exclusions are chance. One trajectory runs per puzzle, from a random initial state and with the constants of Appendix~\ref{app:machine:thought}, until the network rests, with no limit on its thinking time. Ablations remove one multiplier at a time or replace the integrating multipliers by noise or by an occupancy multiplier on single units. These variants, and two classical methods of local search with the same learned rules, min-conflicts \citep{minton1992} and breakout \citep{morris1993breakout}, run on the fixed sample with a budget of $2\times10^6$ module updates. To test composition, invariants and prohibitions are counted separately from 1000 standard solutions and from 1000 random grids whose only rule is that each diagonal holds every digit once, and their union is tested, against the standard knowledge alone, on 200 generated puzzles of X-Sudoku, which has both sets of rules; each of these puzzles has exactly one solution.

\textbf{The full test set.} The dataset index of each puzzle is the seed of its dynamics. A trajectory is fixed by its puzzle and its seed, and the rest rule decides only when it stops, so the test set is computed in two passes. The first pass stops each trajectory after ten quiet sweeps or $2\times10^6$ updates and ends \res{E2:full:solved} trajectories at the solution; at a solution every multiplier relaxes, so the rest rule ends each of them at the same sweep. The second pass runs the other 27 trajectories from the start under the rest rule and without a budget, and runs 1000 randomly chosen trajectories of the first kind again under the rest rule, each of which ends with the same grid after the same number of updates (\res{E2:rest:check}).

\textbf{The sample, the ablations and local search.} The sample is a fixed random sample of 1000 test puzzles (seed 0), run with four seeds of the dynamics. The ablations use the sample and one seed, with the slow multipliers kept and constants chosen on training puzzles and listed in the result files. An ablated run stops as soon as every invariant is satisfied, a test that the full network does not use. The variants are: no integrating multipliers; noise from a heat bath in their place, at a temperature chosen on training puzzles; invariant multipliers only; pair multipliers only; an occupancy multiplier on single units only; and pair multipliers without leak. Min-conflicts runs with a tabu tenure of 10 and breakout with constraint weights, on the same sample and with the same budget.

\textbf{Sample efficiency.} Rule L0 counts the first $N$ training solutions, for $N\in\{75,100,150,\allowbreak200,500,1000,2000\}$.

\textbf{Measures.} On the full test set: puzzles solved, and the median and largest number of updates, overall and by source (Table~\ref{tab:e2full}). On the sample: the solve rate, the median number of updates (Table~\ref{tab:e2}), milliseconds per puzzle, and the Spearman correlation between updates and the difficulty rating. For the knowledge: the number of chance exclusions, and the fraction of the sample whose true solution they would forbid if kept. For sample efficiency: the invariants found (true and spurious), the exclusions and provisional ones, and the solve rate on the sample whenever the knowledge differs from that of $N=1000$ (Table~\ref{tab:e2sample}).

\begin{figure}[t]
\centering
\includegraphics[width=\textwidth]{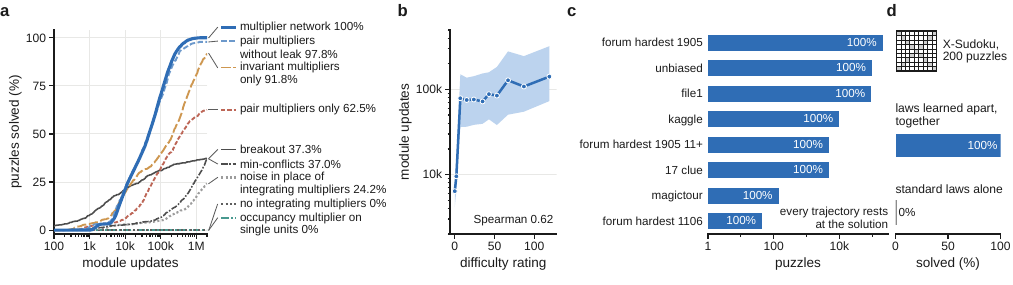}
\vspace{-5mm}
\caption{\textbf{Sudoku-Extreme (Experiment~2).} \textbf{(a)}~Puzzles solved against thinking time, measured in module updates, for the multiplier network, for the network with its integrating multipliers removed or replaced, and for pair multipliers that do not leak; the same learned rules with classical local search are shown for comparison. Shading shows the range of the multiplier network over four seeds; the other curves are single runs. \textbf{(b)}~Thinking time against the difficulty rating of each puzzle, the number of backtracks a classical solver needs (median and interquartile range). \textbf{(c)}~All 422{,}786 test puzzles by source, one trajectory per puzzle run until the network rests: every trajectory rests at the solution. \textbf{(d)}~Composition: prohibitions learned from standard Sudoku and, separately, from grids whose only rule is the two diagonals solve all 200 X-Sudoku puzzles on first encounter; the standard laws alone solve none.}
\label{fig:sudoku}
\vspace{-2mm}
\end{figure}

\textbf{What happens.} Every one of the \res{E2:rest:solved} test puzzles comes to rest at its solution, with one trajectory per puzzle (Table~\ref{tab:main}, Figure~\ref{fig:sudoku}c). Thinking time has a long tail: the median puzzle of the test set needs \res{E2:rest:median} module updates and the slowest \res{E2:rest:maxupd}. On a fixed sample of 1000 test puzzles every seed solves every puzzle (\res{E2:main}\%), in a median of \res{E2:updates} module updates and \res{E2:ms}\,ms on one CPU core. Thinking time grows with difficulty (Spearman correlation \res{E2:spearman}; Figure~\ref{fig:sudoku}b). Both integrating multipliers are needed (Figure~\ref{fig:sudoku}a). With the slow multipliers kept, invariant multipliers alone solve \res{E2:abl:pools}\%, pair multipliers alone \res{E2:abl:pairs}\%, noise in their place \res{E2:abl:noise}\%, an occupancy multiplier on single units \res{E2:abl:rent}\%, and no integrating multiplier at all \res{E2:abl:frozen}\%; pair multipliers without leak solve \res{E2:abl:noleak}\%. Min-conflicts and breakout with the same rules solve \res{E2:minconf}\% and \res{E2:breakout}\%. The laws of standard Sudoku and of the diagonals, learned apart, solve \res{E2:compose}\% of the X-Sudoku puzzles on first encounter, against \res{E2:compose:std}\% for the standard laws alone.

\textbf{What it shows.} The flow alone, with no search procedure outside it, solves every puzzle of Sudoku-Extreme, the hardest known ones among them, from knowledge counted in a fraction of a second. The ablations show which multipliers do the work. An occupancy multiplier on single units erodes correct and incorrect digits alike, while the multiplier of a prohibition acts only on violations. Noise moves the state but forgets where it has been. Classical local search with the same rules solves about a third of the puzzles. Invariant multipliers pull empty groups up and push overfull ones down, pair multipliers remember conflicts, and together they solve every puzzle of the sample. Composition is exact because a prohibition that does not bind contributes no force, so the laws of the diagonals change nothing in a region where they are satisfied. The recurrent reasoning models learn a function of bounded depth from 1000 puzzles with their solutions; the multiplier network learns a law from the same number of solutions and obeys it for as long as a puzzle needs.

\subsection{Experiment 3: planning in mazes}
\label{app:exp:maze}

\textbf{What it tests.} Theorems~\ref{thm:depth} and~\ref{thm:future} say that a goal acts through a multiplier carried backward from the goal, the costate, that at high gain the costate is a front, and that the front reaches each cell at a time equal to its distance from the goal, with parameters that do not depend on the size of the maze. A control that knows what a path looks like and has no costate shows what the goal adds.

\textbf{The data.} Maze-Hard \citep{wang2025hrm} has 1000 training and 1000 test mazes of $30\times30$ cells. Each maze has walls, a start and a goal, and the target is a shortest path between them; the benchmark scores exact match with the stored path. For larger sizes we generate 100 perfect mazes, in which any two cells are joined by exactly one path, at each of the sides 60, 120 and 240, with the recursive backtracker and with start and goal at random open cells whose distance, found by breadth first search, is at least half the side. A maze of side 240 has 64 times the area of a Maze-Hard maze.

\textbf{How it plans.} The planner is that of Appendix~\ref{app:machine:plan}, with the same constants at every size. The units are the open cells. The goal is activated, and a silent unit becomes active when a neighbour is active and its occupancy multiplier is low; every active unit then raises its own occupancy multiplier, which decays slowly and keeps the unit refractory while the front passes. The front reaches each cell once, at its distance from the goal, and leaves behind multipliers that increase with that distance; at high gain these multipliers are a monotone function of the costate of Theorem~\ref{thm:future}, through which the goal acts. The path descends these multipliers from the start. When several shortest paths exist, ties are broken by a preference among the four directions fitted on the training paths; this preference and the choice of the cell where the front starts are all that is fitted on mazes. As a control, rule L0 counts from the 1000 training solutions the window law that every training path obeys, over $3\times3$ windows of (wall, path) labels: a path never enters a wall, every path cell has two path neighbours, and each end has one. The multiplier network without a costate then relaxes this law, with the constants and the budget of $2\times10^6$ module updates used on Sudoku.

\textbf{Measures.} On the test set, the rates of valid paths, shortest paths and exact matches; on the generated mazes, the rates of valid and shortest paths; for the control, the same rates on the test set (Table~\ref{tab:e3}).

\begin{figure}[t]
\centering
\includegraphics[width=\textwidth]{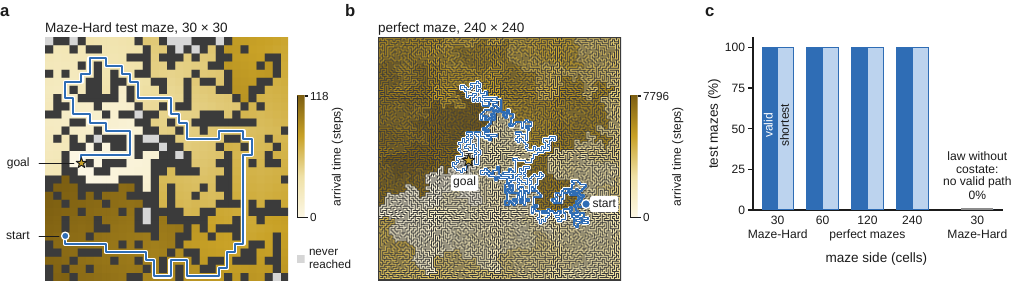}
\vspace{-5mm}
\caption{\textbf{Planning with the future as a multiplier (Experiment~3).} \textbf{(a)}~A Maze-Hard test maze: the front launched at the goal (shading, arrival time; light grey, open cells it never reaches) and the path descended from the start. \textbf{(b)}~A perfect maze of $240\times240$ cells, 64 times the training area, planned by the same rule with the same constants. \textbf{(c)}~Valid and shortest paths at every size, and the same law without a costate, which finds no path.}
\label{fig:maze}
\vspace{-2mm}
\end{figure}

\textbf{What happens.} Paths are valid and shortest in \res{E3:optimal}\% of the Maze-Hard test mazes, and they match the stored path exactly in \res{E3:exact}\%, against 74.5\%, 85.3\% and 86.73\% for the recurrent reasoning models \citep{wang2025hrm,jolicoeur2025trm,sghaier2026ptrm}. Unchanged, the same rule finds a shortest path in \res{E3:large}\% of the perfect mazes at every side up to 240 (Figure~\ref{fig:maze}). The law without a costate rests at a valid path in \res{E3:noprice:valid}\% of the test mazes.

\textbf{What it shows.} The control shows what a goal adds. Locally, a path is only a chain of cells with two neighbours each, and no window of the law can see where the chain should go. The costate supplies the direction, and with it the circuit plans. The front is breadth first search \citep{moore1959} carried by refractoriness, the mechanism of wavefront planning in spiking networks \citep{ponulak2013}, and here it appears as the limit at high gain of the backward wave of Theorem~\ref{thm:future}. Its time grows with the distance to the goal and none of its parameters depends on the size of the maze, so planning extends to any size.

\subsection{Experiment 4: state tracking at every length}
\label{app:exp:s5}

\textbf{What it tests.} Theorems~\ref{thm:closure} and~\ref{thm:depth} say that a law learned on small instances is exact at every size, because it makes no reference to size and runs for as long as the instance requires, and that Transformers of fixed depth cannot track the state of $S_5$ at every length unless $\TCz=\NCo$. State tracking is a building block of reasoning over sequences, from following the moves of a game to keeping track of the variables of a program.

\textbf{The data.} The input is a string of group elements: generators of $\mathbb Z_2$ (parity), $\mathbb Z_5$, $S_3$ or $S_5$. The generators are $0$ and $1$ for $\mathbb Z_2$ and $\mathbb Z_5$, and the identity, a transposition and a cycle of all the objects for $S_3$ and $S_5$. The target at each position is the product so far, which for $S_5$ is one of 120 permutations. Training uses 10{,}000 random strings of length 1 to 40, and the test uses 1000 strings at each of the lengths 40, 100, 500 and 1000. The measure is exact match of the whole string: a string counts as correct only if the target is right at every position (Table~\ref{tab:e4}).

\textbf{How it runs.} The law is counted by rule L0 from the training strings: a triple of state, symbol and next state that never occurs is forbidden. At test time each position is a module holding a state, and the chain relaxes from left to right, each module taking the state that its left neighbour and its own symbol allow, until nothing changes (Appendix~\ref{app:machine:chain}). The baselines are networks trained by gradient descent on the same targets at every position: Transformers with two and six layers (width 128, 4 heads, rotary positions, causal); a looped Transformer, one block with shared weights that receives the input again at every step and is applied $\lceil n/8\rceil$ times or $n$ times to an input of length $n$; and an LSTM with one layer of 256 units. Each is trained with five seeds for 20{,}000 steps of Adam, and the checkpoint is chosen on lengths 30 to 40.

\begin{figure}[t]
\centering
\includegraphics[width=\textwidth]{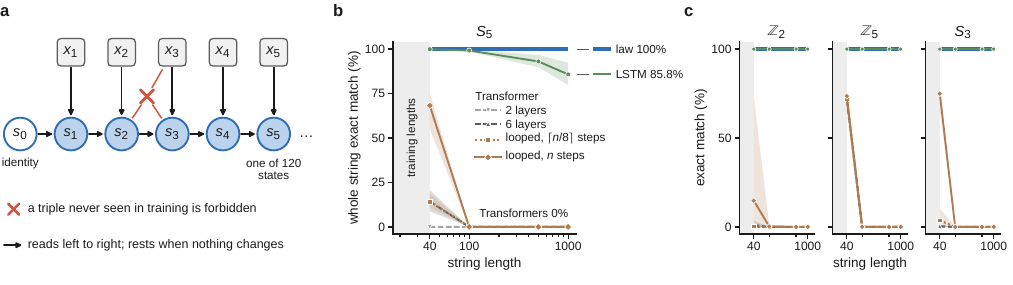}
\vspace{-5mm}
\caption{\textbf{State tracking far beyond the training length (Experiment~4).} \textbf{(a)}~The law for $S_5$ as a chain: each position holds one of 120 states and reads the state to its left and its own symbol; a triple of state, symbol and next state that training never showed is forbidden. \textbf{(b)}~Exact match of the whole string on $S_5$ against length: mean of five seeds per network, shading the range. Training lengths end at 40 (grey band). \textbf{(c)}~The same for $\mathbb Z_2$, $\mathbb Z_5$ and $S_3$.}
\label{fig:tracking}
\vspace{-2mm}
\end{figure}

\textbf{What happens.} The law is exact at every length on every group (Figure~\ref{fig:tracking}). On $S_5$, Transformers of two and six layers and the looped Transformer iterated $\lceil n/8\rceil$ times are exact on at most \res{E4:loop:40}\% of strings of length 40, within the training range. Iterated once per symbol, the looped Transformer reaches \res{E4:loopn:40}\% at length 40 and \res{E4:loopn}\% at length 1000. No Transformer is exact on any string of length 100 or more. The LSTM, which also computes in time, falls from \res{E4:lstm:40}\% at length 40 to \res{E4:lstm}\% at length 1000. On the smaller groups the LSTM stays exact, and the Transformers are exact on at most three strings in a thousand beyond the training length.

\textbf{What it shows.} The Transformers of fixed depth fail on $S_5$ already within their training range. Time is necessary for depth, and the experiment shows that time alone is not enough: the looped Transformer iterated once per symbol has the depth of the law and lacks its exactness, and the LSTM drifts with length. A counted prohibition acts on discrete states, allowing or forbidding each triple, and does not drift. The law holds at length 1000 for the same reason it holds at length 40: it says what may not follow what, and nothing about length.

\subsection{Experiment 5: circuit signatures}
\label{app:exp:circuits}

\textbf{What it tests.} Two predictions of Section~\ref{sec:brain} can be checked in simulated circuits before they are taken to cortex: the paradoxical count of Theorem~\ref{thm:structure}(iii), which prediction P3 carries to cortex, and prediction P5, which follows from Theorem~\ref{thm:time}(i,\,iv): the cumulative imbalance of dominance in rivalry changes less than it would if successive durations were independent.

\textbf{How the count is measured.} We build 200 random circuits of 6 to 12 excitatory units with logistic rates of gain 4, symmetric connections of both signs, and 1 to 11 binding prohibitions with weights drawn from $[0.1,1]$, each carried by a multiplier neuron with leak. The rest point is chosen first, with rates in $[0.3,0.7]$ and multipliers in $[0.5,1.5]$. The curvature of the excitatory network is negative on 0 to 4 directions of the row space of the prohibitions and positive elsewhere, which fixes the connections, and the drive and the bounds then make the chosen point a rest point, with up to four unstable modes held in place by binding prohibitions. The leak is the value in $\{0.02,0.05,0.1,0.2,0.5\}$ with which the circuit settles fastest among those for which the count holds with leak (Appendix~\ref{app:proofs:structure}), and a circuit is kept when it settles within 0.2\,s. The flow of Eq.~\eqref{eq:flow} runs with $\tau_x=10$\,ms and $\tau_k=2$\,ms, integrated by the Runge Kutta method of fourth order with a step of 0.05\,ms. Each circuit starts near its rest point and settles back to it; then a drive of $10^{-4}$ is added to the input of each multiplier neuron in turn, and the steady change of all multipliers gives the matrix of responses. The number of paradoxical patterns is the number of negative eigenvalues of the symmetric part of this matrix; it is compared with the number of unstable modes of the excitatory network with inhibition held fixed, and the simulated responses are compared with the prediction $(S+\rho I)^{-1}$ of Appendix~\ref{app:proofs:structure}.

\textbf{How rivalry is simulated.} Two percepts compete in one module. The dominant percept receives a bonus of 1, each percept has an occupancy multiplier that grows at rate 0.2 per second while it dominates and leaks with a time constant of 30\,s, and each input carries slow noise with a time constant of 0.3\,s and a standard deviation of 0.35, chosen so that dominance durations have the mean and the variability of human rivalry. The module is simulated for 4000\,s in steps of 1\,ms after 30\,s of burn in. The measures are the mean and the coefficient of variation of dominance durations; the root mean square change of the cumulative imbalance of dominance within windows of 1 to 160 full cycles, against 100 renewal surrogates that draw the same durations in shuffled order; and the power of the cumulative imbalance below 0.05\,Hz relative to that of the surrogates (Table~\ref{tab:e5}).

\textbf{What happens.} Every circuit settles, and in every circuit the simulated count equals the number of unstable modes (\res{E5:paradox}); the simulated responses agree with the theory to within 2\%. In rivalry, dominance lasts \res{E5:rivalry:mean}\,s on average, with a coefficient of variation of \res{E5:rivalry:cv}, as in human rivalry \citep{levelt1965,brascamp2006,shpiro2009}. Single durations are as irregular as in perception, yet at every window from 5 to 800\,s the cumulative imbalance of dominance changes less than in each of 100 renewal processes that draw the same durations in shuffled order (Figure~\ref{fig:structure}e), and its power below 0.05\,Hz is \res{E5:rivalry:lowpct}\% of theirs.

\textbf{What it shows.} The number of unstable excitatory modes that inhibition holds in place is written in the responses of the interneurons, and an experiment that drives them pattern by pattern can read it, as P3 proposes for cortex. An occupancy multiplier leaves each dominance period as irregular as perception shows it and still balances the time shares of the two percepts over many minutes. The memory sits in the multiplier, and it is the signature that P5 asks of recordings of rivalry.

\subsection{Numerical checks of the theorems}
\label{app:exp:theory}

Numerical checks that take about a minute produce the theory panels of Figures~\ref{fig:structure}, \ref{fig:learning} and~\ref{fig:future} and the corresponding numbers. The checks are: the rhythm law against direct integration over 2.6 decades of the time constant of the multiplier; herding against independent sampling; rest and ramp on a Sudoku-Extreme puzzle with consistent and with contradictory givens; the dwell fractions under occupancy multipliers against sparsemax (error \res{T:sparse:err}, bound $\sqrt{2\rho}=\res{T:sparse:bound}$); the bridge variance on a lattice; and the relation $q=\min(q_0,r\,p)$ in a world of 18 units with forbidden and rare pairs at $r=1$, 10 and 100.

\subsection{Compute and data}
\label{app:exp:compute}

The multiplier network runs on CPUs, one trajectory per core. The full Sudoku-Extreme test set takes about 25 CPU hours (mean 0.2\,s per puzzle on one core); Experiments~1, 3 and~5 take minutes to hours on a laptop. The networks of Experiment~4 were trained on GPUs. Every number, table and figure of the paper is generated by code from the stored results of the runs, each of which records its configuration. Sudoku-Extreme and Maze-Hard are public datasets released with \citet{wang2025hrm}; the perfect mazes, the strings of the four groups and the X-Sudoku puzzles are generated with fixed seeds.

\subsection{Result tables}
\label{app:exp:tables}

Tables~\ref{tab:e1} to~\ref{tab:e5} list every run of the five experiments.

\begin{table*}[t]
\centering\footnotesize
\caption{Laws from dreams (E1). (a, b) Planted worlds with $n$ units and $N$ training examples: minimal laws, laws recovered, false prohibitions, false prohibitions of patterns unseen in training, and the percentage of fresh waking states wrongly forbidden, measured and predicted by $\sum_Sp_S(1-p_S)^N$. (c, d) Latin squares and Sudoku: true invariants and prohibitions after consolidation, and the share of the true prohibitions stored by sleep and wake before it.}\label{tab:e1}
\begin{tabular}{lrrrrrr}
\toprule
World & Laws & Recovered & False & Unseen & Forbidden (\%) & Predicted (\%) \\
\midrule
(a) planted $n=12$, $N=1000$, world 0 & 9 & 9 & 0 & 0 & 0.00 & 0.01 \\
(a) planted $n=12$, $N=1000$, world 1 & 9 & 9 & 0 & 0 & 0.00 & 0.01 \\
(a) planted $n=12$, $N=1000$, world 2 & 9 & 9 & 0 & 0 & 0.00 & 0.03 \\
(a) planted $n=24$, $N=1000$, world 0 & 18 & 18 & 0 & 0 & 0.00 & 0.12 \\
(a) planted $n=24$, $N=1000$, world 1 & 18 & 18 & 2 & 2 & 0.44 & 1.06 \\
(a) planted $n=24$, $N=1000$, world 2 & 18 & 18 & 0 & 0 & 0.00 & 0.17 \\
(a) planted $n=48$, $N=1000$, world 0 & 36 & 36 & 1 & 1 & 0.47 & 4.79 \\
(a) planted $n=48$, $N=1000$, world 1 & 36 & 36 & 8 & 8 & 3.17 & 4.96 \\
(a) planted $n=48$, $N=1000$, world 2 & 36 & 36 & 3 & 3 & 1.19 & 2.92 \\
(b) planted $n=24$, $N=125$, world 0 & 18 & 18 & 299 & 299 & 70.10 & 418.50 \\
(b) planted $n=24$, $N=125$, world 1 & 18 & 18 & 250 & 250 & 68.20 & 371.28 \\
(b) planted $n=24$, $N=125$, world 2 & 18 & 18 & 340 & 340 & 76.09 & 415.46 \\
(b) planted $n=24$, $N=250$, world 0 & 18 & 18 & 88 & 88 & 42.75 & 89.66 \\
(b) planted $n=24$, $N=250$, world 1 & 18 & 18 & 71 & 71 & 23.36 & 80.51 \\
(b) planted $n=24$, $N=250$, world 2 & 18 & 18 & 78 & 78 & 39.34 & 89.00 \\
(b) planted $n=24$, $N=500$, world 0 & 18 & 18 & 4 & 4 & 3.00 & 6.75 \\
(b) planted $n=24$, $N=500$, world 1 & 18 & 18 & 27 & 27 & 9.26 & 12.20 \\
(b) planted $n=24$, $N=500$, world 2 & 18 & 18 & 5 & 5 & 5.96 & 7.10 \\
(b) planted $n=24$, $N=1000$, world 0 & 18 & 18 & 0 & 0 & 0.00 & 0.12 \\
(b) planted $n=24$, $N=1000$, world 1 & 18 & 18 & 2 & 2 & 0.44 & 1.06 \\
(b) planted $n=24$, $N=1000$, world 2 & 18 & 18 & 0 & 0 & 0.00 & 0.17 \\
(b) planted $n=24$, $N=2000$, world 0 & 18 & 18 & 0 & 0 & 0.00 & 0.00 \\
(b) planted $n=24$, $N=2000$, world 1 & 18 & 18 & 0 & 0 & 0.00 & 0.03 \\
(b) planted $n=24$, $N=2000$, world 2 & 18 & 18 & 0 & 0 & 0.00 & 0.00 \\
(b) planted $n=24$, $N=4000$, world 0 & 18 & 18 & 0 & 0 & 0.00 & 0.00 \\
(b) planted $n=24$, $N=4000$, world 1 & 18 & 18 & 0 & 0 & 0.00 & 0.00 \\
(b) planted $n=24$, $N=4000$, world 2 & 18 & 18 & 0 & 0 & 0.00 & 0.00 \\
(c) Latin squares $4\times4$, seed 0 & \multicolumn{6}{l}{32/32 invariants and 192/192 prohibitions; 100.0\% stored before consolidation} \\
(d) Sudoku, sleep and wake, seed 0 & \multicolumn{6}{l}{243/243 invariants and 7290/7290 prohibitions; 98.6\% stored before consolidation} \\
(d) Sudoku, sleep and wake, seed 1 & \multicolumn{6}{l}{243/243 invariants and 7290/7290 prohibitions; 98.6\% stored before consolidation} \\
(d) Sudoku, sleep and wake, seed 2 & \multicolumn{6}{l}{243/243 invariants and 7290/7290 prohibitions; 98.6\% stored before consolidation} \\
\bottomrule
\end{tabular}
\end{table*}

\begin{table*}[t]
\centering\footnotesize
\caption{Sudoku-Extreme: the fixed sample of 1000 test puzzles under four seeds of the dynamics; ablations and local search with the same learned rules and update budget (one seed); composition on 200 X-Sudoku puzzles. Percent solved, stopped at a wrong grid, and unresolved within the budget; median module updates of solved puzzles.}\label{tab:e2}
\begin{tabular}{lrrrrr}
\toprule
Run & $n$ & Solved & Wrong stop & Unresolved & Median updates \\
\midrule
Multiplier network (seed 0) & 1000 & 100.0 & 0.0 & 0.0 & 44626 \\
Multiplier network (seed 1) & 1000 & 100.0 & 0.0 & 0.0 & 50046 \\
Multiplier network (seed 2) & 1000 & 100.0 & 0.0 & 0.0 & 46174 \\
Multiplier network (seed 3) & 1000 & 100.0 & 0.0 & 0.0 & 48027 \\
Invariant multipliers only (seed 0) & 1000 & 91.8 & 0.0 & 8.2 & 172685 \\
Pair multipliers only (seed 0) & 1000 & 62.5 & 36.2 & 1.3 & 95253 \\
Noise instead of integrating multipliers (seed 0) & 1000 & 24.2 & 34.8 & 41.0 & 584414 \\
Occupancy multiplier on single units only (seed 0) & 1000 & 0.0 & 52.1 & 47.9 & -- \\
No integrating multipliers (seed 0) & 1000 & 0.0 & 100.0 & 0.0 & -- \\
Pair multipliers without leak (seed 0) & 1000 & 97.8 & 2.2 & 0.0 & 45269 \\
Min-conflicts, same rules (seed 0) & 1000 & 37.0 & 0.0 & 63.0 & 549564 \\
Breakout, same rules (seed 0) & 1000 & 37.3 & 0.0 & 62.7 & 6703 \\
X-Sudoku, union of laws learned apart (seed 0) & 200 & 100.0 & 0.0 & 0.0 & 56079 \\
X-Sudoku, standard laws only (seed 0) & 200 & 0.0 & 100.0 & 0.0 & -- \\
\bottomrule
\end{tabular}
\end{table*}
\begin{table*}[t]
\centering\footnotesize
\caption{Sample efficiency of counting: knowledge from the first $N$ training solutions. Same: invariants and prohibitions identical to those from $N=1000$, so the solver behaves identically; otherwise the percentage of the fixed sample solved with one trajectory per puzzle. Seconds: duration of the run, which counts the solutions and finds the invariants, and solves the sample when the knowledge differs from that of $N=1000$.}\label{tab:e2sample}
\begin{tabular}{lrrrrrrr}
\toprule
$N$ & Exclusions & True invariants & Spurious & Prohibitions & Provisional & Solved (\%) & Seconds \\
\midrule
75 & 105450 & 243 & 6 & 7330 & 98120 & 79.3 & 285.3 \\
100 & 79596 & 243 & 0 & 7290 & 72306 & same & 5.2 \\
150 & 46631 & 243 & 0 & 7290 & 39341 & same & 0.3 \\
200 & 28258 & 243 & 0 & 7290 & 20968 & same & 0.1 \\
500 & 7771 & 243 & 0 & 7290 & 481 & same & 0.2 \\
1000 & 7293 & 243 & 0 & 7290 & 3 & same & 0.6 \\
2000 & 7290 & 243 & 0 & 7290 & 0 & same & 1.7 \\
\bottomrule
\end{tabular}
\end{table*}
\begin{table*}[t]
\centering\footnotesize
\caption{The full Sudoku-Extreme test set by source, one trajectory per puzzle run until the network rests, that is, until no module has changed for ten sweeps and no multiplier is rising: puzzles, percent solved, and the median and largest number of module updates.}\label{tab:e2full}
\begin{tabular}{lrrrr}
\toprule
Source & $n$ & Solved (\%) & Median updates & Largest \\
\midrule
all sources & 422786 & 100.0 & 49610 & 3076726 \\
01\_file1 & 93552 & 100.0 & 82940 & 2347830 \\
puzzles0\_kaggle & 9954 & 100.0 & 1504 & 2940 \\
puzzles1\_unbiased & 99749 & 100.0 & 7840 & 137368 \\
puzzles2\_17\_clue & 4855 & 100.0 & 27776 & 577152 \\
puzzles3\_magictour\_top1465 & 143 & 100.0 & 29232 & 173696 \\
puzzles4\_forum\_hardest\_1905 & 209601 & 100.0 & 81240 & 3076726 \\
puzzles5\_forum\_hardest\_1905\_11+ & 4887 & 100.0 & 90573 & 1666167 \\
puzzles6\_forum\_hardest\_1106 & 45 & 100.0 & 113042 & 1704215 \\
\bottomrule
\end{tabular}
\end{table*}

\begin{table*}[t]
\centering\footnotesize
\caption{Maze planning and the control without a costate. Rates are percentages. The front starts at the goal, and ties are broken in the order down, left, right, up; both were chosen on the training mazes.}\label{tab:e3}
\begin{tabular}{lrrrr}
\toprule
Condition & $n$ & Valid & Shortest & Exact \\
\midrule
Maze-Hard $30\times30$, test & 1000 & 100.0 & 100.0 & 100.0 \\
perfect mazes $60\times60$ & 100 & 100.0 & 100.0 & -- \\
perfect mazes $120\times120$ & 100 & 100.0 & 100.0 & -- \\
perfect mazes $240\times240$ & 100 & 100.0 & 100.0 & -- \\
Maze-Hard, law without costate & 1000 & 0.0 & 0.0 & -- \\
\bottomrule
\end{tabular}
\end{table*}

\begin{table*}[t]
\centering\footnotesize
\caption{Exact match of the whole string (percent). Entries for the trained networks average five seeds; the law has one training sample. TF-2 and TF-6 are Transformers with two and six layers; Loop $n/8$ and Loop $n$ apply the looped Transformer $\lceil n/8\rceil$ and $n$ times to an input of length $n$.}\label{tab:e4}
\begin{tabular}{lrrrrrrr}
\toprule
Group & Length & Law & TF-2 & TF-6 & Loop $n/8$ & Loop $n$ & LSTM \\
\midrule
Z2 & 40 & 100.0 & 0.0 & 1.0 & 0.1 & 14.8 & 100.0 \\
Z2 & 100 & 100.0 & 0.0 & 0.0 & 0.0 & 0.3 & 100.0 \\
Z2 & 500 & 100.0 & 0.0 & 0.0 & 0.0 & 0.0 & 100.0 \\
Z2 & 1000 & 100.0 & 0.0 & 0.0 & 0.0 & 0.0 & 100.0 \\
Z5 & 40 & 100.0 & 70.9 & 72.3 & 71.9 & 73.6 & 100.0 \\
Z5 & 100 & 100.0 & 0.0 & 0.0 & 0.0 & 0.1 & 100.0 \\
Z5 & 500 & 100.0 & 0.0 & 0.0 & 0.0 & 0.0 & 100.0 \\
Z5 & 1000 & 100.0 & 0.0 & 0.0 & 0.0 & 0.0 & 100.0 \\
S3 & 40 & 100.0 & 0.0 & 0.4 & 3.6 & 75.0 & 100.0 \\
S3 & 100 & 100.0 & 0.0 & 0.0 & 0.0 & 0.0 & 100.0 \\
S3 & 500 & 100.0 & 0.0 & 0.0 & 0.0 & 0.0 & 100.0 \\
S3 & 1000 & 100.0 & 0.0 & 0.0 & 0.0 & 0.0 & 100.0 \\
S5 & 40 & 100.0 & 0.0 & 14.3 & 14.4 & 68.2 & 99.9 \\
S5 & 100 & 100.0 & 0.0 & 0.0 & 0.0 & 0.0 & 99.2 \\
S5 & 500 & 100.0 & 0.0 & 0.0 & 0.0 & 0.0 & 93.1 \\
S5 & 1000 & 100.0 & 0.0 & 0.0 & 0.0 & 0.0 & 85.8 \\
\bottomrule
\end{tabular}
\end{table*}

\begin{table*}[t]
\centering\footnotesize
\caption{Circuit signatures (Experiment 5): the paradoxical count in simulated circuits and rivalry under occupancy multipliers.}\label{tab:e5}
\begin{tabular}{lr}
\toprule
Check & Result \\
\midrule
Simulated circuits whose paradoxical count equals their unstable modes & 200/200 \\
Simulated circuits that settle back after a perturbation & 200/200 \\
Largest deviation of a simulated response from the theory (relative) & 0.021 \\
Mean dominance duration (s) & 2.55 \\
Coefficient of variation of dominance durations & 0.547 \\
Imbalance power below 0.05\,Hz, relative to renewal & 0.024 \\
Renewal surrogates with more power below 0.05\,Hz & 100/100 \\
\bottomrule
\end{tabular}
\end{table*}

\clearpage

\end{document}